\documentclass{article}
\PassOptionsToPackage{numbers, compress, sort}{natbib}

   \usepackage[final]{neurips_2020}

\usepackage[utf8]{inputenc} 
\usepackage[T1]{fontenc}    
\usepackage{hyperref}       
\usepackage{url}            
\usepackage{booktabs}       
\usepackage{amsfonts}       
\usepackage{nicefrac}       
\usepackage{microtype}      
\usepackage{csquotes}
\usepackage{graphicx}
\usepackage{comment}
\usepackage{algorithm}
\usepackage[noend]{algorithmic}
\usepackage{multicol}
\usepackage{multirow}
\usepackage{makecell}
\usepackage{subcaption}
\usepackage{amsmath}
\usepackage{bbm}
\usepackage{caption}

\title{Hierarchically Organized Latent Modules for Exploratory Search in Morphogenetic Systems}

\author{%
Mayalen Etcheverry\thanks{Source code, videos and additional results can be found at \url{http://mayalenE.github.io/holmes/}}, ~Clément Moulin-Frier, ~Pierre-Yves Oudeyer\\
Flowers Team\\
Inria, Univ. Bordeaux, Ensta ParisTech (France)\\
\texttt{\{mayalen.etcheverry,clement.moulin-frier,pierre-yves.oudeyer\}@inria.fr} \\
}

\newcommand{\sidebysidecaption}[4]{%
\raggedright%
  \begin{minipage}[t]{#1}
    \vspace*{0pt}
    #3
  \end{minipage}
  \hfill%
  \begin{minipage}[t]{#2}
    \vspace*{0pt}
    #4
\end{minipage}%
}

\begin{document}

\maketitle

\begin{abstract}
Self-organization of complex morphological patterns from local interactions is a fascinating phenomenon in many natural and artificial systems. In the artificial world, typical examples of such morphogenetic systems are cellular automata. Yet, their mechanisms are often very hard to grasp and so far scientific discoveries of novel patterns have primarily been relying on manual tuning and ad hoc exploratory search. The problem of automated \textit{diversity-driven} discovery in these systems was recently introduced~\cite{Reinke2020Intrinsically, grizou2020curious}, highlighting that two key ingredients are autonomous exploration and unsupervised representation learning to describe \enquote{relevant} degrees of variations in the patterns. In this paper, we motivate the need for what we call \textit{Meta-diversity} search, arguing that there is not a unique ground truth \textit{interesting} diversity as it strongly depends on the final observer and its motives. Using a continuous game-of-life system for experiments, we provide empirical evidences that relying on monolithic architectures for the behavioral embedding design tends to bias the final discoveries (both for hand-defined and unsupervisedly-learned features) which are unlikely to be aligned with the interest of a final end-user. To address these issues, we introduce a novel \textit{dynamic} and \textit{modular} architecture that enables unsupervised learning of a hierarchy of diverse representations. Combined with intrinsically motivated goal exploration algorithms, we show that this system forms a discovery assistant that can efficiently adapt its diversity search towards preferences of a user using only a very small amount of user feedback.
\end{abstract}

\section{Introduction}
\label{sec:introduction}
\textit{Self-organisation} refers to a broad range of pattern-formation processes in which globally-coherent structures spontaneously emerge out of local interactions. In many natural and artificial systems, understanding this phenomenon poses fundamental challenges~\cite{ball1999self}.
In biology, \textit{morphogenesis}, where cellular populations self-organize into a structured morphology, is a prime example of complex developmental process where much remains to be assessed. Since Turing's influential paper \textit{\enquote{The Chemical Basis of Morphogenesis}}~\cite{turing1990chemical}, many mathematical and computational models of morphogenetic systems have been proposed and extensively studied~\cite{glen2019agent}. For example, cellular automata abstract models like Conway’s Game of Life (GoL), despite their apparent simplicity, generate a wide range of life-like structures ranging from Turing-like patterns reminiscent of animal skins stripes and spots~\cite{ball1999self} to localized autonomous patterns showing key behaviors like self-replication~\cite{langton1984self}.

A modern goal of morphogenesis research has become the manipulation, exploration and control of self-organizing systems. With the recent advances in synthetic biology and high-precision roboticized experimental conditions, potential applications range from automated material design~\cite{tabor2018accelerating}, drug discovery~\cite{schneider2019rethinking} and regenerative medicine~\cite{dzobo2018advances} to unraveling the chemical origins of life~\cite{grizou2020curious,meisner2019computational}. Yet, even for simple artificial systems like Conway's Game of Life, we do not fully grasp the mapping from local rules to global structures nor have a clear way to represent and classify the discovered structures. While scientific discoveries of novel patterns have primarily been relying on manual tuning and ad hoc exploratory search, the growing potential of data-driven search coupled with machine learning (ML) algorithms allows us to rethink our approach. In fact, contemporary work opened the path toward three promising directions in applying ML to morphogenesis research, namely in 1) the design of richer computational models, coupling neural networks data-structures~\cite{gilpin2019cellular} with deep-learning techniques to \textit{learn} the update rules of the system, allowing to directly achieve key properties such as self-regeneration and self-replication~\cite{mordvintsev2020growing}; 2) by formulating morphological search as a reinforcement learning problem where an artificial agent learns to control the morphologies self-assembly to achieve a certain task such as locomotion ~\cite{pathak2019learning}; and 3) formulating the problem of automated \textit{diversity-driven} discovery in morphogenetic systems, and proposing to transpose intrinsically-motivated exploration algorithms, coming from the field of developmental robotics, to this new kind of problem~\cite{Reinke2020Intrinsically, grizou2020curious}.

Pure objective-driven search is unlikely to scale to complex morphogenetic systems. 
It can even be inapplicable when scientists do not know how to characterize desired behaviours from raw observations (reward definition problem), or merely aim to find novel patterns.
As an illustration of these challenges, it took 40 years before the first replicator \enquote{spaceship} pattern was spotted in Conway's Game of Life.
We believe that \textit{diversity-driven} approaches can be powerful discovery tools~\cite{Reinke2020Intrinsically, grizou2020curious}, and can potentially be coupled with objective-driven searches~\cite{pathak2017curiosity,pugh2016quality,colas2018gep}. Recent families of machine learning have shown to be effective at creating \textit{behavioral diversity}, namely Novelty Search (NS)~\cite{lehman2008exploiting,lehman2011abandoning} and Quality-Diversity (QD)~\cite{pugh2016quality,cully2015robots} coming from the field of evolutionary robotics; and intrinsically-motivated goal-directed exploration processes (IMGEP)~\cite{baranes2013active, forestier2017intrinsically} coming from developmental robotics. A known critical part of these algorithms, is that they require the definition of a \textit{behavioral characterization} (BC) feature space which formalizes the \enquote{interesting} degrees of behavioral variation in the environment~\cite{pugh2015confronting}. So far, this behavior space has either been hand-defined in robotic scenarios~\cite{cully2015robots,forestier2017intrinsically} or unsupervisedly learned with deep auto-encoders directly from raw observations ~\cite{pere2018unsupervised,laversanne2018curiosity,nair2018visual,pong2019skew,cully2019autonomous}. While deep auto-encoders have shown to recover the \enquote{ground truth} factor of variations in simple generative datasets~\cite{burgess2018understanding}, it is impossible (and not desirable) to recover \textit{all} the degrees of variations in the targeted self-organizing systems. 

In this paper, we follow the proposed experimental testbed of Reinke et al.~(2020)~\cite{Reinke2020Intrinsically} on a continuous game-of-life system (Lenia,~\cite{chan2018lenia}). We provide empirical evidence that the discoveries of an IMGEP operating in a \textit{monolithic} BC space are highly-diverse in that space, yet tend to be poorly-diverse in other potentially-interesting BC spaces. This draws several limitations when it comes to applying such system as a tool for assisting discovery in morphogenetic system, as the suggested discoveries are unlikely to align with the interests of a end-user. How to build an artificial \enquote{discovery assistant} learning to generate diverse patterns in the eyes of a future, yet unknown, observer? A central challenge in that direction is to extend the standard notion of \textit{diversity}, where an agent discovers diverse patterns in a monolithic BC space, to what we call \textit{meta-diversity}, where an agent incrementally learns diverse behavioral characterization spaces and discovers diverse patterns within each of them. A second key challenge is to build exploration strategies that can quickly adapt to the preferences of a human end-user, while requiring minimal feedback. To address these challenges, we propose a novel model architecture for unsupervised representation learning with Hierarchically Organized Latent Modules for Exploratory Search, called \textit{HOLMES}. We compare the behavioral characterizations learned by an IMGEP equipped with HOLMES hierarchy of goal space representations to an IMGEP using a single monolithic goal space representation, and the resulting discoveries. We consider two end-user models respectively interested in two types of diversities (diverse spatially localized and diverse turing-like patterns), and show that a monolithic IMGEP will make discoveries that are strongly uneven in relation to these user preferences, while IMGEP-HOLMES is better suited to escape this bias by learning divergent feature representations. Additionally, we show how HOLMES can be efficiently \textit{guided} to drive the search toward those two types of \textit{interesting} diversities with very little amount of (simulated) user feedback. 

Our contributions are threefold. We introduce the novel objective of \textit{meta-diversity} search in the context of automated discovery in morphogenetic systems. We propose a \textit{dynamic} and \textit{modular} model architecture for unsupervised learning of \textit{diverse} representations, which, to our knowledge, is the first work that proposes to progressively grow the capacity of the agent visual world model into an organized hierarchical representation. We show how this architecture can easily be leveraged to \textit{drive} exploration, opening interesting perspectives for the integration of a human in the loop.

\section{Problem Formulation and Motivation for \textit{Meta-Diversity} Search}
\label{sec:problem_formulation}
We summarize the problem of automated discovery of morphogenetic systems as formulated in \cite{Reinke2020Intrinsically}, on a continuous game of life environment example. We identify limits of this formulation and associated approach. The novel process of \textit{meta-diversity} search is proposed within this framework.

\paragraph*{A morphogenetic system: Lenia} Morphogenetic systems are characterized by an \textit{initial state} ($A^{t=1}$, seed of the system) as well as a set of \textit{local update rules} that are iteratively applied to evolve the state of the system through time ($A^{t=1} \rightarrow \ldots \rightarrow A^{t=T}$). Typically observed from raw images, the emerging patterns depend on a set of \textit{controllable parameters} $\theta$ that, for each experimental run, condition the system initial state and update rules. We use Lenia cellular automaton~\cite{chan2018lenia,chan2020lenia}, a continuous generalization to Conway’s Game of Life, as testbed. It can generate a wide range of complex behaviors and life-like structures, as testified by the collection of \enquote{species} that were manually identified and classified in \cite{chan2018lenia}. To organize the experimental study of exploration algorithms, several restrictions were proposed \cite{Reinke2020Intrinsically}: the lattice resolution is fixed to $256\times256$ and the evolution is stopped after 200 time steps. Moreover, only the system final state is observed ($A^{t=200}$), focusing the search on morphological appearance traits and leaving out dynamical traits (yet see section~\ref{sec:experimental_results} for side-effect discoveries of interesting dynamics). 
The controllable parameters include a 7-dimensional set of parameters controlling Lenia's update rule as well as parameters governing the generation of the initial state $A^{t=1}$. Compositional-pattern producing networks (CPPN)~\cite{stanley2006exploiting} are used to produce the initial patterns. 

\paragraph*{Automated discovery problem} The standard automated discovery problem (\cite{Reinke2020Intrinsically}) consists of generating a maximally-diverse set of observations through sequential experimentation of controllable parameters $\theta$. Each controllable parameter vector $\theta$ generates a rollout $A^{t=1} \rightarrow \ldots \rightarrow A^{t=T}$ of the morphogenetic system leading to an observation $o(\theta)=A^{t=T}$. This observation is encoded as a vector $r=R(o)$ in a BC space representing interesting features of the observation (e.g. based on the color or frequency content of the observation image). Given a budget of N experiments, the objective of the automated discovery problem is to sample a set of controllable parameters $\Theta$ where $\{R(o(\theta)) | \theta \in \Theta\}$ maximally covers the BC space.

\paragraph*{Problem definition: Meta-Diversity Search} The standard automated discovery problem defined above assumes that the intuitive notion of diversity can be captured within a single BC space (what we call a monolithic representation). However, as our results will show, maximizing the diversity in one BC space may lead to poor diversity in other, possibly-interesting, BC spaces. Thus, using a single representation space to drive exploration algorithms limits the scope of their discoveries, as well as the scope of their external evaluation. To address this limit, we formulate the novel process of \textit{meta-diversity} search: in an outer loop, one aims to learn a \textit{diverse} set of behavioral characterizations (called the \textit{meta-diversity}); then in an inner loop, one aims to discover a set of maximally diverse patterns in each of the BC spaces (corresponding to the standard notion of \textit{diversity} in previous work). The objective of this process is to enable continuous seeking of novel niches of diversities while being able to quickly adapt the search toward a new unknown \textit{interesting} diversity. Here, a successful discovery assistant agent is one which can leverage its diverse BCs to specialize efficiently towards a particular type of diversity, corresponding to the initially unknown preferences of an end-user, and expressed through simple and very sparse feedback.

\paragraph*{Proposed approach: IMGEP Agent with modular BC spaces} A goal-directed intrinsically-motivated exploration process (IMGEP) was used for the parameter sampling strategy in~\cite{Reinke2020Intrinsically}. After an initialization phase, the IMGEP strategy iterates through 1) sampling a goal in a learned BC space $R$, conditioned on the explicit memory of the system $\mathcal{H}$ and based on the goal-sampling strategy $g \sim G(H)$; 2) sampling a set of parameters $\theta$ for the next rollout to achieve that goal, based on its parameter-sampling policy $\Pi=Pr(\theta;g,H)$; 3) let the system rollout, observe outcome $o$, and store the resulting $(\theta, o, R(o))$ triplet in an explicit memory $\mathcal{H}$. However, the behavioral characterization was based on a monolithic representation. Although being learned, this representation was limited to capture diversity in a single BC space and was therefore unable to perform meta-diversity search as defined above. To solve this problem, we introduce a modular architecture where a hierarchy of behavioral characterization spaces is progressively constructed, allowing flexible representations and intuitive guidance during the discovery process. 

\section{Hierarchically Organized Latent Modules for Exploratory Search}
\label{sec:HOLMES}



\begin{figure}[t]
\includegraphics[width=\textwidth]{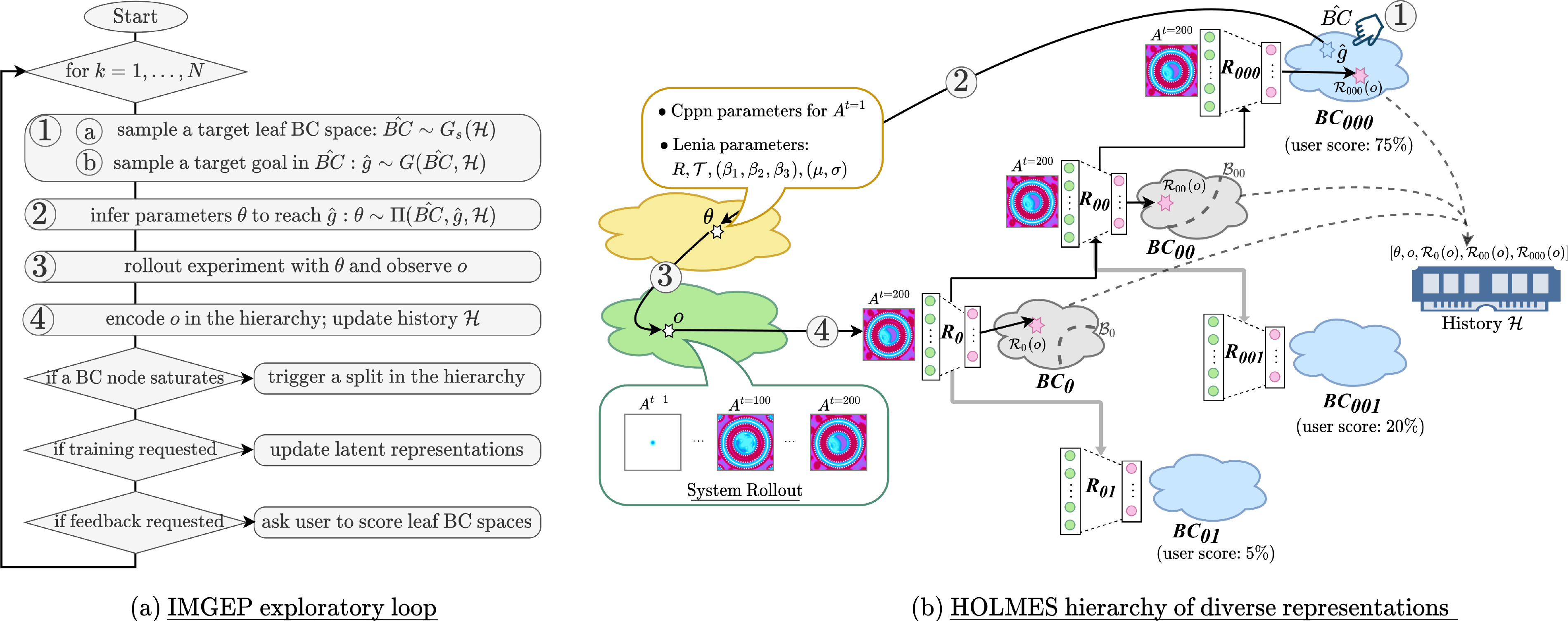}
\caption{IMGEP-HOLMES framework integrates a goal-based intrinsically-motivated exploration process (IMGEP) with the incremental learning of a hierarchy of behavioral characterization spaces (HOLMES). HOLMES unsupervisedly clusters and encodes discovered patterns into the different nodes of the hierarchy of representations. The architecture of HOLMES and clustering part is detailed in section ~\ref{subsec:HOLMES_architecture}. The exploratory loop and its interaction with the hierarchy of behavioral characterization (BC) spaces, enabling \textit{meta-diversity search}, is detailed in section ~\ref{subsec:IMGEP-HOLMES}. 
}
\vspace{-3pt}
\label{fig:IMGEP-HOLMES}
\end{figure}

\subsection{HOLMES: Architecture}
\label{subsec:HOLMES_architecture}

HOLMES is a dynamic architecture that \enquote{starts small} both on the task data distribution and on the network memory capacity, following the intuition of Elman~(1993)~\cite{elman1993learning}. A hierarchy of embedding networks $\mathcal{R}=\{\mathcal{R}_i\}$ is actively expanded to accommodate to the different niches of patterns discovered in the environment, resulting in a progressively deeper hierarchy of specialized BC spaces. The architecture has 4 main components: (i) a base \textit{module} embedding neural network, (ii) a \textit{saturation} signal that triggers the instantiating of new nodes in the hierarchy, (iii) a \textit{boundary} criteria that unsupervisedly clusters the incoming patterns into the different modules; and (iv) a \textit{connection-scheme} that allows feature-wise transfer from a parent module to its children. \\
We refer to Figure~\ref{fig:IMGEP-HOLMES} and Figure~\ref{sm:fig:HOLMES_focus} (section~\ref{sm:sec:design_choices_IMGEP-HOLMES}, appendix) for an illustration of the following.

In this paper, we use a variational auto-encoder (VAE)~\cite{kingma2013auto} as base \textit{module}. The hierarchy starts with a single VAE $\mathcal{R}_0$ that is incrementally trained on the incoming data, encoding it into its latent characterization space $BC_0$. A split is triggered when a node of the hierarchy \textit{saturates} i.e. when the reconstruction loss of its VAE reaches a plateau, with additional conditions to prevent premature splitting (minimal node population and a minimal number of training steps). Each time a node gets \textit{saturated}, the split procedure is executed as follows. First, the parameters of the parent $\mathcal{R}_p$ are frozen and two child networks $\mathcal{R}_{p0}$ and $\mathcal{R}_{p1}$ are instantiated with their own neural capacity. Besides, additional learnable layers called \textit{lateral connections} are instantiated between the parent and child VAE feature-maps, drawing inspiration from \textit{Progressive Neural Networks} (PNN)~\cite{rusu2016progressive}. Here, these layers allow the child VAE to reuse its parent knowledge while learning to characterize novel dissimilar features in its BC space (see section~\ref{subsec:results_incrementalBC} and appendix~\ref{sm:subsec:impact_lateral_connections} for an ablation study). Finally, a boundary $\mathcal{B}_p$ is fitted in the parent frozen embedding space $BC_p$ and kept fixed for the rest of the exploration. In this paper, the boundary is unsupervisedly fitted with K-means by assigning two clusters from the points that are currently in the node's BC space. Based on this boundary criteria, incoming observations forwarding through $\mathcal{R}_p$ will be either redirected through the left child $\mathcal{R}_{p0}$ or thought the right child $\mathcal{R}_{p1}$. After the split, training continues: leaf node VAEs as well as their incoming lateral connections are incrementally trained on their own niches of patterns while non-leaf nodes are frozen and serve to redirect the incoming patterns.

In HOLMES, the clustering of the different patterns is central to learn \textit{diverse} BC spaces. The clustering is influenced by the choice of the module and connections training strategy, that determines the latent distribution of patterns in the latent space, and by the clustering algorithm itself. Note that the genericity of HOLMES architecture (agnostic to the base module) allows many other design choices to be considered in future work, and that current choices are discussed in Appendix~\ref{sm:subsec:HOLMES}. 

\subsection{IMGEP-HOLMES: Interaction with the Exploration Process}
\label{subsec:IMGEP-HOLMES}
The goal space of an IMGEP is usually defined as the BC space of interest, with a representation based on a monolithic architecture $\mathcal{R}$~\cite{pere2018unsupervised,laversanne2018curiosity}. In this paper, we propose a variant where the IMGEP operates in a hierarchy of goal spaces $\{BC_i\}$, where observations and hence goals are encoded at different levels or granularity, as defined by HOLMES embedding hierarchy $\{\mathcal{R}_i\}$. The exploration process iterates N times through steps 1-to-4, as illustrated in Figure~\ref{fig:IMGEP-HOLMES}. In this section we detail the implementation choices for each step. We refer to algorithm~\ref{sm:algo:IMGEP-HOLMES} in appendix for a general pseudo-code.

\textbf{1)} The goal-sampling strategy is divided in two sub-steps. \textbf{a)} Sample a target BC space $\hat{BC_i}$ with a goal space sampling distribution $G_s$. Here, the agent is given an additional degree of control allowing to prioritize exploration in certain nodes of the hierarchy (and therefore on a subset population of patterns). In this paper we considered two setups for the goal space sampling strategy: (i) a \textit{non-guided} variant where the target BC space is sampled uniformly over all the leaf nodes and (ii) a \textit{guided} variant where, after each split in the hierarchy we \enquote{pause} exploration and ask for evaluator feedback to assign an \textit{interest score} to the different leaf modules. The guided variant simulates a human observer that could, by visually browsing at few representative images per module and simply "click" and score the leaf nodes with the preferred discoveries. Then during exploration, the agent selects over the leaf goal spaces with softmax sampling on the assigned score probabilities. \textbf{b)} Sample a target goal $\hat{g}$ in the selected BC space with a goal sampling distribution $G$. In this paper, we use a uniform sampling strategy: the goal is uniformly sampled in the hypercube envelope of currently-reached goals. Because the volume of the hypercube is larger than the cloud of currently-reached goals, uniform sampling in the hypercube incentivizes to search in unexplored area outside this cloud (this is equivalent to novelty search in the selected BC space). Note that other goal-sampling mechanisms can be considered within the IMGEP framework~\cite{baranes2013active}.

\textbf{2)} The parameter-sampling strategy $\Pi$ generates the CPPN-generated initial state and Lenia's update rule in 2 steps: (i) given the goal $\hat{g} \in \hat{BC_i}$, select parameters $\hat{\theta}$ in $\mathcal{H}$ whose corresponding outcome $R_i(o)$ is closest to $\hat{g}$ in $\hat{BC_i}$; (ii) mutate the selected parameters by a random process $\theta = \textsc{mutation}(\hat{\theta})$ (see appendix~\ref{sm:subsec:parameter_sampling} for implementation details).

\textbf{3)} Rollout experiment with parameters $\theta$ and observe outcome $o$.  

\textbf{4)} Forward $o$ top-down through the hierarchy and retrieve respective embeddings $\{R_i(o)\}$ along the path. Append respective triplets $\{(\theta, o, R_i(o))\}$ to the episodic memory $\mathcal{H}$.

\paragraph*{HOLMES online training.} The data distribution collected by the IMGEP agent directly influences HOLMES splitting procedure and training procedure by determining which nodes get populated and when. In this paper, we incrementally train the goal space hierarchy every $N_T=100$ exploration step for $E=100$ epochs on the observations collected in the history $\mathcal{H}$. Importance sampling is used at each training stage, giving more weight to recently discovered patterns. We refer to appendix~\ref{sm:subsec:imgep_variants} for implementation details on HOLMES training procedure.

\section{Experimental Results}
\label{sec:experimental_results}

In this section we compare the results of an IMGEP equipped with a goal space based on different types of BCs. We denote \textbf{IMGEP-X} an IMGEP operating in a goal space $X$ ($X$ can be e.g. an analytical BC space based on Fourier descriptors, or a modular representation learned by the HOLMES architecture as in section~\ref{subsec:IMGEP-HOLMES}). 
In order to evaluate meta-diversity, we make the distinction between the BC used as the goal space of an IMGEP and the BC used for evaluating the level of diversity reached by that IMGEP. For example, we might want to evaluate the diversity reached by \textbf{IMGEP-HOLMES} in a BC space based on Fourier descriptors. For quantitative evaluation of the diversity in a given BC, the BC space is discretized with $n=20$ bins per dimension, and the diversity is measured as the number of bins in which at least one explored entity falls (details are provided in Appendix ~\ref{sm:subsubsec:diversity_measure}). Each experiment below consists of $N=5000$ runs starting with $N_{init} = 1000$ initial random explorations runs. For all algorithms, we conduct 10 repetitions of the exploration experiment with different random seeds. Please refer to appendix (sections~\ref{sm:sec:evaluation_procedure} and~\ref{sm:sec:experimental_settings}) for all details on the evaluation procedure and experimental settings. Additionally, the source code and full database of discoveries are provided on the project website.


\newpage
\subsection{Does maximizing diversity in a given BC space lead to high diversity in other BC spaces?}
\label{subsec:results_predefinedBC}
\begin{figure}[t!]
\centering
\setlength\tabcolsep{1pt}
\renewcommand{\arraystretch}{0.2} 
    \begin{tabular}{ccl}
    \multirow{5}{*}[5mm]{
    \vspace{-5mm}
    \includegraphics[height=4.5cm]{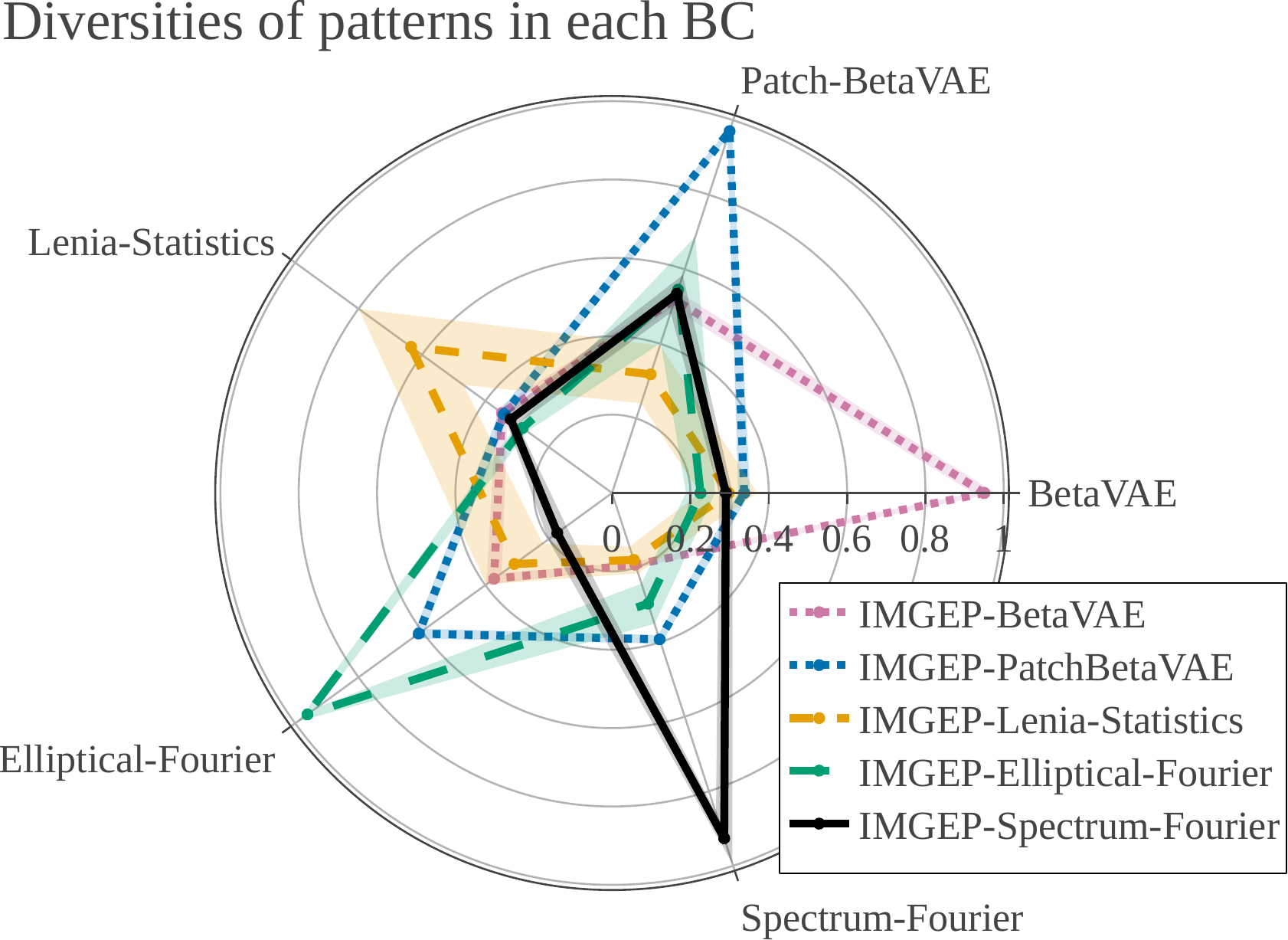}
    } &
        \begin{tabular}{c}
            \tiny{Diverse in} \\
            \tiny{\textsc{Spectrum-Fourier}}:
        \end{tabular}&
        \begin{tabular}{l}
            \includegraphics[height=.91cm]{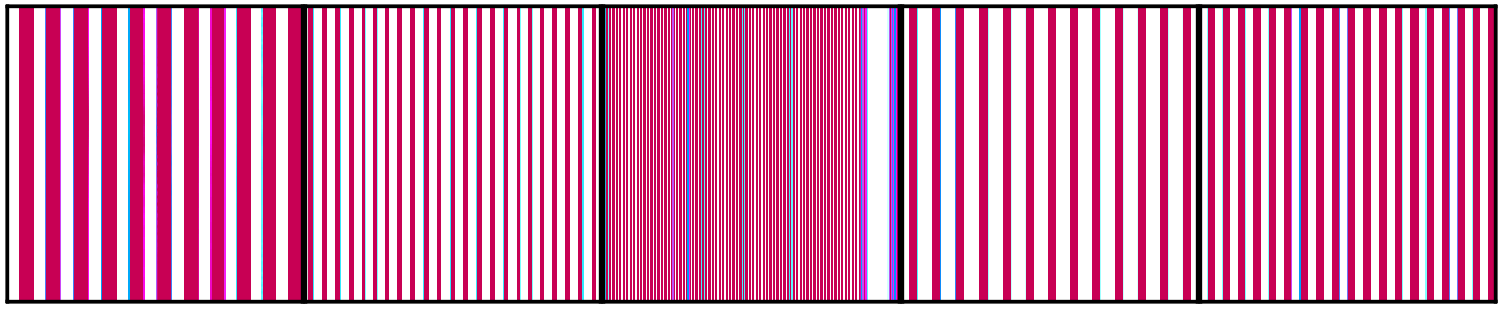}
        \end{tabular}\\[1.25ex]
        &
        \begin{tabular}{c}
            \tiny{Diverse in} \\
            \tiny{\textsc{Elliptical-Fourier}:}
        \end{tabular}&
        \begin{tabular}{l}
            \includegraphics[height=.902cm]{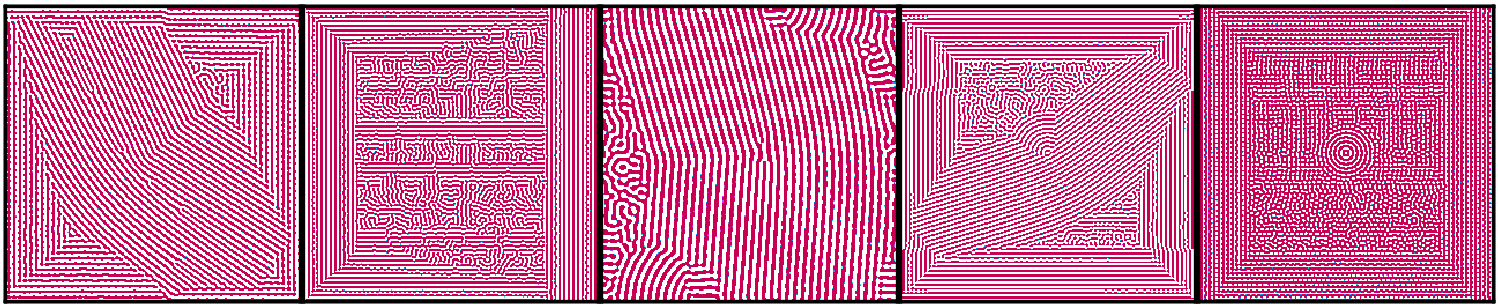}
        \end{tabular}\\[1.25ex]
        &
        \begin{tabular}{c}
            \tiny{Diverse in} \\
            \tiny{\textsc{Lenia-Statistics}:}
        \end{tabular}&
        \begin{tabular}{l}
            \includegraphics[height=.91cm]{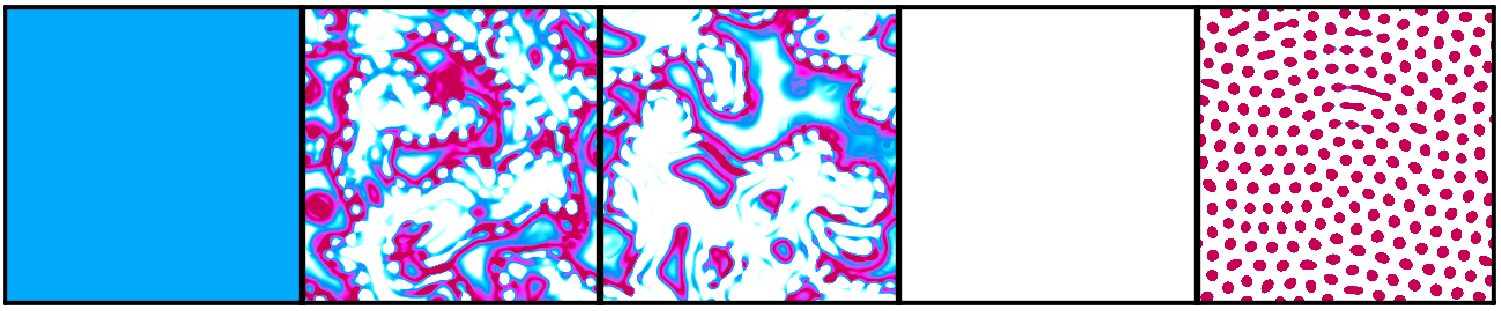}
        \end{tabular}\\[1.25ex]
        &
        \begin{tabular}{c}
            \tiny{Diverse in} \\
            \tiny{\textsc{BetaVAE}:}
        \end{tabular}&
        \begin{tabular}{l}
            \includegraphics[height=.9cm]{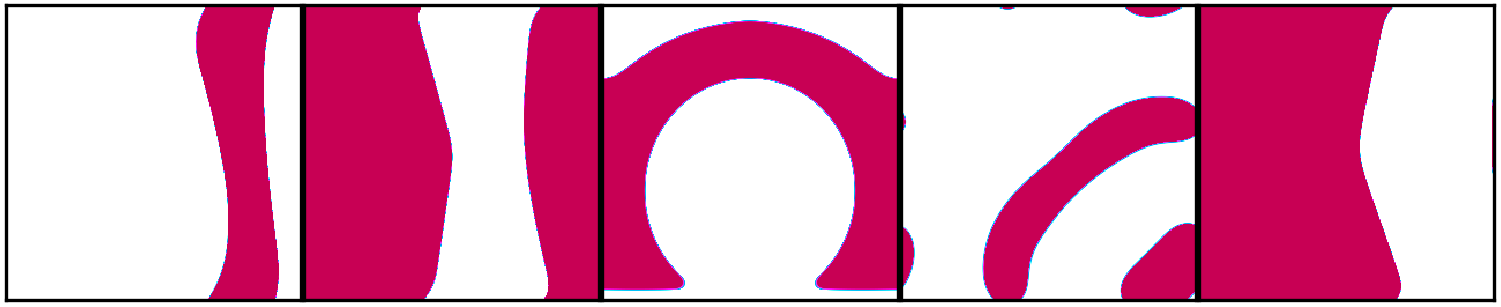}
        \end{tabular}\\[1.25ex]
        &
        \begin{tabular}{c}
            \tiny{Diverse in} \\
            \tiny{\textsc{Patch-BetaVAE}:}
        \end{tabular}&
        \begin{tabular}{l}
            \includegraphics[height=.9cm]{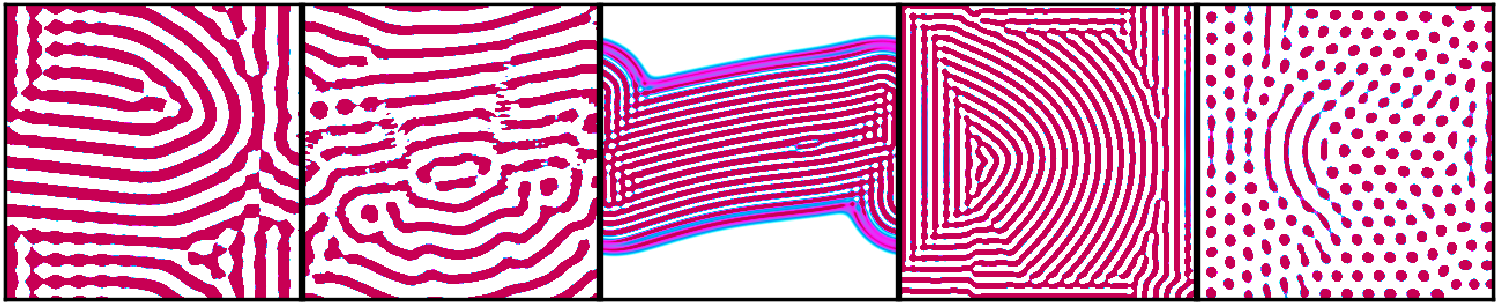}
        \end{tabular}\\[5.25ex]
     \end{tabular}
     \vspace{-15pt}
 \caption{Although IMGEPs succeed to reach a high-diversity in their respective BC space, they are poorly-diverse in all the others. (left) Diversity for all IMGEP variants measured in each analytic BC space. For better visualisation the resulting diversities are divided by the maximum along each axis. Mean and std-deviation shaded area curves are depicted. (right). Examples of patterns discovered by the IMGEPs that are consider diverse in their respective BC space. See Appendix~\ref{sm:subsubsec:analytic_BCs} for details.}
\label{fig:diversity_per_BC}
\end{figure}

We construct 5 BC spaces with 8 dimensions each, among which 3 representations are predefined and 2 are pre-learned on a database of Lenia patterns. The exploration discoveries of an IMGEP equipped with the different BCs as (fixed) goal space are evaluated and compared in figure~\ref{fig:diversity_per_BC}. The two first BCs rely on Fourier descriptors, intended to characterize the frequencies of textures (Spectrum-Fourier) and shape of closed contours (Elliptical-Fourier~\cite{kuhl1982elliptic}), typically used in cellular-automata~\cite{machicao2018cellular} and biology~\cite{cope2012plant,zhang2019crop}. The third BC relies on statistical features that were proposed in the original Lenia paper~\cite{chan2018lenia} to describe the activation of patterns. The fourth uses a set of features unsupervisedly learned by a $\beta$-VAE~\cite{burgess2018understanding} on a large database of Lenia patterns, as proposed in~\cite{Reinke2020Intrinsically}. Because $\beta$-VAE can poorly represent high-frequency patterns, another variant trained on cropped patches is proposed. 

\paragraph*{Limits of monolithic goal spaces} The results of Figure~\ref{fig:diversity_per_BC} suggest that, if we could have a theoretical BC model that aligns with what a user considers as diverse under the form of a goal space, the IMGEP should be efficient in finding a diversity of patterns in that space. In practice however, constructing such a BC is very challenging, if not infeasible. Each BC was carefully designed or unsupervisedly learned to represent what could be \enquote{relevant} factors of variations and yet, the IMGEP seems to exploit degrees of variations that might not be aligned with what we had in mind when constructing such BCs. Spectrum-Fourier is a clear example that was constructed to describe textures (in a general sense) but where the discoveries exhibit only vertical-stripe patterns with all kind of frequencies.

\subsection{What is the impact of modularity in incremental learning of goal space(s)?}
\label{subsec:results_incrementalBC}

\paragraph*{Baselines} We compare \textbf{IMGEP-VAE} which uses a monolithic VAE as goal space representation to \textbf{IMGEP-HOLMES} which is defined in section~\ref{sec:HOLMES}. HOLMES expansion is stopped after 11 splits (resulting in a hierarchy of 23 VAEs) and uses small-capacity modules and connections, such that its final capacity is still smaller than the monolithic VAE. Other variants for the monolithic architecture and training strategies are considered in  Appendix~\ref{sm:subsec:monolithic_baselines}.

\begin{figure}[h!]
\sidebysidecaption{0.555\linewidth}{0.35\linewidth}{%
    \centering
    \setlength\tabcolsep{1pt}
    \renewcommand{\arraystretch}{0.5}
    \def\imagetop#1{\vtop{\null\hbox{#1}}}
    \begin{tabular}[t]{cc}
VAE &  HOLMES\\
 \midrule
    \imagetop{\includegraphics[height=3.7cm]{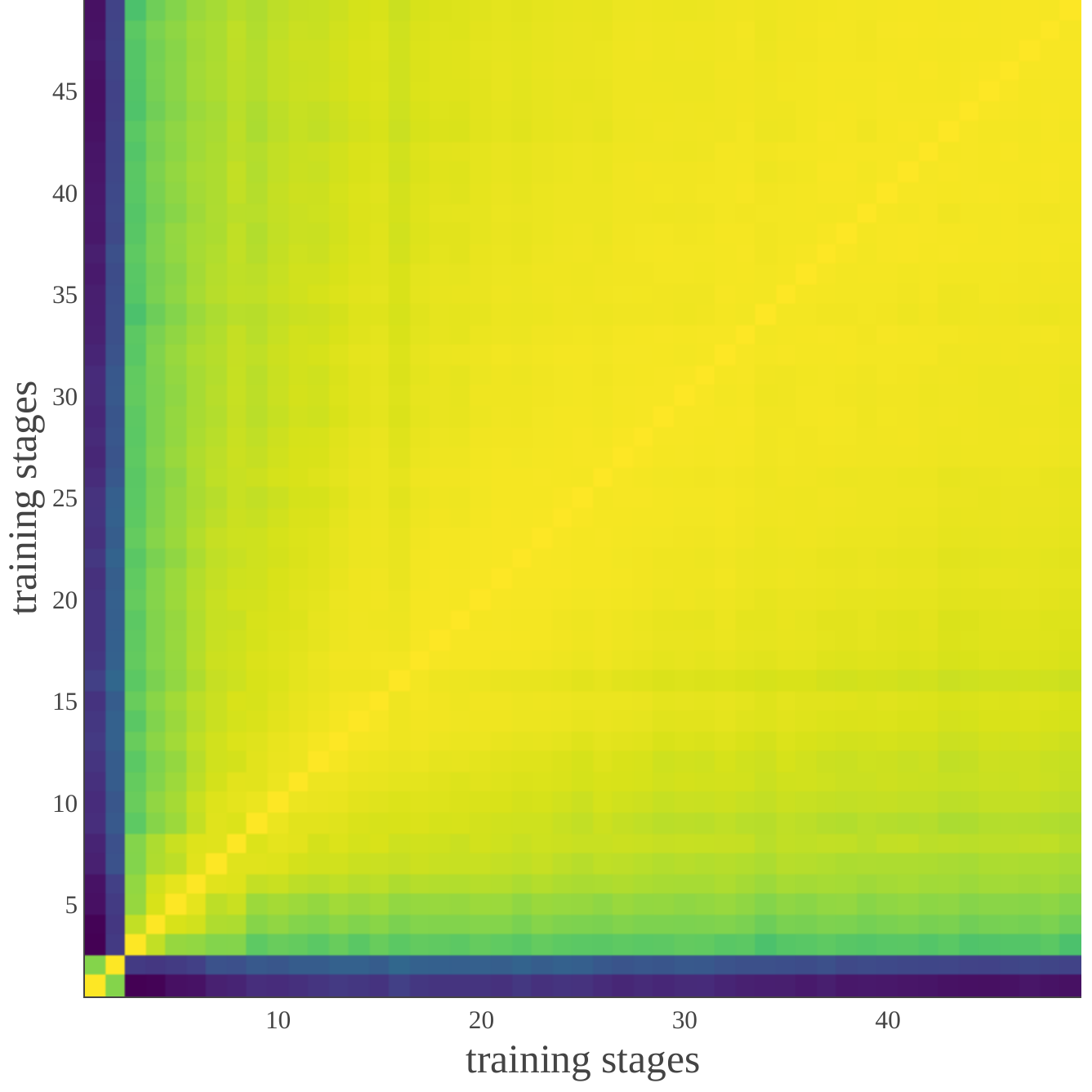}}
    & 
    \imagetop{\includegraphics[height=4.2cm]{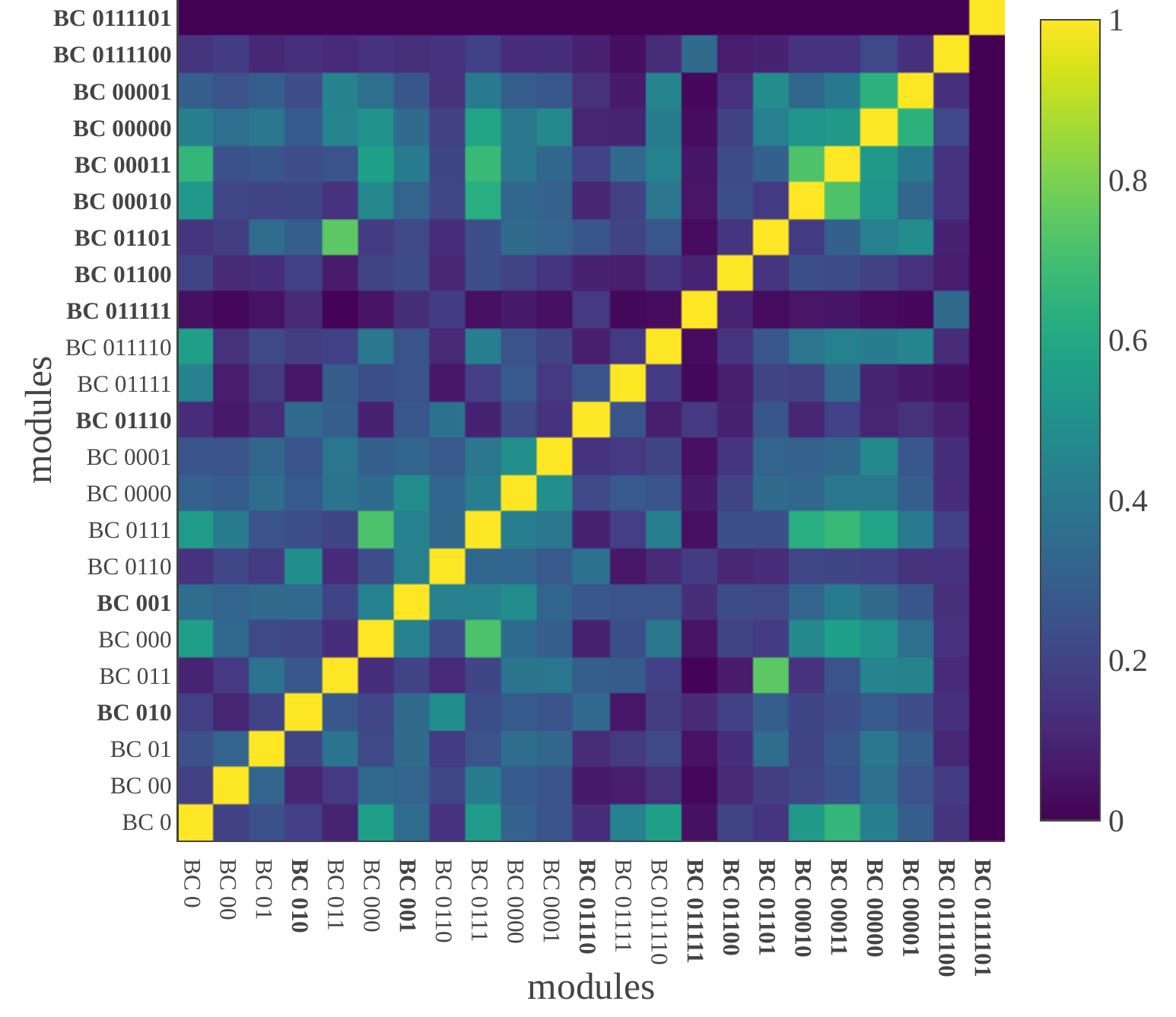}}
    \end{tabular}%
}{%
  \caption{RSA similarity index between 0 (dark blue, not similar at all) and 1 (yellow, identical). (VAE) representations are compared in time between the different training stages. (HOLMES) representations are compared at the \textit{end} of exploration between the different modules. Leafs are depicted in bold and modules are ordered by their creation time (left-to-right in x-axis). See Appendix ~\ref{sm:subsec:RSA_results} for full temporal analysis and statistical results. 
  }
\label{fig:CKA}
}
\vspace{-15pt}
\end{figure}

\paragraph*{Learning to characterize different niches} We use representational similarity analysis (RSA)~\cite{kriegeskorte2008representational} to quantify how much the representations embeddings (encoded behaviors) evolve through the exploration inner loop (Figure~\ref{fig:CKA}). Different metrics have been proposed to compute the representational similarity, here we use the linear centered kernel alignment (CKA) index~\cite{kornblith2019similarity} (see Appendix~\ref{sm:subsec:RSA_evaluation}). Results show that the learned features of the monolithic VAE stop evolving after only 2 training stages, i.e. 200 explored patterns. This suggests that, even though the VAE is \textit{incrementally trained} during exploration (at the difference of the pretrained variants in section~\ref{subsec:results_predefinedBC}), it still fails to adapt to the different niches of patterns which will lead to limited discoveries. However, RSA results suggest that HOLMES succeeds to learn features that are highly dissimilar from one module to another, which allows to target discovery of a \textit{meta-diversity}. An ablation study (see section ~\ref{sm:subsec:impact_lateral_connections} of appendix) highlighted the importance of HOLMES \textit{lateral connections} to escape the VAE learning biases.

\begin{figure}[b]
\centering
\setlength\tabcolsep{1pt}
\begin{tabular}{@{}cc@{}}
Spatially-Localized Patterns (SLP) & Turing-like Patterns (TLP) \\
 \includegraphics[width=0.49\textwidth]{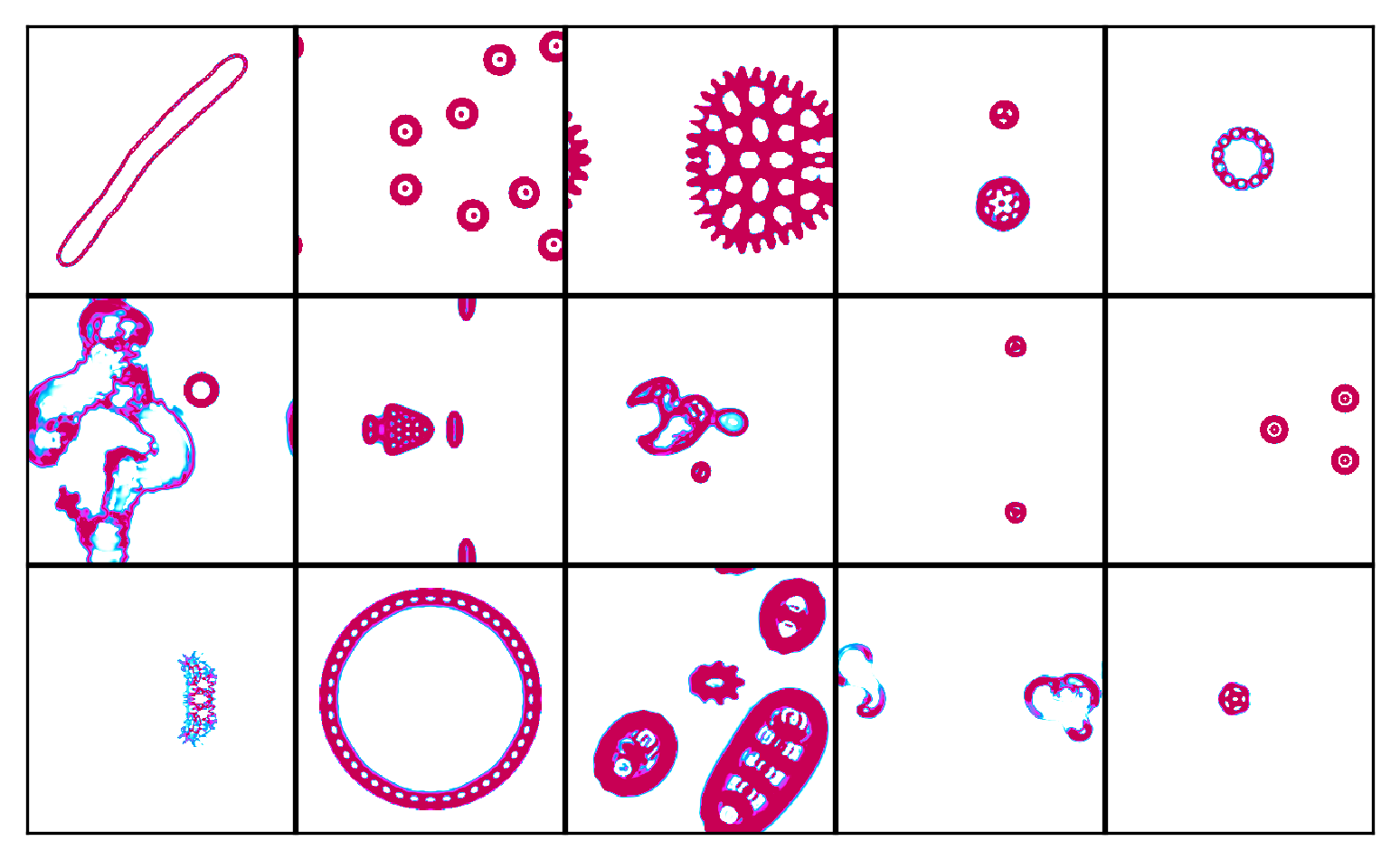} 
& 
 \includegraphics[width=0.49\textwidth]{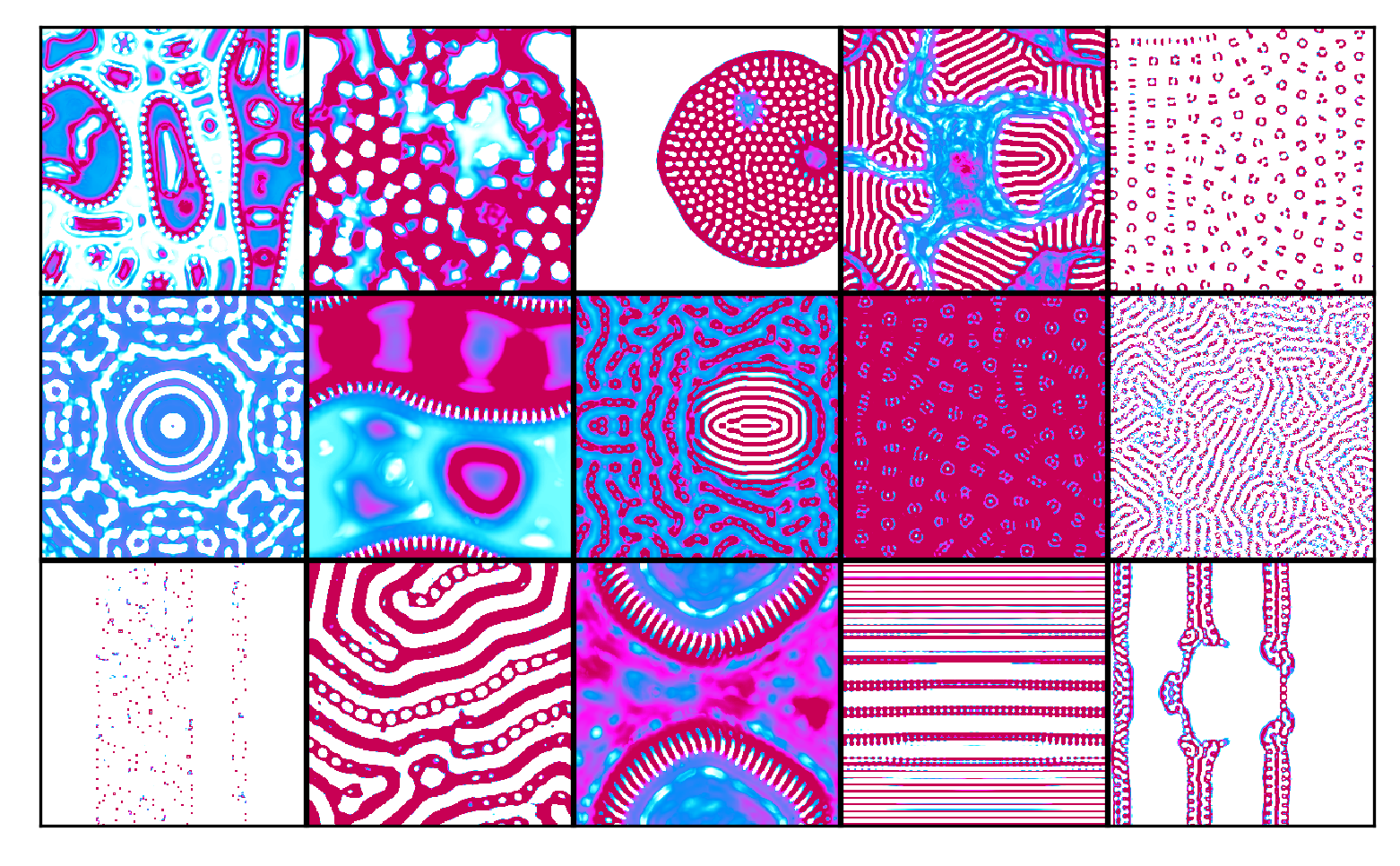}
 \end{tabular}
 \vspace{-10pt}
 \caption{(Left) SLP are autonomous stable patterns, that show interesting behaviors such as locomotion and metamorphosis(shape-shifting). (Right) TLP patterns are characterized by an unlimited spatial growth resembling reaction-diffusion pattern-formation of fronts, spirals, stripes and dissipative solitons. The displayed patterns where \textit{autonomously} discovered in Lenia~\cite{chan2018lenia} by IMGEP-HOLMES (without guidance) and, considered by us (human evaluator), as \textit{interesting}.}
\label{fig:SLP_TLP_examples}
\end{figure}

\paragraph*{Learning to explore different niches} Qualitative browsing through the discoveries confirms that IMGEP-HOLMES is able to explore diverse niches, allowing the discovery of very interesting behaviors in Lenia from quite unexpected types of patterns. To further investigate these discoveries, the exploration was prolonged for 10000 additional runs (without expanding more the hierarchy). Not only many of the \enquote{lifeform} patterns of the \textit{species} identified in~\cite{chan2018lenia} were discovered, but it allowed to assist to  the \textit{birth} of these creatures from fluid-like structures that have shown capable of \textit{pattern-emission} (behavior which was, to our knowledge, never discovered in 2D Lenia). Example of our discoveries can be seen in Figure~\ref{fig:SLP_TLP_examples} and on the website  { \small \url{http://mayalenE.github.io/holmes/}}. 

\subsection{Can we drive the search toward an \textit{interesting} diversity?}
\label{subsec:guided_discovery}

Two categories of patterns have been extensively studied in cellular-automata, known as Spatially-Localized Patterns (SLP) and Turing-Like Patterns (TLP) (Figure~\ref{fig:SLP_TLP_examples}). 
We investigate if our \textit{discovery assistant} search can be guided to specialize toward a diversity of either SLPs or TLPs.  
For experimentation, we propose to use a \textit{simulated} user, preferring either SLPs or TLPs, and a \textit{proxy} evaluation of diversity tailored to these preferences. For \textit{guidance}, the classifiers of \enquote{animals} (SLP) and \enquote{non-animals} (TLP) from \cite{Reinke2020Intrinsically} are used to score the different nodes in IMGEP-HOLMES with the number of SLP (or TLP) that fall in that node at split time (see section ~\ref{subsec:IMGEP-HOLMES}). 
This simulates preferences of a user toward either SLPs or TLPs, who would use a GUI to score the patterns found in each leaves of IMGEP-HOLMES. The total number of user interventions is 11 (one per split) with, for each intervention, an average of 6 \enquote{clicks} (scores), which represents very sparse feedback.
For evaluation, an experiment with a human evaluator has been conducted for selecting the BC (among the 5 proposed in section ~\ref{subsec:results_predefinedBC}) that correlates the most with the evaluator judgement of what represents a diversity of SLP and a  diversity of TLP. \textsc{BC\textsubscript{Elliptical-Fourier}} was designed as the best proxy space for evaluating the diversity of SLP (98\% agreement score) and \textsc{BC\textsubscript{Lenia-Statistics}} was designated for TLP (92\% agreement). Experiment details are provided in appendix~\ref{sm:subsec:human_evaluation}.

\begin{figure}[t]
\centering
\setlength\tabcolsep{1pt}
\begin{tabular}{@{}cc@{}}
(a) Diversity of SLP in \textsc{bc\textsubscript{Elliptical-Fourier}} & (b) Diversity of TLP in \textsc{bc\textsubscript{Lenia-Statistics}}\\
 \includegraphics[width=0.45\textwidth]{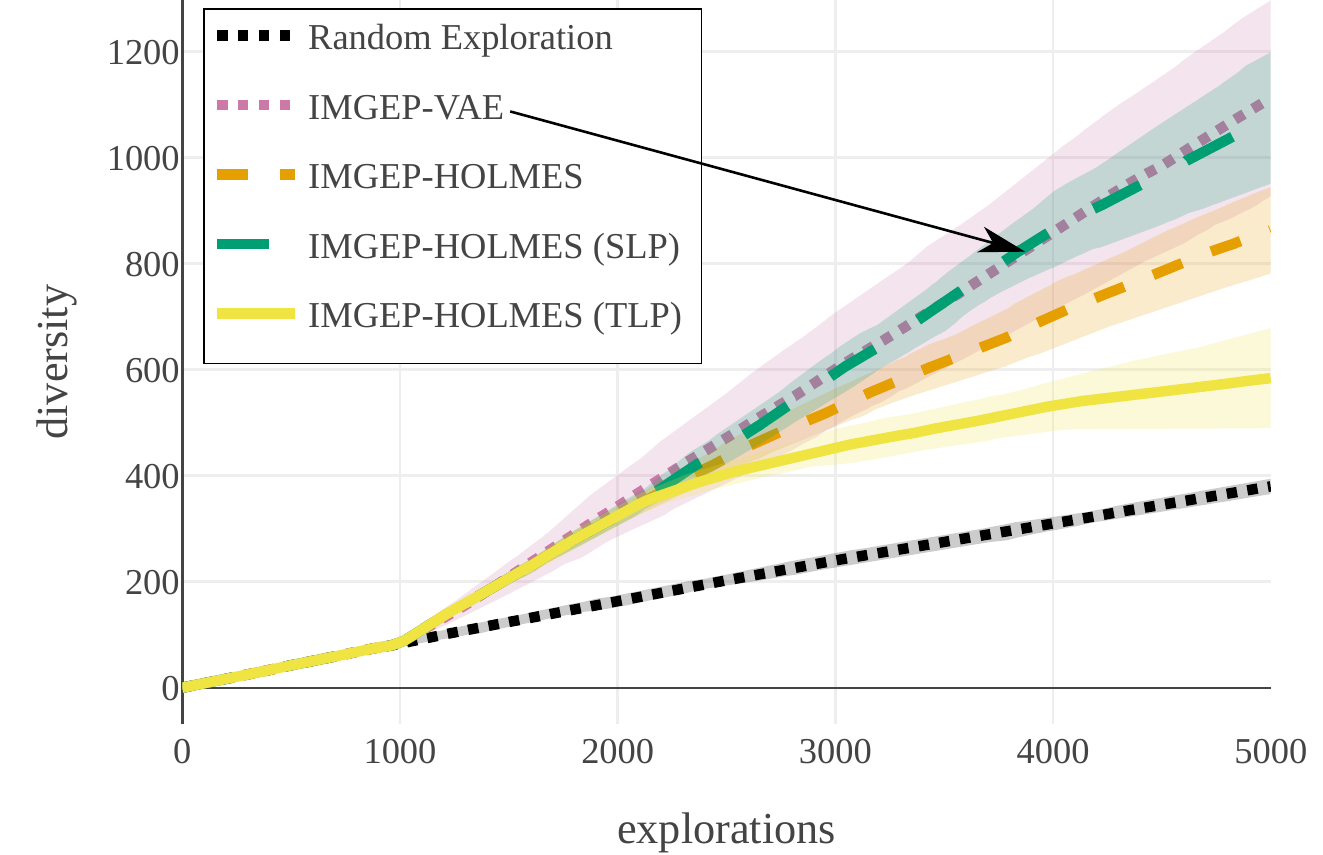} 
& 
 \includegraphics[width=0.45\textwidth]{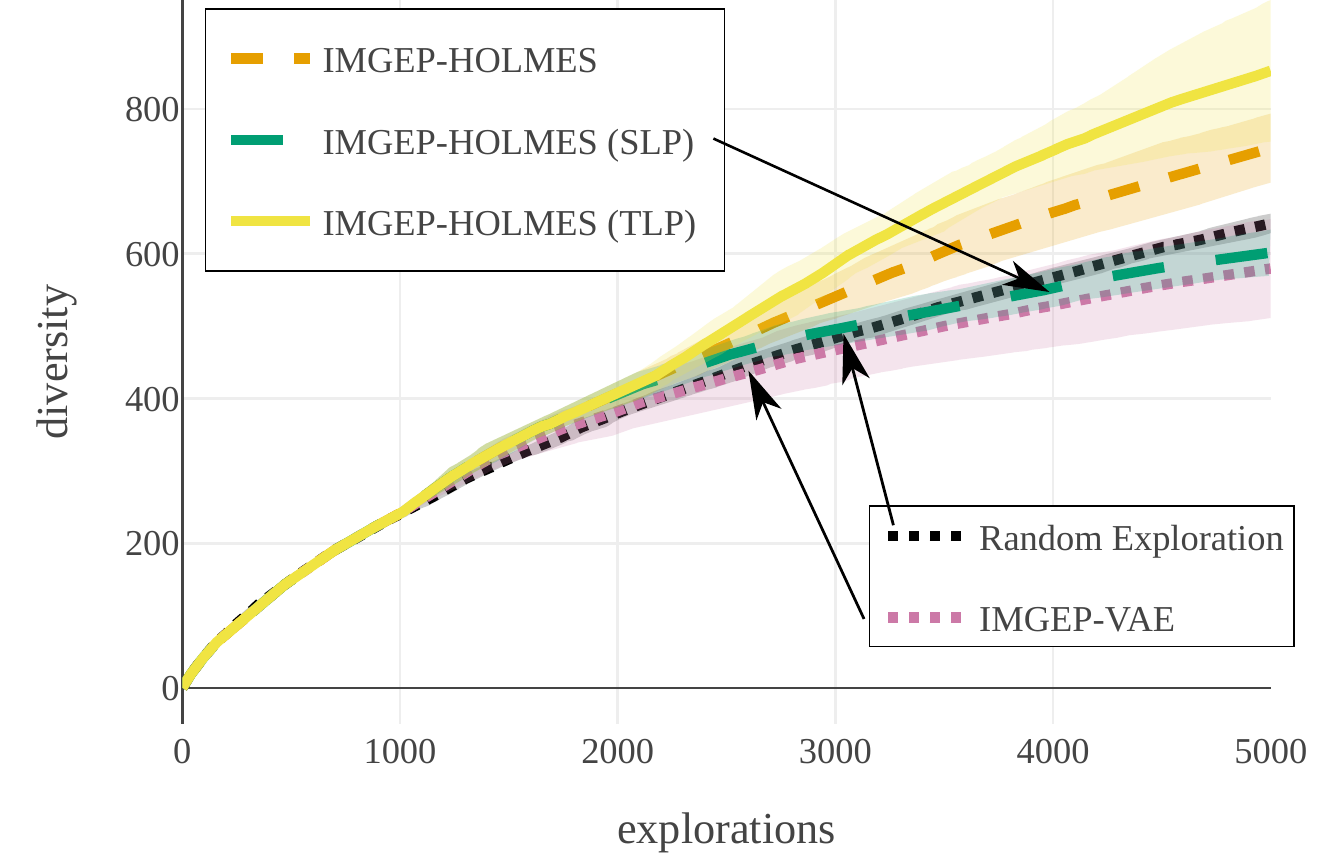}
 \end{tabular}
 \vspace{-5pt}
 \caption{Depicted is the diversity discovered by the algorithms during exploration. The discovered patterns classified as  SLP in (a) (resp TLP in (b)) are projected in \textsc{bc\textsubscript{Elliptical-Fourier}} space in (a)  (resp \textsc{bc\textsubscript{Lenia-Statistics}} in (b)), where the binning-based measure is used. Mean and std-deviation shaded area curves are depicted.  }
\label{fig:diversity_interesting}
\vspace{-5pt}
\end{figure}

\newpage
\paragraph*{Baselines} Non-guided IMGEP variants (section~\ref{subsec:results_incrementalBC}) are compared with guided IMGEP variants. The results are compared to Random Exploration baseline (where parameters $\theta$ are randomly sampled for the 5000 explorations) which serves as reference for the \textit{default} diversity found in Lenia (and by all algorithms during the first 1000 explorations). 

\paragraph*{Results} The results in Figure~\ref{fig:diversity_interesting} show that the bias of the monolithic VAE allows IMGEP-VAE to find a high diversity of SLPs but leads to poor diversity of TLPs. When non-guided, IMGEP-HOLMES finds a higher diversity than Random Exploration both for SLPs and TLPs. When guided, IMGEP-HOLMES can even further increase the diversity in the category of interest.

\section{Related Work}

\paragraph{Machine Learning for Automated Discovery in Science} Many work successfully applied ML for discovery in chemistry~\cite{duros2017human,raccuglia2016machine,gomez2018automatic}, physics~\cite{bloom2012automating,lu2018accelerated} and biology~\cite{king2004functional,hackett2020learning}. However, they focus the search on predefined structures with target properties and not, to our knowledge, on diversity search. 

\paragraph*{Diversity Search} Diversity-driven approaches in ML are presented in the introduction. In the QD literature, the work of Pugh et al.~(2016)~\cite{pugh2016searching} relates to what we frame as \textit{meta-diversity} search. In a maze environment where 2 sets of hand-defined BC are provided (one for agent position and one for agent direction), they show that driving the search with the two BC sets simultaneously leads to higher probability to discover \textit{quality} behavior (meaning here solving the task) than with a single \textit{unaligned} BC. However, BCs are predefined and fixed, limiting the generalisation to complex environments.

\paragraph*{State Representation Learning} Many approaches have been proposed for state representation learning in reinforcement learning~\cite{lesort2018state}. As presented in the introduction, goal-directed approaches generally rely on deep generative models such as VAEs~\cite{pere2018unsupervised,nair2018visual}. Others tune the representation to achieve target properties at the feature level such as disentanglement~\cite{higgins2017darla} or linear predictability~\cite{watter2015embed}. This includes coupling with predictive forward  and inverse models~\cite{van2016stable,karl2016deep,higgins2017darla,pathak2017curiosity,hafner2018learning,zhang2018decoupling} or priors of independent controllability  ~\cite{thomas2017independently,laversanne2018curiosity}. However, they all rely on a single embedding space, where all the observed instances are mapped to the same set of features. 

\paragraph*{Continual Unsupervised Representation Learning} Recent work also proposed to dynamically expand the network capacity of a VAE~\cite{rao2019continual, lee2020neural,li2020progressive}. Similarities and differences with HOLMES are discussed in Appendix~\ref{sm:sec:CURL}. However, these approaches were applied to passively observed datasets, either targeting unsupervised clustering of sequentially-received class of images or disentanglement of factors of variations in generative datasets.


\paragraph*{Interactive Exploration of Patterns} Interactive evolutionary computation (IEC) ~\cite{langdon2005pfeiffer,secretan2011picbreeder,simon2019ganbreeder} aims to integrate external human feedback to explore high complexity pattern spaces, via intuitive interfaces. However, the user must provide feedback at each generation by individually selecting interesting patterns for the next generation; whereas our framework requires much sparser feedback.

\newpage
\section{Discussion}
As stated in the Introduction, our contributions in this paper are threefold. First, in section~\ref{sec:problem_formulation}, we introduced the novel objective of \textit{meta-diversity} search in the context of automated discovery in morphogenetic systems. Then, in section~\ref{sec:HOLMES}, we proposed a \textit{dynamic} and \textit{modular} model architecture for meta-diversity search through unsupervised learning of \textit{diverse} representations. Finally, in section
~\ref{sec:experimental_results}, we showed that search can easily be guided toward complex pattern preferences of a simulated end user, using very little user feedback.

To our knowledge, HOLMES is the first algorithm that proposes to progressively grow the capacity of the agent visual world model into an organized hierarchical representation. There are however several limitations to be addressed in future works. The architecture remains quite \textit{rigid} in the way it is isolating the different niches of patterns (binary tree with frozen boundaries) whereas other approaches, further leveraging human feedback, could be envisaged. 

The question whether \textit{machines} can really help scientists for crucial discoveries in Science, although appealing, is still an open question~\cite{bertolaso2020critical}.
We believe that machine learning algorithms integrating flexible modular representation learning with intrinsically-motivated goal exploration processes for \textit{meta-diversity} search are very promising directions. As an example, despite the limitations mentioned above, IMGEP-HOLMES was able to discover many types of solitons including unseen pattern-emitting lifeforms in less than 15000 training steps without guidance, when their existence remained an open question raised in the original Lenia paper~\cite{chan2018lenia}.


\section*{Broader Impact Statement}

We introduced methods that can be used as tools to help human scientists discover novel structures in complex dynamical systems.
While experiments presented in this article were performed using an artificial system (continuous cellular automaton), they also target to be used for automated discovery of novel structures in fields ranging from biology to physics. As an example, Grizou et al. \cite{grizou2020curious} recently showed how IMGEPs can be used to automate chemistry experiments addressing fundamental questions related to the origins of life (how oil droplets may self-organize into proto-cellular structures), leading to new insights about oil droplet chemistry.
As experiments in Grizou et al. used a single pre-defined BC, one can expect that the new approach presented in this paper may boost
the efficiency of its use in bio-physical systems, that could include systems related to design of new materials or new drugs.
As a tool enabling scientist to better understand the space of dynamics of such systems, we believe it could help them
better understand how to leverage such dynamics for societally useful purposes, and avoid negative effects, e.g. due to unpredicted self-organized dynamics.

However, technological and scientific discoveries might have a considerable impact in modern societies. Introducing machine decisions in the process should therefore be done with great responsibility, taking care of carefully identifying and balancing the biases inherent to any ML algorithms. The methods proposed in this paper constitute a first step in this direction by quantitatively measuring the influence of biases, in both predefined and learned BC spaces, on the algorithm discoveries. With an increasing interest in ML for automated discovery, it will be fundamental to to improve and extend these methods in the near future.

Besides, by releasing our code, we believe that we help efforts in reproducible science and allow the wider community to build upon and extend our work in the future. 

\section*{Acknowledgements and Disclosure of Funding}
We would like to thank Chris Reinke and Bert Chan for useful inputs and discussions for the paper, as well as Jonathan Grizou for valuable suggestions.
The authors acknowledge support from the Human Frontiers Science Program (Collaborative Research Grant RGP0018/2016) and from Inria.
This work also benefited from access to the HPC resources of IDRIS under the allocation 2020-[A0091011996] made by GENCI, using the Jean Zay supercomputer.

\newpage
\bibliography{neurips_2020}

\clearpage

\section*{Supplementary Material}
\appendix
\setcounter{page}{1}

This supplementary material provides implementation details, hyper-parameters settings, additional results and visualisations. 

\begin{itemize}
    \item Section~\ref{sm:sec:design_choices_IMGEP-HOLMES} presents a focus on the design choices we use for IMGEP-HOLMES
    \item Section~\ref{sm:sec:evaluation_procedure} provides implementation details for the main paper evaluation procedure
    \begin{itemize}
        \item \ref{sm:subsec:diversity_evaluation}: Quantitative evaluation of diversity
        \item \ref{sm:subsec:RSA_evaluation}: Quantitative evaluation of Representational Similarity
        \item  \ref{sm:subsec:human_evaluation}: Human-evaluator selection of the BC spaces for evaluating SLP and TLP diversity
    \end{itemize}
    \item Section~\ref{sm:sec:experimental_settings} provides all necessary implementation details for reproducing the main paper experiments
    \begin{itemize}
        \item \ref{sm:subsec:lenia_settings}: Lenia environment settings
        \item \ref{sm:subsec:parameter_sampling}: Parameter-sampling policy $\Pi$ settings for Lenia’s initial state and update rule
        \item \ref{sm:subsec:imgep_variants}: Settings for training the BC spaces in IMGEP-VAE and IMGEP-HOLMES
    \end{itemize}
    \item Section~\ref{sm:sec:additional_results} provides additional results that complete the ones from the main paper
    \begin{itemize}
        \item \ref{sm:subsec:RSA_results}: Complete RSA analysis of the hierarchy of behavioral characterizations learned in HOLMES
        \item \ref{sm:subsec:monolithic_baselines}: Additional IMGEP baselines with a monolithic BC space are compared 
        \item \ref{sm:subsec:impact_lateral_connections}: Ablation study of the impact of the lateral connections in HOLMES
    \end{itemize}
    \item Section~\ref{sm:sec:CURL} discusses the comparison of HOLMES with other model-expansion architectures
    \item Section~\ref{sm:sec:additional_visualisations} provides qualitative visualisations of the hierarchical trees that were autonomously constructed by the different IMGEP-HOLMES variants.
\end{itemize}

\paragraph*{Source code:} Please refer to the project website \url{http://mayalenE.github.io/holmes/} for the source code and complete database of discoveries for our experiments. 


\clearpage
\section{Focus on IMGEP-HOLMES}
\label{sm:sec:design_choices_IMGEP-HOLMES}

\paragraph*{Algorithm~\ref{sm:algo:IMGEP-HOLMES}: IMGEP-HOLMES pseudo-code.} Please refer to section~\ref{sec:HOLMES} of the main paper for the step-by-step implementation choices and to section~\ref{sm:sec:experimental_settings} in apppendix for the implementation details.

\begin{algorithm}[h!]
\caption{IMGEP-HOLMES}
\label{sm:algo:IMGEP-HOLMES}
\vspace{-10pt}
\begin{multicols}{2}
\begin{algorithmic}
\STATE \textbf{Inputs: }Parameter-sampling policy $\Pi$
\STATE Initialize root representation $\mathcal{R} = \{\mathcal{R}_0\}$

\FOR {$k\leftarrow 1 \: \mathbf{to} \: N$}

\IF [initial random iterations]{\small $k<N_{init}$}
\STATE Sample $\theta \sim \mathcal{U}(\Theta)$

\ELSE[goal-directed iterations]
\STATE Sample a target BC space $\hat{BC} \sim G_{s}(\mathcal{H})$
\STATE Sample a goal $\hat{g} \sim G(\hat{BC}, \mathcal{H})$ in $\hat{BC}$ 
\STATE Choose $\theta \sim \Pi(\hat{BC}, \hat{g}, \mathcal{H})$  
\ENDIF
\STATE Rollout experiment with $\theta$ and observe $o$ 

\STATE
\STATE \COMMENT{Encode reached goals in the hierarchy}
\STATE Start with root node $i \gets 0, parent(i)=\emptyset $ 
\WHILE{$i$ exists in the hierarchy (until leaf)}
\STATE $ r_i = \mathcal{R}_i(o, \mathcal{R}_{parent(i)}(o))$ 
\STATE Append $(\theta, o, r_i)$ to the history $\mathcal{H}$ 
\STATE  $i \gets ic $, $c = \mathcal{B}_i(r_i)$  \COMMENT{go to left or right child}
\ENDWHILE

\STATE
\STATE \COMMENT{Augment representational capacity}
\IF{a BC space $BC_i$ is \textit{saturated}} 
\STATE Freeze $\mathcal{R}_i$ weights
\STATE Define a boundary $\mathcal{B}_i: BC_i \rightarrow \{0,1\}$ 
\STATE Instantiate child modules $\mathcal{R}_{i0}$ and $\mathcal{R}_{i1}$
\STATE \COMMENT{Project past discoveries to children BCs}
\FOR {$(\theta, o, r) \in \mathcal{H}[BC_i]$} 
\STATE  $\mathcal{R}_j \gets \mathcal{R}_{ic}$, $c = B_i(r)$
\STATE Append $(\theta, o, \mathcal{R}_j(o))$ to $\mathcal{H}[j]$
\ENDFOR
\ENDIF

\STATE
\STATE \COMMENT{Periodically train HOLMES}
\IF{training requested}
\FOR {E epochs} 
\STATE Train the hierarchy $\mathcal{R}$ on observations in $\mathcal{H}$ with importance sampling
\ENDFOR
\STATE  \COMMENT {Update the database of reached goals}
\FOR {$i \in \textnormal{hierarchy}$}
\FOR {$(\theta, o, r) \in \mathcal{H}[i]$}
\STATE $\mathcal{H}[i][r] \gets \mathcal{R}_i(o)$
\ENDFOR
\ENDFOR
\ENDIF

\STATE
\STATE \COMMENT{Ask for user feedback}
\IF{user feedback requested}
\STATE Ask user to score leaf BC spaces
\STATE Update $G_s$ with assigned scores
\ENDIF

\ENDFOR
\end{algorithmic}
\end{multicols}
\vspace{-8pt}
\end{algorithm}

\subsection{Design choices in HOLMES}
\label{sm:subsec:HOLMES}

\begin{figure}[h]
\centering
 \includegraphics[width=\linewidth]{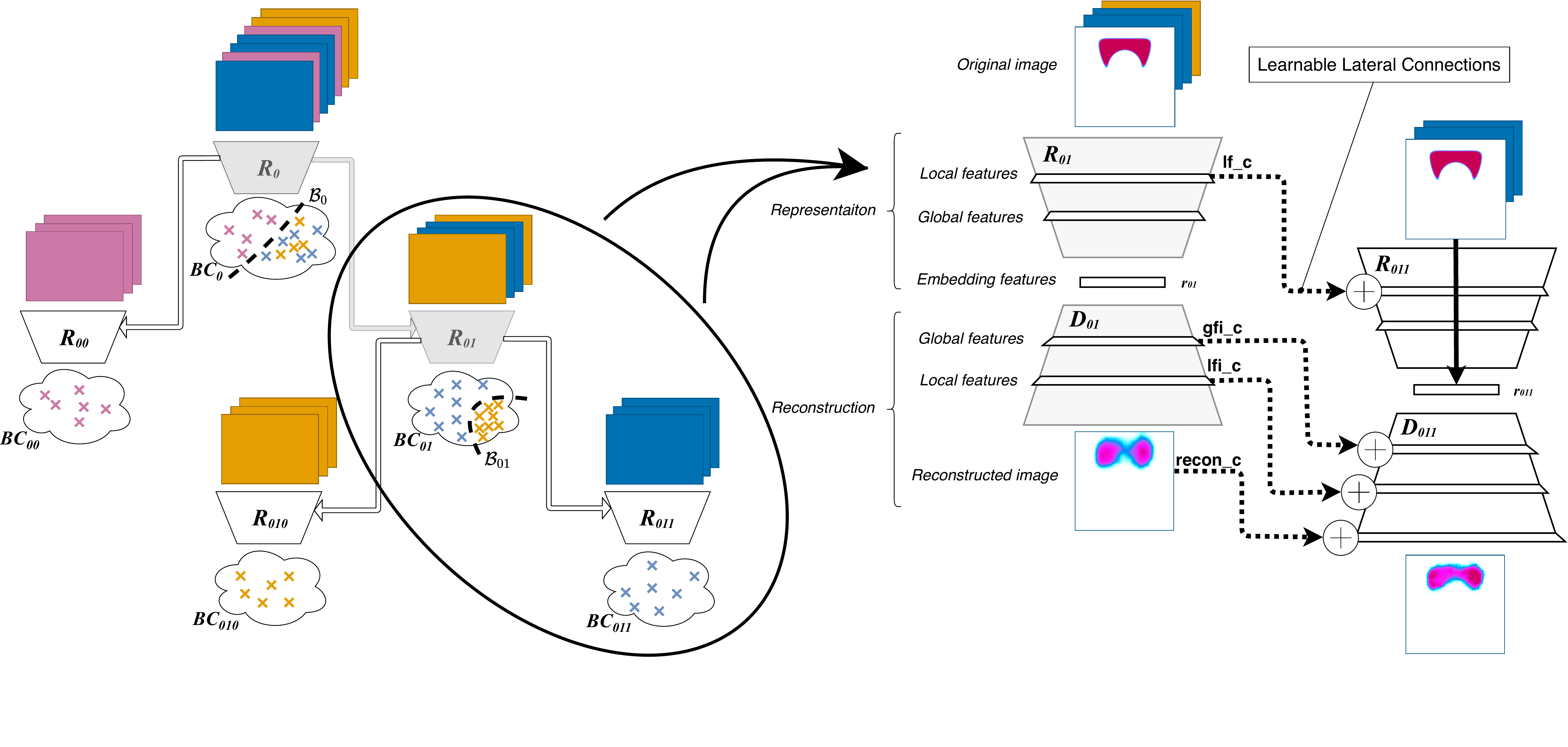} 
 \vspace{-20pt}
 \caption{Focus on the different design choices made for the HOLMES architecture. (Left) Each module uses a VAE~\cite{kingma2013auto} as the base architecture, where the embedding $R_i$ is coupled to a decoder $D_i$ ($D_i$ is not shown on the left panel for readability). All non-leaf node VAEs are frozen as well as their incoming lateral connections (light grey). The leaf nodes are incrementally trained on their own niches of patterns (represented as colored squares above the embeddings) defined by the boundaries fitted at each node split (curved dotted lines in each BC space, represented as clouds). (Right) $R_{011}$ is trained to encode new information in a latent representation $r_{011}$ (plain vertical arrow) by learning to reuse its parent knowledge via the lateral connections (dotted arrows, denoted as l\_f, gfi\_c, lfi\_c, recon\_c).}
\label{sm:fig:HOLMES_focus}
\end{figure}




While the global architecture is generic and numerous design choices can be made, this section details the practical implementation for the \textit{modules}, \textit{connection scheme}, and \textit{splitting criteria} used in this paper. We summarize those components in Figure~\ref{sm:fig:HOLMES_focus}.

\paragraph*{Choice for the base module} Each module has an embedding network $R_i$ that maps an observation $o$ to a low-dimensional vector $r=R(o)$. To learn such embedding, we rely on a variational autoencoder network~\cite{kingma2013auto} for the base module. The encoder network $R_i: q_\phi(r|x)$ is coupled with a decoder network $D_i: p_\theta(x|r)$ that enables a generative process from the latent space, and the networks are jointly trained to maximize the marginal log-likelihood of the training data with a regularizer on the latent structure. \\
The training loss is $\mathcal{L}_{\textsc{vae}} (\theta, \phi; \textbf{x}, \textbf{r} )= \underbrace{\mathbb{E}_{\hat{p}(\textbf{x})}\left(\mathbb{E}_{q_\phi(\textbf{r}|\textbf{x})}\left( -\log p_\theta(\textbf{x}|\textbf{r}) \right)\right)}_a + \underbrace{\mathbb{E}_{\hat{p}(\textbf{x})}\left( D_{\textsc{KL}} \left( q_\phi(\textbf{r}|\textbf{x}) || p(\textbf{r}) \right) \right)}_b $, where (a) represents the expected reconstruction error (computed with binary cross entropy) and (b) is the regularizer KL divergence loss of the approximate diagonal Gaussian posterior $q_\phi(r|x)$ from the standard Gaussian prior $p(z) = \mathcal{N} (0, I)$. Please note that input observations are partitioned between the different nodes in HOLMES, therefore each module VAE is trained only on its niche of patterns. 
Only the encoder network $R_i$ is kept in IMGEP-HOLMES (Algorihm~\ref{sm:algo:IMGEP-HOLMES}), therefore other choices for the base module and training strategy could be envisaged in future work, for instance with contrastive approaches instead of generative approaches.

\paragraph*{Choice for the connection scheme} The connection scheme takes inspiration from \textit{Progressive Neural Networks} (PNN) \citep{rusu2016progressive}), where
transfer is enabled by connecting the different modules via learned \textit{lateral connections}. To mitigate the growing number of parameters, we opted for a sparser connection scheme that in \citep{rusu2016progressive}.  The connection scheme is summarized in Figure~\ref{sm:fig:HOLMES_focus}. Connections are only instantiated between a child and its parent (hierarchical passing of information). Connections are only instantiated between a reduced number of layers (denoted as l\_f, gfi\_c, lfi\_c, recon\_c in the figure). We hypothesize that transfer is beneficial in the decoder network so a child module can reconstruct \enquote{as well as} its parent, however connections are removed between encoders as new complementary type of features should be learned. We preserve the connections only at the local feature level, as the CNN first layers tend to learn similar features \citep{yosinski2014transferable}. Connections between linear layers are defined as linear layers and connections between convolutional layers are defined as convolutions with $1 \times 1$ kernel. At each connection level, the output of the connection is summed to the current feature map in the VAE. Other connection schemes could be envisaged in future work, for instance with FiLM layers~\cite{pere2018unsupervised} (feature-wise affine transformation instead of sum) which have recently been proposed for vision models.

\paragraph*{Choice for the splitting criteria} There are two main choices: \textit{when} to split a node and \textit{how} to redirect the patterns toward either the left or right children. For both, we opted for simple design choices that allow the split to be unsupervisedly and autonomously handled during the exploration loop. We trigger a split in a node when the reconstruction loss of its VAE reaches a plateau, with additional conditions to prevent premature splitting (minimal node population and minimal number of training steps) or to limit the total number of splits. When splitting a node, we use K-means algorithm in the embedding space to fit 2 clusters on the points that are currently in the node. This generates a boundary in the latent space of the node, that we keep fixed for the rest of the exploration loop. Again, many other choices could be envisaged in future work, for instance by including human feedback to fit the boundary or with more advanced clustering algorithms.

\clearpage
\section{Complete Description of the Evaluation Procedure}
\label{sm:sec:evaluation_procedure}

\subsection{Evaluation of diversity}
\label{sm:subsec:diversity_evaluation}

\subsubsection{Construction of 5 analytic BC spaces}
\label{sm:subsubsec:analytic_BCs}
This section details the 5 BC spaces introduced in section~\ref{subsec:results_predefinedBC} of the main paper. Each set of BC features relies either on \textit{engineered} representation based on existing image descriptors from the literature or on \textit{pretrained} representations unsupervisedly learned on Lenia patterns. Those BCs were constructed to characterize different \textit{types} of diversities in the scope of evaluating \textit{meta-}diversity as defined in section
~\ref{sec:problem_formulation}, but obviously many others could be envisaged. The 5 BC models are provided with the source code of this paper.

Each set of BC features is defined by a mapping function $BC_X: o \in [0,1]^{256 \times 256} \mapsto \hat{z} \in [0,1]^8 $ where $X$ is the corresponding BC space, $o$ is a Lenia pattern and $\hat{z}$ represents its 8-dimensional behavioral descriptor in the corresponding BC space. 

We denote $\mathcal{D}_{ref}$ an external dataset of 15000 Lenia patterns. The patterns in $\mathcal{D}_{ref}$  were randomly collected from prior exploration experiments in Lenia, experiments that include different random seeds and different exploration variants and comport 50\% SLPs and 50\% TLPs. $\mathcal{D}_{ref}$ is a large database that is intended to cover a diversity of patterns orders of magnitude larger than what could be found in any single algorithm experiment, and that we use as reference dataset to construct and normalize the different evaluation BC spaces.

\paragraph*{Spectrum-Fourier} The 2-dimensional discrete Fourier transform is a mathematical method that projects an image (2D spatial signal) into the frequency domain, from which frequency characteristics can be extracted and used as texture descriptors~\cite{smach2008generalized}. Applications range from material description~\cite{muller1998fourier}, leaf texture description in biology~\cite{cope2010plant} and rule classification in cellular automata~\cite{machicao2018cellular}.

\begin{figure}[h!]
\centering
 \includegraphics[width=\linewidth]{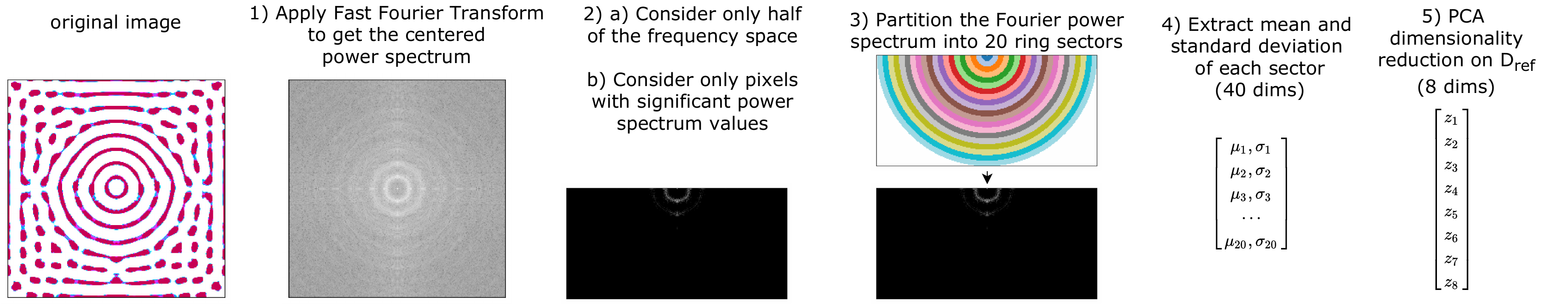} 
 \caption{Construction of \textsc{Spectrum-Fourier} analytic space. See text for details. Please note that for visualisation purposes: (left) the original image is colorized but is originally a $256 \times 256$ grayscale image; (step 1-2-3) the power spectrum is depicted in logarithmic scale.}
\label{sm:fig:spectrum_fourier}
\end{figure}

The construction of \textsc{BC\textsubscript{Spectrum-Fourier}} is summarized in Figure~\ref{sm:fig:spectrum_fourier} and follows the below procedure: \begin{enumerate}

    \item The 2D Fast Fourier Transform transforms the image $o=f(x,y)$ into the $u,v$ frequency domain function $F$, the zero-frequency component is shifted to the center of the array and the power specrum $PS$ (or power spectral density) is computed: 
    \begin{align*}
    F(u,v) & = \frac{1}{256 \times 256}\sum_{x=0}^{255} \sum_{y=0}^{255} f(x,y) \exp^{-j2\pi \frac{ux}{256} \frac{vx}{256}} \\
    F(u,v) & \gets Roll(F(u,v), (\frac{256}{2}, \frac{256}{2})) \\
    PS(u,v) & = Real(F(u,v))^2 + Imaginary(F(u,v))^2
     \end{align*}
    
    \item The power spectrum is filtered to keep only the lower half (symmetry property of the FFT) and the significant values:
    \begin{align*}
        & PS(u,v) = \{PS(u,v), 0 \le u \le \frac{256}{2}, - \frac{256}{2} \le v \le \frac{256}{2} - 1 \} \\
        & PS(u,v) = 0 \: if \: PS(u,v) < mean(PS(u,v))
    \end{align*}
    
    \item  The power spectrum is partitioned into 20 ring-shaped sectors:
    $$ \left[ R_i = \{ PS(u,v) | r_1^2 \le u^2 + v^2 \le r_2^2 \} \: with \: (r_1,r_2) = (\frac{i}{20}  \times \frac{256}{2}, \frac{i+1}{20}  \times \frac{256}{2});  \: for \: i \in [0 .. 19 ] \right] $$
    
    \item A 40-dimensional feature vector (FV) representing radially-aggregating measures (mean $\mu_i$ and standard deviation $\sigma_i$ of each sector) is extracted:
    \begin{align*}
      FV(o) = [\mu_1, \sigma_1, \dots, \mu_{20}, \sigma_{20}] & , \\
      \text{ where } \mu_i & = mean(PS[R_i]), \sigma_i = std(PS[R_i]) 
    \end{align*}
    
    \item The 40-dimensional feature vector $FV$ is projected into a normalized 8-dimensional behavioral descriptor $\hat{z}$ using a transformation $\hat{T}: FV \mapsto \hat{z}$. $\hat{T}$ is constructed with Principal Component Analysis (PCA)~\cite{wold1987principal} dimensionality reduction on $\mathcal{D}_{ref}$:
    \begin{align*}
       & X_{ref} = \{ FV(o), o \in \mathcal{D}_{ref} \} \\
       & \text{Fit a PCA with 8 components on $X_{ref}$,  } PCA: FV \in \mathbb{R}^{40} \mapsto z \in \mathbb{R}^8 \\
       & z_{ref} = PCA(X_{ref}),  z_{min} = percentile(z_{ref}, 0.01), z_{max} = percentile(z_{ref}, 99.9)) \\ 
       & \hat{T}: FV  \mapsto \Hat{z} = \frac{PCA(FV) - z_{min}}{z_{max} - z_{min}} 
    \end{align*}
    
    \item \textsc{BC\textsubscript{Spectrum-Fourier}}$(o) = \hat{T} \circ FV (o)$

\end{enumerate} 

\paragraph*{Elliptical-Fourier} Elliptical Fourier analysis (EFA)~\cite{kuhl1982elliptic} is a mathematical method for contour description which has been widely-used for shape description in image processing~\cite{russ1990image}. These descriptors have been applied to morphometrical analysis in biology~\cite{lestrel1997fourier}, for instance to characterize the phenotype of plants leaf and petal contours~\cite{neto2006plant} or anatomical shape changes ~\cite{chen2000describing, friess2003exploring}. 

A closed contour $\{ x_p, y_p \}_{p=1}^K$ (K points polygon) can be seen as a continuous periodic function of the \textit{length} parameter $T = \sum_{p=1}^K \Delta t_p$ where $t_p$ is the distance from the $p-1^{th}$ to the $p^{th}$ point. Therefore it can be represented as a sum of cosine and sine functions of growing frequencies (harmonics) under Fourier approximation. Each harmonic is an ellipse which is defined by 4 coefficients $a, b, c, d$. 

\begin{figure}[h!]
\centering
 \includegraphics[width=\linewidth]{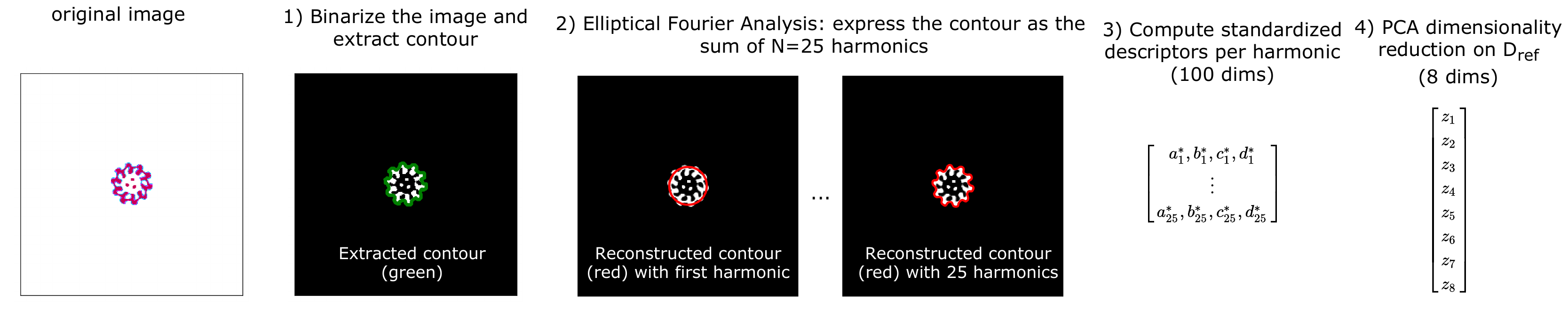} 
 \caption{Construction of \textsc{Elliptical-Fourier} analytic space. See text for details. (step 1) The contour depicted in green is extracted with OpenCV \texttt{findContours()} function (step 2) The contours depicted in red are reconstructed from the EFA coefficients (like in other Fourier series transforms the shape signal can be approximated by summing the harmonics~\cite{kuhl1982elliptic}).}
\label{sm:fig:elliptical_fourier}
\end{figure}

The construction of \textsc{BC\textsubscript{Elliptical-Fourier}} is summarized in Figure~\ref{sm:fig:elliptical_fourier} and follows the below procedure: \begin{enumerate}

    \item Binarize the image \texttt{o\textsubscript{binary} = o > 0.2} and extract the external contour as the a list of the (x,y) positions of the pixels that make up the boundary using OpenCV function \texttt{contour = cv2.findContours(o\textsubscript{binary}, cv2.RETR\_EXTERNAL, cv2.CHAIN\_APPROX\_SIMPLE)}
    
    \item Extract the set of $\{a_n, b_n, c_n, d_n\}_{n=1}^N$ coefficients for a series of N ellipses (N=25) from the x- and y-deltas ($\Delta x_p$ and $\Delta y_p$) between each consecutive point p in the K points polygon:
    \begin{align*}
    a_n & = \frac{T}{2 n^2 \pi^2} \sum\limits_{p=1}^K \frac{\Delta x_p}{\Delta t_p} \left[ \cos{\frac{2 n \pi t_p}{T}} - \cos{\frac{2 n \pi t_{p-1}}{T}} \right]\\
    b_n & = \frac{T}{2 n^2 \pi^2} \sum\limits_{p=1}^K \frac{\Delta x_p}{\Delta t_p} \left[ \sin{\frac{2 n \pi t_p}{T}} - \sin{\frac{2 n \pi t_{p-1}}{T}} \right]\\
    c_n & = \frac{T}{2 n^2 \pi^2} \sum\limits_{p=1}^K \frac{\Delta y_p}{\Delta t_p} \left[ \cos{\frac{2 n \pi t_p}{T}} - \cos{\frac{2 n \pi t_{p-1}}{T}} \right]\\
    d_n & = \frac{T}{2 n^2 \pi^2} \sum\limits_{p=1}^K \frac{\Delta y_p}{\Delta t_p} \left[ \sin{\frac{2 n \pi t_p}{T}} - \sin{\frac{2 n \pi t_{p-1}}{T}} \right]
    \end{align*}
    \item The coefficients are standardized (i.e. made invariant to size, rotation and shift): \\
    $ \begin{bmatrix} a_n^* & b _n^* \\ c_n^* & d _n^*\end{bmatrix} = \frac{1}{L} \begin{bmatrix} \cos \phi & \sin \phi \\ -\sin \phi & \cos \phi \end{bmatrix} \begin{bmatrix} a_n & b _n \\ c_n & d _n\end{bmatrix} \begin{bmatrix} \cos N \theta & \sin N \theta \\ -\sin N \theta & \cos N \theta \end{bmatrix} $, where $L= \sqrt{\left[(A_0 - x_m)^2 + (C_0-x_m)^2 \right]}$, $(A_0, C_0)$ is the center of the $1^{st}$ harmonic ellipse, $(x_m, y_m)$ is the location of the modified starting point (on the major axis of the ellipse), $\theta = \frac{2 \pi t_m}{T} $ and $\phi = \tan^{-1} \frac{y_m - C_0}{x_m - A_0}$ (angle between the major axis of the ellipse and xaxis).
    
    \item The 100-dimensional feature vector $FV = \{a_n^*, b_n^*, c_n^*, d_n^*\}_{n=1}^{25}$ is projected into a normalized 8-dimensional behavioral descriptor using a transformation  $\hat{T}: FV \mapsto \hat{z}$. $\hat{T}$ is constructed with Principal Component Analysis (PCA) dimensionality reduction on $\mathcal{D}_{ref}$ (similar procedure as in point 5 of \textsc{BC\textsubscript{Spectrum-Fourier}}).
  
    \item \textsc{BC\textsubscript{Elliptical-Fourier}}$(o) = \hat{T} \circ FV (o)$

\end{enumerate} 

\paragraph*{Lenia-Statistics} The original Lenia paper proposes several measures for statistical analysis of the Lenia patterns (section 2.4.2 in~\cite{chan2018lenia}), also defined in Reinke et al.~(2020) (section B.3 in ~\cite{Reinke2020Intrinsically}).

\textsc{BC\textsubscript{Lenia-Statistics}} is constructed on top of these measures according to the below procedure: \begin{enumerate}
    \item Among all the statistical measures proposed in~\cite{chan2018lenia} we selected the 17 measures that are time-independent, i.e. that can be computed from the final Lenia pattern $o = I(x,y)$, namely: \begin{itemize}
        \item the activation mass $m =  \frac{1}{256 \times 256} \sum\limits_{(x,y) \in I} I(x,y)$
        \item the activation volume $V_m =  \frac{1}{256 \times 256} \sum\limits_{(x,y) \in I} \delta_{I(x,y) > \epsilon}$ ($\epsilon = 10^{-4}$)
        \item the activation density $\rho_m =  \frac{m}{V_m}$
        \item the centeredness of the activation mass distribution \\ 
        $C_m =  \frac{1}{m} \sum\limits_{(x,y) \in I} w_{xy} \cdot I(x-\bar{x}_m, y-\bar{y}_m)$  where $(\bar{x}_m, \bar{y}_m)$ is the activation centroid \\
        and $w_{x,y} = \left(1 - \frac{d(x,y)}{\underset{x,y}{\max}d(x,y)} \right)^2$ with $d(x,y) = \sqrt{(x-\bar{x}_m)^2 + (y-\bar{y}_m)^2}$
        \item the 8 invariant image moments by Hu~\cite{hu1962visual}
        \item the 5 extra invariant image moments by Flusser~\cite{flusser2006moment}
    \end{itemize}
    
     \item The 17-dimensional feature vector $FV = \begin{bmatrix} m, V_m, \rho_m, C_m, hu_1, \dots , hu_7, flusser_8, \dots, flusser_{13} \end{bmatrix}$ is projected into a normalized 8-dimensional behavioral descriptor using a transformation  $\hat{T}: FV \mapsto \hat{z}$. $\hat{T}$ is constructed with Principal Component Analysis (PCA) dimensionality reduction on $\mathcal{D}_{ref}$ (similarly than for \textsc{BC\textsubscript{Spectrum-Fourier}}).
  
    \item \textsc{BC\textsubscript{Lenia-Statistics}}$(o) = \hat{T} \circ FV (o)$
\end{enumerate}

\paragraph*{BetaVAE} Reinke et al.~(2020)~\cite{Reinke2020Intrinsically} propose to train a $\beta$-VAE~\cite{burgess2018understanding} on a large database of Lenia pattterns and to reuse the learned features as behavioral descriptors for the analytic BC space. 

\textsc{BC\textsubscript{BetaVAE}} is constructed according to the below procedure : \begin{enumerate}
\item A $\beta$-VAE with 8-dimensional latent space is instantiated with the architecture detailed in table~\ref{sm:tab:BetaVAE_architecture}.
\begin{table}[b!]
	\centering
    \resizebox*{1.0\textwidth}{!}{%
	\begin{tabular}{ll}
   & \\
   \textbf{Encoder} & \textbf{Decoder} \\
   \cmidrule(lr){1-1} \cmidrule(lr){2-2}
   Input pattern A: $256\times256\times1$ & Input latent vector z: $8\times1$ \\
   Conv layer: 32 kernels $4\times4$, stride $2$, $1$-padding + ReLU & FC layers : 256 + ReLU,  256 + ReLU,  $4\times4\times32$ + ReLU \\
   Conv layer: 32 kernels $4\times4$, stride $2$, $1$-padding + ReLU & TransposeConv layer: 32 kernels $4\times4$, stride $2$, $1$-padding + ReLU \\
   Conv layer: 32 kernels $4\times4$, stride $2$, $1$-padding + ReLU & TransposeConv layer: 32 kernels $4\times4$, stride $2$, $1$-padding + ReLU \\
   Conv layer: 32 kernels $4\times4$, stride $2$, $1$-padding + ReLU & TransposeConv layer: 32 kernels $4\times4$, stride $2$, $1$-padding + ReLU \\
   Conv layer: 32 kernels $4\times4$, stride $2$, $1$-padding + ReLU & TransposeConv layer: 32 kernels $4\times4$, stride $2$, $1$-padding + ReLU \\
   Conv layer: 32 kernels $4\times4$, stride $2$, $1$-padding + ReLU & TransposeConv layer: 32 kernels $4\times4$, stride $2$, $1$-padding + ReLU \\
   FC layers : 256 + ReLU, 256 + ReLU, FC: $2\times8$ & TransposeConv layer: 32 kernels $4\times4$, stride $2$, $1$-padding \\
\end{tabular}}
    \vspace{0.2cm}
	\caption{$\beta$-VAE architecture used for \textsc{BC\textsubscript{BetaVAE}}.}
\label{sm:tab:BetaVAE_architecture}
\end{table}

\item The construction of the training dataset, training procedure and hyperparameters follows~\cite{Reinke2020Intrinsically}:  \begin{itemize}
    \item The $\beta$-VAE is trained on an external database $\mathcal{D}_{ref}^{(big)}$ of 42500 Lenia patterns (with 50\% SLP and 50\% TLP, 37500 as training set, 5000 as validation set) which were randomly collected from independent previous experiments (with the same procedure than $\mathcal{D}_{ref}$). 
    \item The $\beta$-VAE is trained for more than 1250 epochs with hyperparameters $\beta = 5$, Adam optimizer ($lr=1\mathrm{e}{-3}$, $\beta_1=0.9$, $\beta_2=0.999$, $\epsilon=1\mathrm{e}{-8}$, weight decay=$1\mathrm{e}{-5}$) and a batch size of 64.
    \item  The network weights which resulted in the minimal validation set error during the training are kept.
\end{itemize}
\item The resulting pretrained encoder serves as mapping function from a Lenia pattern $o$ to a 8-dimensional feature vector $FV(o) = \begin{bmatrix} z_1, z_2, z_3, z_4, z_5, z_6, z_7, z_8 \end{bmatrix}$

\item Similarly to the other analytic BC spaces in this paper, we use the reference dataset $\mathcal{D}_{ref}$ to normalize the 8-dimensional behavioral descriptors between $[0,1]$:
    \begin{align*}
       & z_{ref} = \{ FV(o), o \in \mathcal{D}_{ref} \}, z_{min} = percentile(z_{ref}, 0.01), z_{max} = percentile(z_{ref}, 99.9)) \\ 
       & \hat{T}: FV  \mapsto \Hat{z} = \frac{FV - z_{min}}{z_{max} - z_{min}} 
    \end{align*}

\item \textsc{BC\textsubscript{BetaVAE}}$(o) = \hat{T} \circ FV (o)$
\end{enumerate}

\paragraph*{Patch-BetaVAE} Reinke et al.~(2020) noticed that the $\beta$-VAE is not able to encode finer details and texture of patterns as the compression of the images to a 8-dimensional vector results in a general blurriness in the reconstructed patterns~\cite{Reinke2020Intrinsically}. Therefore, we also implemented an additional variant denoted as \textsc{Patch-BetaVAE} where the $\beta$-VAE is trained on \enquote{zoomed} $32 \times 32$ patches.
A preprocessing step extracts the cropped patch around the image activation centroid $P: o \mapsto o[\bar{x}_m-16:\bar{x}_m+16, \bar{y}_m-16:\bar{y}_m+16]$. Then, the construction of \textsc{BC\textsubscript{Patch-BetaVAE}} follows exactly the construction of \textsc{BC\textsubscript{BetaVAE}}, except that the network architecture has only 3 convolutional layers instead of 6. Following the notations of the previous paragraph, \textsc{BC\textsubscript{Patch-BetaVAE}}$(o) = \hat{T} \circ FV \circ P(o)$ with $FV$ the pretrained model on image patches and  $\hat{T}$ a normalizing function computed on $\mathcal{D}_{ref}$.

\subsubsection{Binning-based Diversity Metric}
\label{sm:subsubsec:diversity_measure}


We follow existing approaches in the QD and IMGEP literature~\cite{pugh2015confronting, pere2018unsupervised} and use binning-based measure to quantify the diversity of a set of explored instances into a predefined BC space. The entire BC space is discretized into a collection of t bins ${N_1, …, N_t}$ and the diversity is quantified as the number of bins filled over the course of exploration: $D_{|BC} = \sum\limits_{i=1}^t \delta_i$ where $\delta_i=1$ if the $N_i^{th}$ bin is filled, $\delta_i=0$ otherwise.

 We opt for a regular binning where each dimension of the BC space is discretized into equally sized bins. For all the results in the main paper, 20 bins per dimension are used for the discretization of the BC spaces. For recall, all the analytic BC spaces used in this paper are 8-dimensional and bounded in $[0,1]^8$ (see previous section). This results in a total of $25.6 \times 10^9$ bins. Note however that for a given BC space, the maximum number of bins that can be filled by all possible Lenia patterns is unknown.

\begin{figure}[h!]
\centering
\setlength\tabcolsep{1pt}
\begin{tabular}{@{}cc@{}}
(a) Diversity of SLP in \textsc{bc\textsubscript{Elliptical-Fourier}} & (b) Diversity of TLP in \textsc{bc\textsubscript{Lenia-Statistics}}\\
 \includegraphics[width=0.5\textwidth]{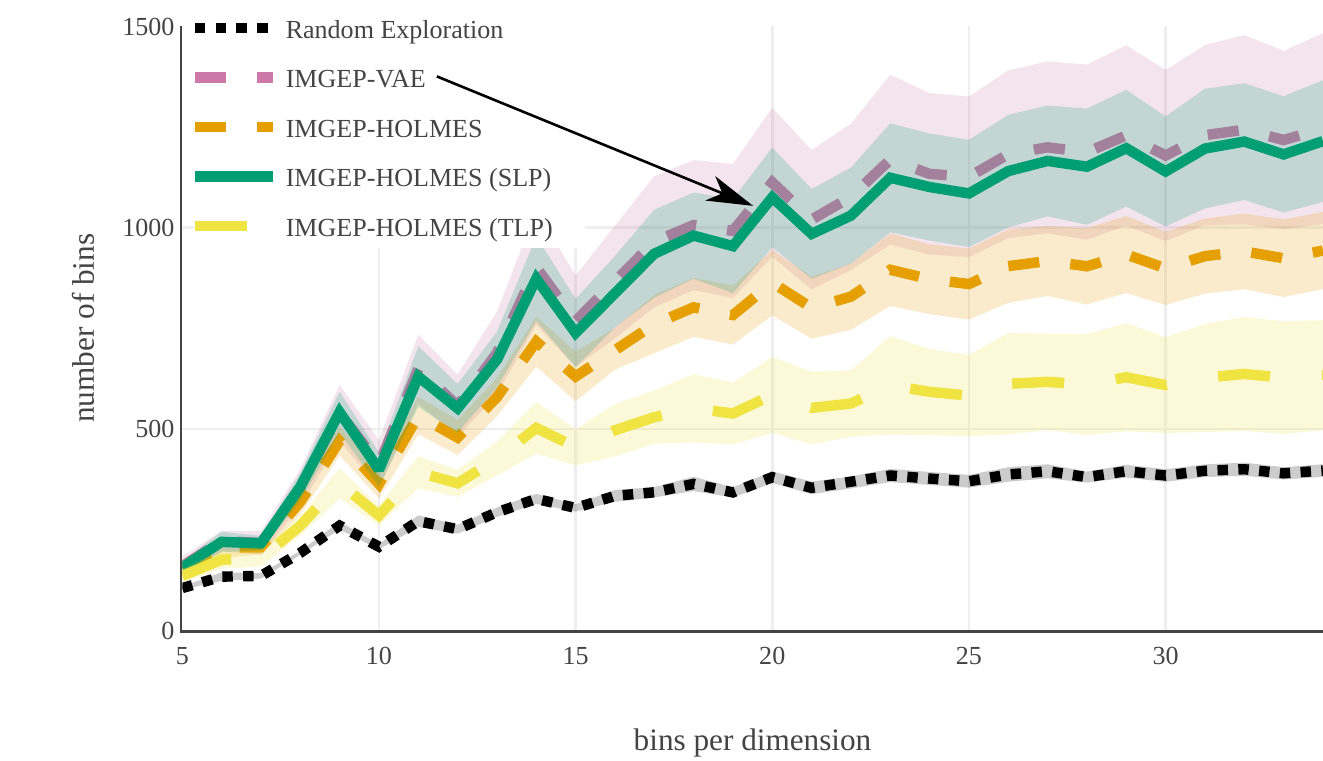} 
& 
 \includegraphics[width=0.5\textwidth]{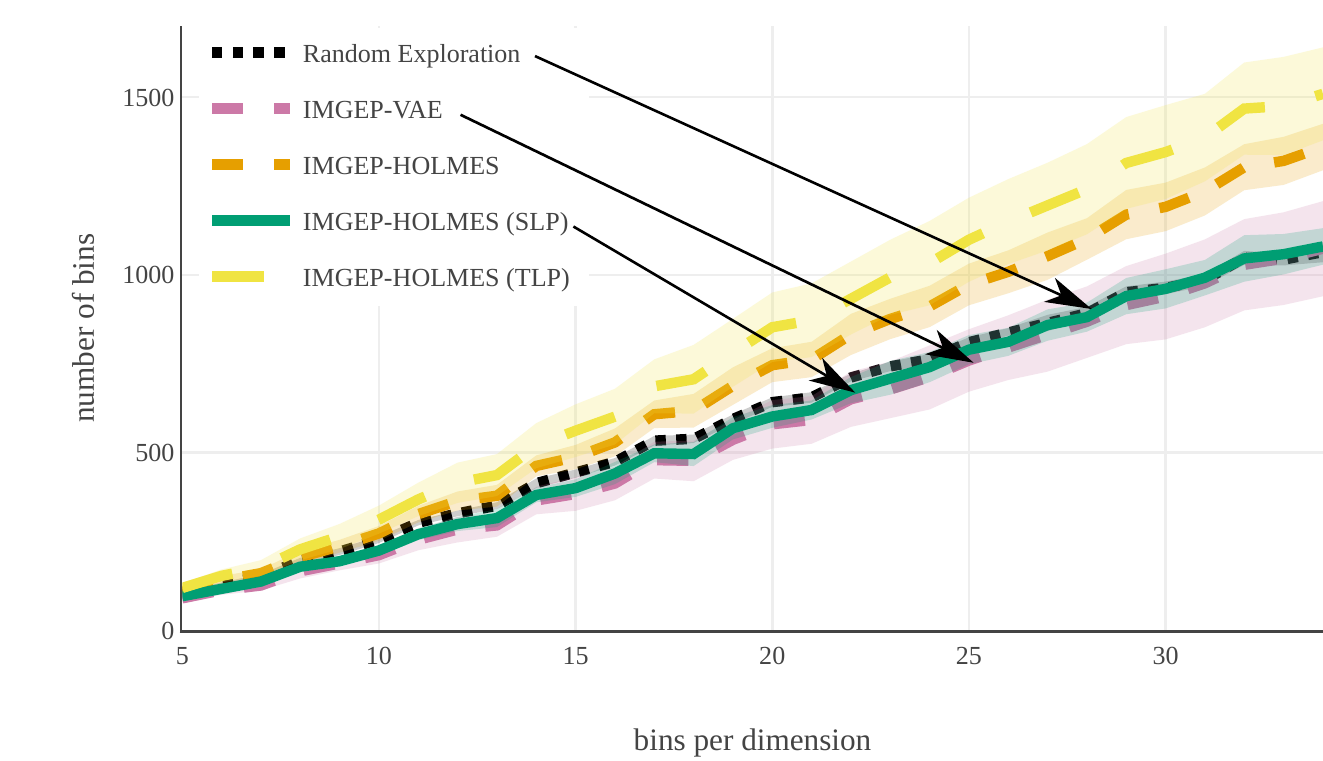}
 \end{tabular}
 \vspace{-5pt}
 \caption{Influence of the choice of the number of bins on the diversity measure presented in Figure 5 of the main paper. The final diversity (number of occupied bins at the end of exploration, as shown in the y axis) is measured by varying the number of bins per dimension from 5 to 35. Results in the main paper use n=20 bins. Mean and std-deviation shaded area curves are depicted.}
\label{sm:fig:diversity_dependance_num_of_bins}
\end{figure}

Because binning-based metrics directly depend on the choice of the bins discretization, we analyze in Figure~\ref{sm:fig:diversity_dependance_num_of_bins} the impact of the choice of the number of bins on the final diversity measure. As we can see, the ranking of the different IMGEP algorithms compared in Figure 5 of the main paper is invariant to this choice. 

\subsection{Representational Similarity Analysis}
\label{sm:subsec:RSA_evaluation}

We denote $\mathcal{D}_{ref}^{(small)}$ an external dataset of 3000 Lenia patterns (50\% SLPs and 50\% TLPs) which were collected with the same procedure than $\mathcal{D}_{ref}$. 

Given two representations embedding networks $R_i$ and $R_j$ with 8-dimensional latent space, the RSA similarity index $RSA_{ij}$ is computed with the  the linear Centered Kernel Alignment index (CKA) as proposed in~\cite{kornblith2019similarity}: \begin{enumerate}
    \item Compute the matrix of behavioral descriptors responses from each representation \\
    $Z_i = [R_i(o), o \in \mathcal{D}_{ref}^{(small)}] \in [0,1]^{3000 \times 8}$ and  $Z_j = [R_j(o), o \in \mathcal{D}_{ref}^{(small)}] \in [0,1]^{3000 \times 8}$ 
    
    \item Center the matrices responses: \\ $Z_i \gets Z_i - mean (Z_i, axis = 0)$ and $Z_j \gets Z_j - mean (Z_j, axis = 0)$
    
    \item $  RSA_{ij} = CKA (Z_i Z_i^T, Z_j Z_j^T)  = \frac{|| Z_i \cdot Z_j ^ T||_F^2}{|| Z_i \cdot Z_i ^ T||_F || Z_j \cdot Z_j ^ T||_F} \\
    \text{ where } || \cdot ||_F \text{ represents the Frobenius norm} $
\end{enumerate}

Representation Similarity Analysis (RSA) is used in Figure 3 of the main paper in two ways: \begin{itemize}
    \item To compare representations in \textit{time}, i.e. where the the embedding networks $R_i$ and $R_j$ come from the same network but from different training stages 
    \item To compare representations from different \textit{modules} in HOLMES where the the embedding networks $R_i$ and $R_j$ are taken from the same time step (end of exploration) but from different networks. 
\end{itemize}

\subsection{Human-Evaluator Selection of a \textit{proxy}-BC for Evaluation of SLP and TLP Diversity}
\label{sm:subsec:human_evaluation}

Relying on the external database $\mathcal{D}_{ref}$ of 15000 Lenia patterns (50\% SLP - 50\% TLP) and the SLP/TLP classifiers, we conducted an experiment with a human evaluator to select the analytic BC space that correlates the most with human judgement of what represents a \textit{diversity of SLP} and a  \textit{diversity of TLP}. 

\begin{figure}[h!]
\centering
 \includegraphics[width=\linewidth]{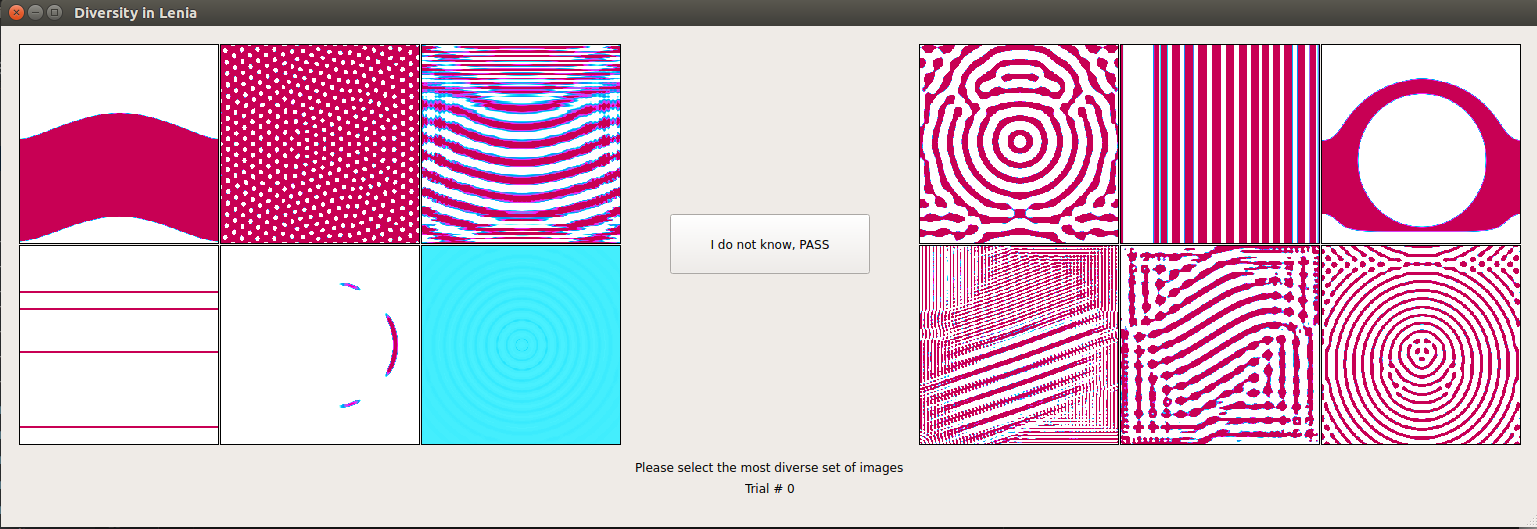} 
 \caption{Interface used for selecting of the best proxy-BC analytic space that correlates with human judgement of what represents a \textit{diversity of SLP} and a  \textit{diversity of TLP}.}
\label{sm:fig:human_interface}
\end{figure}

The experiment consisted in repeatedly showing the human with two sets of patterns (as shown in Figure~\ref{sm:fig:human_interface}) and asking the human to click on the set that he considers is the more \textit{diverse}, according to its intuitive notion of diversity. If the human cannot choose between the two sets, he can click on the \enquote{pass} button. In background, the procedure to generate the sets is the following: \begin{enumerate}
    \item Randomly select a (BC, category) pair, where BC $\in$ \{\textsc{Spectrum-Fourier}, \textsc{Elliptical-Fourier}, \textsc{Lenia-Statistics}, \textsc{BetaVAE}, \textsc{Patch-BetaVAE}\} and category $\in$ \{SLP,TLP\}.
    \item Randomly draw 750 candidate sets of 6 images among the 7500 patterns of the current category.
    \item Select the most \textit{similar} set and the most \textit{dissimilar} (i.e. diverse) set among those 750 sets. The (di-)ssimilarity of a set of 6 images is measured as a function of all the distances between each pair of images in the set, with distances being computed in the current BC space. This distance-based measure of diversity $D$, proposed in~\cite{scheiner2017decomposing}, measures the magnitude $M$ (dispersion) and variability $E$ (equability) of the set of S=6 points in the BC~\cite{scheiner2017decomposing}:
        \begin{align*}
            M &= \frac{S}{S-1} \sum_{i=1}^S \sum_{j=1}^S \frac{d_{ij}}{S^2}, \text{ where $d_{ij}$ is the pairwise euclidean distance}\\
            E &= \frac{1+\sqrt{1+4 H}}{2 S }, \text{ where } H = \left[ \sum_{i=1}^S \sum_{j=1}^S \left( \frac{d_{ij}}{\sum_{i=1}^S \sum_{j=1}^S d_{ij}} \right)^2 \right]^{\frac{1}{1-2}} \\
            D &= 1 + (S-1) \times E \times M, M \in [0,1] and  E \in [0,1]
        \end{align*}
       This measure replaces the binning-based measure which can hardly be used here (as they are only 6 images most candidate sets are likely to fall in the same number of bins and be equally diverse).
    \item The sets are displayed to the human in random presentation order.
\end{enumerate}

This experiment was conducted with one human evaluator which performed a total of 500 clicks, i.e 50 times per (BC, category) pair. For each click per (BC,category) pair, the agreement score is 0 if the human selected the opposite set that the one considered as diverse by the BC, 0.5 is the human selected the \enquote{pass} button and 1 if the human selected the set considered as diverse by the BC. Table~\ref{sm:table:human_agreement_scores} reports the mean and standard deviation agreement scores of the human evaluator for each (BC, category) pair. The agreement score is significant at level $\alpha = 5\%$ if it is above $0.64 = 0.5 + 1.96 \times \sqrt{\frac{0.25}{50}}$.

\begin{table}[h!]
  \caption{Human-evaluator agreement scores (mean $\pm$ std). Best scores are shown in bold. }
  \label{sm:table:human_agreement_scores}
  \centering
  \resizebox{\textwidth}{!}{%
  \begin{tabular}{lccccc}
    \toprule
        & \textbf{Spectrum-Fourier} &  \textbf{Elliptical-Fourier} &  \textbf{Lenia-Statistics} & \textbf{BetaVAE} &  \textbf{Patch-BetaVAE}\\
    \midrule
    \textbf{\textsc{SLP}} & $ 0.5 \pm 0.18 $ & $\mathbf{ 0.98 \pm 0.04 }$ & $ 0.59 \pm 0.12$ & $ 0.1 \pm 0.06 $ & $ 0.89 \pm 0.08$ \\
    \textbf{\textsc{TLP}} & $ 0.2 \pm 0.13 $ & $ 0.47 \pm 0.1$ & $\mathbf{ 0.92 \pm 0.07} $ & $ 0.75 \pm 0.08 $ & $ 0.38 \pm 0.08$ \\
    \bottomrule
  \end{tabular}
  }
\end{table}

As we can see, the human evaluator designated \textsc{bc\textsubscript{Elliptical-Fourier}} as the best proxy space for evaluating the diversity of SLP (98\% agreement score) and \textsc{bc\textsubscript{Lenia-Statistics}} as the best proxy space for evaluating the diversity TLP (92\% agreement score). This is why those BCs are used in Figure 5 of the main paper.

\clearpage
\section{Experimental Settings}
\label{sm:sec:experimental_settings}

\subsection{Environment Settings}
\label{sm:subsec:lenia_settings}

All experiments are done in the Lenia environment, as described in ~\cite{chan2018lenia, Reinke2020Intrinsically}. 
As stated in the main paper, we use a $256 \times 256$ state size ($A \in \mathbb{R}^{256 \times 256}$) and a number of $T = 200$ steps for each run.

The $256\times 256$ Lenia grid is a torus where the neighborhood is circular (i.e pixels on the top border are neighbors of the pixels on the bottom border and same between the left and right borders). 

Lenia's update rule ($A^t \rightarrow A^{t+1}$) is defined as $A^{t+1} = A^t + \Delta_\mathcal{T}  G(K \ast A^t)$, where:
\begin{itemize}
	\item G defines a parametrized \textit{growth mapping} function: exponential with $G(u;\mu,\sigma) = 2 \exp \left(- \frac{(u-\mu)^2}{2 \sigma^2}  \right)-1$
	\item K defines a parametrized concentric-ring Kernel with:
	\begin{align*}
	    K_C(r) &= \exp \left( \alpha - \frac{\alpha}{4 r (1-r)}\right)\text{, with }\alpha=4 \text{~~(Kernel core)} \\
	    K_S(r;\beta) &= \beta_{\lfloor Br \rfloor} K_C(Br~mod~1)\text{, with }\beta = (\beta_1, \beta_2, \beta_3) \text{~~(Kernel shell)} \\
	    K &= \frac{K_S}{|K_S|}
	\end{align*}
	
\end{itemize}

The update rule is therefore determined with 7 parameters: \begin{itemize}
    \item $R$: radius of the Kernel (i.e. radius of the local neighborhood that influences the evolution of each cell in the grid),
    \item $\mathcal{T} = \frac{1}{\Delta_\mathcal{T}}$: fraction of the growth update that is applied per time step,
    \item $\mu, \sigma$: growth center and growth width,
    \item $\beta_1, \beta_2, \beta_3$: concentring rings parameters that control the shape of the kernel.
\end{itemize}

See the project website \url{http://mayalenE.github.io/holmes/} for videos of the Lenia dynamics.

\subsection{Sampling of parameters $\theta$}
\label{sm:subsec:parameter_sampling}

The set of \textit{controllable} parameters $\theta$ of the artificial agent include:\begin{itemize}
    \item The update rules parameters $[R, \mathcal{T}, \mu, \sigma, \beta_1, \beta_2, \beta_3]$ 
    \item CPPN-parameters that control the generation of the initial state $A^{t=1}$
\end{itemize}

For each exploration run, the IMGEP agent samples a set of parameters $\theta \in \Theta$ that generates a rollout $A^{t=1} \rightarrow \dots \rightarrow A^{t=200} $.

Following the notations of Algorithm~\ref{sm:algo:IMGEP-HOLMES}, there are two ways parameters $\theta \in \Theta$ are sampled: \begin{enumerate}
    \item During the $N_{init}$ initial runs, parameters are randomly sampled $\theta \sim \mathcal{U}(\Theta)$
    \item During the goal-directed exploration runs, parameters are sampled from a policy $\theta \sim \Pi (\hat{BC_i}, \hat{g}, \mathcal{H})$. The  $\Pi$ policy operates in two steps: \begin{enumerate}
        \item given a goal $g \in \hat{BC_i}$, select parameters $\hat{\theta} \in \mathcal{H}$ whose corresponding outcome is closest to $g$ in $\hat{BC_i}$
        \item mutate the parameters by a random process $\theta = \textsc{mutation}(\hat{\theta})$
    \end{enumerate}
\end{enumerate}

We therefore need to define 1) the random process $\mathcal{U}$ used to randomly initialize the parameters $\theta$ and 2) the random \textsc{mutation} process used to mutate an existing set of parameters $\hat{\theta}$. 

Please note that for both we follow exactly the implementation proposed in Reinke et al.~(2020)~\cite{Reinke2020Intrinsically}. We therefore refer to section B.4 of their paper for a complete description of the implementation of the random initialization process and random mutation process. This includes the procedure used for parameters that control the generation of the initial pattern $A^{t=1}$ and for parameters that control Lenia's update rule. 

In this paper, we use the exact same hyperparameters as in ~\cite{Reinke2020Intrinsically} for initialization $\mathcal{U}$ and \textsc{mutation} of the CPPN-parameters that control the generation of the initial state $A^{t=1}$. We use slightly different hyper-parameters for the \textsc{mutation} of the parameters that control the generation of the update rule $[R, \mathcal{T}, \mu, \sigma, \beta_1, \beta_2, \beta_3]$, as detailed in table~\ref{sm:table:update_rule_sampling_parameters}.

\begin{table}[h!]
  \caption{Sampling of parameters for the update rule. The random initialization process $\mathcal{U}$ uses uniform sampling in an interval $[a,b]$. The random \textsc{mutation} is a Gaussian process $\theta = [\hat{\theta} + \mathcal{N}(\sigma_M)]_a^b$. }
  \label{sm:table:update_rule_sampling_parameters}
  \centering
  \begin{tabular}{lccccc}
    \toprule
        & $R$ &  $\mathcal{T}$ &  $\mu$ & $\sigma$ &  $(\beta_1, \beta_2, \beta_3)$ \\
    \midrule
     $[a,b]$ & $[2,20]$ & $[1,20]$ & $[0,1]$ & $[0.001, 0.3]$ & $[0, 1]$ \\
    \midrule 
    $\sigma_M$ & 0.5 & 0.5 & 0.1 & 0.05 & 0.1 \\
    \bottomrule
  \end{tabular}
  \end{table}

\subsection{Incremental Training of the BC Spaces}
\label{sm:subsec:imgep_variants}

\paragraph{Training Procedure} The networks are trained 100 epochs every 100 runs of exploration (resulting in 50 training stages and 5000 training epochs in total). The networks are initialized with \textit{kaiming} uniform initialization. We used the Adam optimizer ($lr=1\mathrm{e}{-3}$, $\beta_1=0.9$, $\beta_2=0.999$, $\epsilon=1\mathrm{e}{-8}$, weight decay=$1\mathrm{e}{-5}$) with a batch size of 128.

\paragraph*{Training Dataset} The datasets are incrementally constructed during exploration by gathering the discovered patterns. One pattern every ten is added to the validation set (10\%) and the rest is used in the training set (the validation dataset only serves for checking purposes and has no influence on the learned BC spaces). Importance sampling is used to give the newly-discovered patterns more weights. A weighted random sampler is used as follow: at each training stage t, there are $X$ patterns discovered so far among which $X_{new}$ have been discovered during the last 100 steps, we create a dataset $D_t$ of $X$ images that we construct by sampling $30\%$ among the $X_{new}$ lastly discovered images and $70\%$ among the $X-X_{new}$ old patterns. We also use data-augmentation, i.e at each training stage t, the images in $D_t$ are augmented online by random x and y translations (up to half the pattern size and with probability 0.6), rotation (up to 20 degrees and with probability 0.6), horizontal and vertical flipping (with probability 0.2), zooming (up to factor 3 with probability 0.6). The augmentations are preceded by spherical padding to preserve Lenia spherical continuity.

\paragraph*{IMGEP-VAE}  The monolithic VAE architecture used in the IMGEP-VAE baseline is detailed in table~\ref{sm:tab:monolithicVAE_architecture}. It has a total neural capacity of 2258657 parameters.

\begin{table}[h!]
	\centering
    \resizebox*{1.0\textwidth}{!}{%
	\begin{tabular}{ll}
   & \\
   \textbf{Encoder} & \textbf{Decoder} \\
   \cmidrule(lr){1-1} \cmidrule(lr){2-2}
   Input pattern A: $256\times256\times1$ & Input latent vector z: $16\times1$ \\
   Conv layer: 64 kernels $4\times4$, stride $2$, $1$-padding + ReLU & FC layers : 512+ ReLU,  512+ ReLU,  $4\times4\times64$ + ReLU \\
   Conv layer: 64 kernels $4\times4$, stride $2$, $1$-padding + ReLU & TransposeConv layer: 64 kernels $4\times4$, stride $2$, $1$-padding + ReLU \\
   Conv layer: 64 kernels $4\times4$, stride $2$, $1$-padding + ReLU & TransposeConv layer: 64 kernels $4\times4$, stride $2$, $1$-padding + ReLU \\
   Conv layer: 64 kernels $4\times4$, stride $2$, $1$-padding + ReLU & TransposeConv layer: 64 kernels $4\times4$, stride $2$, $1$-padding + ReLU \\
   Conv layer: 64 kernels $4\times4$, stride $2$, $1$-padding + ReLU & TransposeConv layer: 64 kernels $4\times4$, stride $2$, $1$-padding + ReLU \\
   Conv layer: 64 kernels $4\times4$, stride $2$, $1$-padding + ReLU & TransposeConv layer: 64 kernels $4\times4$, stride $2$, $1$-padding + ReLU \\
   FC layers : 512+ ReLU, 512+ ReLU, FC: $2\times16$ & TransposeConv layer: 1 kernels $4\times4$, stride $2$, $1$-padding \\
\end{tabular}}
    \vspace{0.4cm}
	\caption{VAE architecture used for \textsc{IMGEP-VAE}.}
\label{sm:tab:monolithicVAE_architecture}
\end{table}

\paragraph*{IMGEP-HOLMES} For the IMGEP-HOLMES variant, the hierarchical representation starts with a single root module $R_0$ at the beginning of exploration. During each training stage, one node is \textit{split} if it meets the following conditions: \begin{itemize}
    \item the reconstruction loss for that node reaches a plateau (running average over the last 50 training epochs is below $\epsilon=20$)
    \item at least 500 patterns populate the node
    \item the node has not just been created (must has been trained for at least 200 epochs)
    \item it is not too early in the exploration loop (there must be at least 2000 patterns are explored)
    \item the total number of nodes in the hierarchy is below the maximum number allowed (we stop the expansion after 11 splits i.e. 23 modules)
\end{itemize}
Each time a split is triggered in a BC space node of the hierarchy $BC_i$, the boundary $\mathcal{B}_i$ is fitted in the latent space as follows: K-Means algorithm with 2 clusters is ran on the patterns that currently populate the node. The resulting clusters are kept fixed for the rest of the exploration, therefore when a pattern is projected in the split node, it is sent to the left children if it belongs to the first cluster on the latent space and to the right children otherwise. \\
For the IMGEP-HOLMES variant, the final hierarchy has a total of 23 VAE modules. The architecture is  identical for each module and is detailed in table~\ref{sm:tab:holmesVAE_architecture}. At the end of exploration, HOLMES has a total neural capacity of 2085981 parameters. Each base module VAE has a capacity of 86225 parameters and connections of 4673 parameters ($2085981 = 23 \times 86225 + 22 \times 4673$).

\begin{table}[h!]
	\centering
    \resizebox*{0.9\textwidth}{!}{%
    \setlength\tabcolsep{2pt}
	\begin{tabular}{ll}
   & \\
   \multicolumn{2}{l}{\textbf{Encoder}}  \\
   \cmidrule(lr){1-2} 
   Input pattern A: $256\times256\times1$ &  \\
   Conv layer: 16 kernels $4\times4$, stride $2$, $1$-padding + ReLU & \\
   Conv layer: 16 kernels $4\times4$, stride $2$, $1$-padding + ReLU &  \\
   Conv layer: 16 kernels $4\times4$, stride $2$, $1$-padding + ReLU & \textbf{lf\_c:} 16 kernels $1\times1$, stride $1$, $1$-padding \\
   Conv layer: 16 kernels $4\times4$, stride $2$, $1$-padding + ReLU &  \\
   Conv layer: 16 kernels $4\times4$, stride $2$, $1$-padding + ReLU & \\
   Conv layer: 16 kernels $4\times4$, stride $2$, $1$-padding + ReLU &  \\
   FC layers : 64+ ReLU, 64+ ReLU, FC: $2\times16$ & \\

   & \\
   \multicolumn{2}{l}{\textbf{Decoder}} \\
   \cmidrule(lr){1-2} 
    Input latent vector z: $16\times1$ & \\
    FC layers : 64+ ReLU, & \textbf{gfi\_c:} 64+ReLU \\
    FC layers: 64+ ReLU, $4\times4\times16$ + ReLU \\
    TransposeConv layer: 16 kernels $4\times4$, stride $2$, $1$-padding + ReLU & \\
    TransposeConv layer: 16 kernels $4\times4$, stride $2$, $1$-padding + ReLU & \\
    TransposeConv layer: 16 kernels $4\times4$, stride $2$, $1$-padding + ReLU & \textbf{lfi\_c:}  16 kernels $1\times1$, stride $1$, $1$-padding \\
    TransposeConv layer: 16 kernels $4\times4$, stride $2$, $1$-padding + ReLU & \\
    TransposeConv layer: 16 kernels $4\times4$, stride $2$, $1$-padding + ReLU & \\
    TransposeConv layer: 1 kernel $4\times4$, stride $2$, $1$-padding & \textbf{recon\_c:} 1 kernel $1\times1$, stride $1$, $1$-padding\\

\end{tabular}}
    \vspace{0.4cm}
	\caption{Module architecture used for \textsc{IMGEP-HOLMES}. All the modules $R_i$ have this architecture for the base VAE network as well as the connections (except $R_0$ which does not have the connections). }
\label{sm:tab:holmesVAE_architecture}
\end{table}

\clearpage
\section{Additional Results}
\label{sm:sec:additional_results}

\subsection{RSA complete temporal analysis and statistics}
\label{sm:subsec:RSA_results}

\begin{figure}[b!]
\centering
 \includegraphics[width=\linewidth]{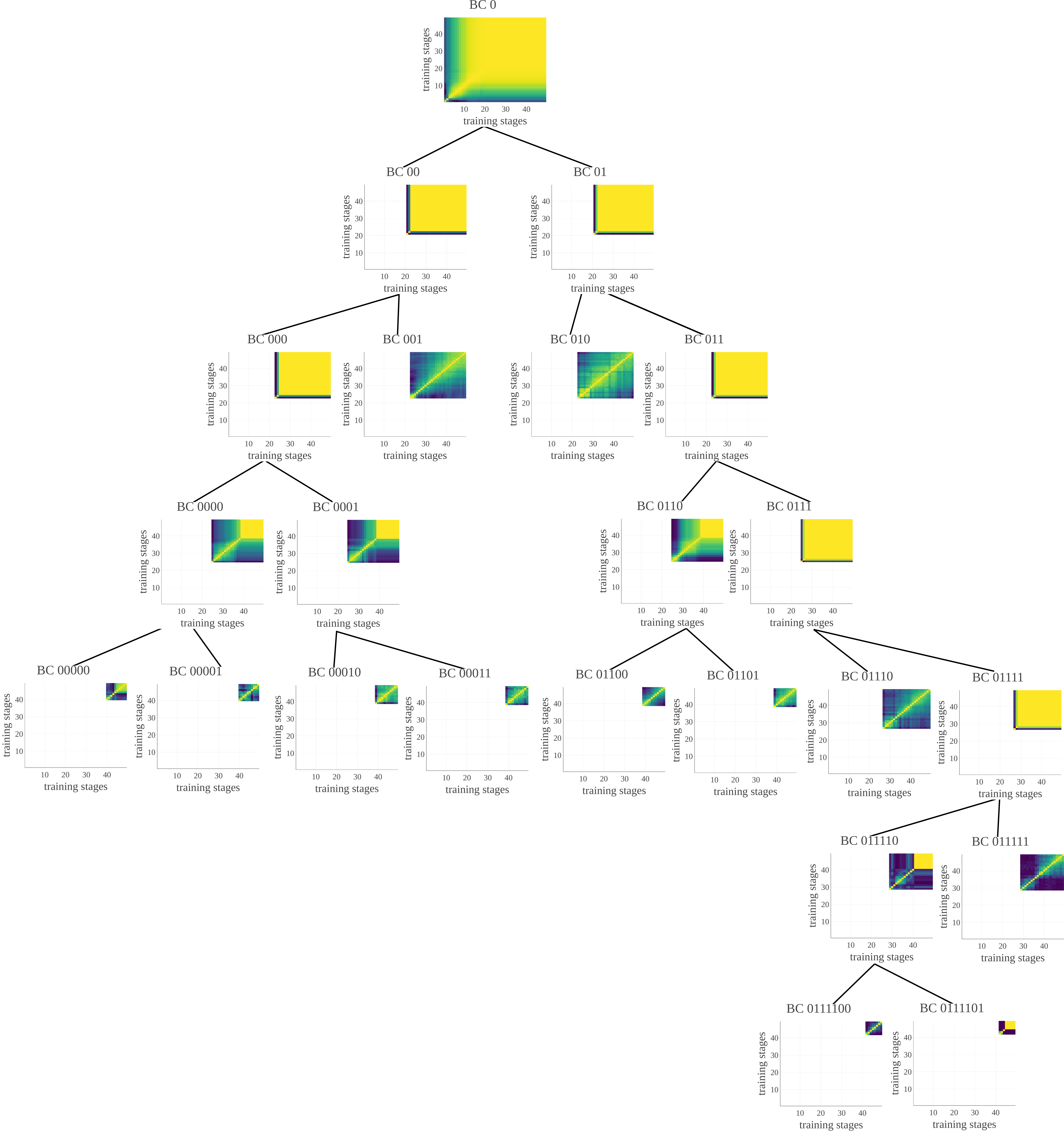} 
 \caption{Example of a hierarchy of behavioral characterization spaces learned by HOLMES. It starts with a root node (BC 0, top) and iteratively splits the learned latent spaces, resulting in a tree structure (with leaf nodes at the bottom). In each node, we display the RSA similarity index between 0 (dark blue, not similar at all) and 1 (yellow, identical), where representations are compared in time between the different training stages.}
\label{sm:fig:rsa-hierarchy}
\end{figure}

Figure~\ref{sm:fig:rsa-hierarchy} shows how HOLMES is able to progressively builds a hierarchy of behavioral characterization spaces from the incoming data. The data is here collected by the IMGEP-HOLMES algorithm, with 50 training stages occurring each 100 steps. At start (stage 0), the hierarchy contains only the root node at the top of the figure (BC 0). Node saturation occurs at training stages where the RSA similarity index between the representation at that stage and the representations at all subsequent stages is high (yellow). For example, we see on the figure that the root node saturates after approximately 15-20 training stages. When a node saturates, HOLMES splits it in two child nodes (see section~\ref{sm:subsec:HOLMES} for details on the splitting procedure). For example, the root node BC 0 is split into the child nodes BC 00 and BC 01 at training stage 21, as indicated by the fact that the RSA plots of these child nodes start at that stage. When a node is split, the parent node is frozen and learning only continues in leaf nodes (as indicated by the RSA indexes of a parent node being all at 1 after a split). We observe that some child nodes saturate much more quickly than others. For example, node BC 000 saturates only a few training stages after its split from BC 00, while its sibling BC 001 never saturates until the end of the training at stage 50. The RSA analysis of node BC 001 indeed shows that the learned representation continues to evolve as training occurs. This means that this node corresponds to a part of the BC space constituting a rich progress niche for the base module VAE associated with that node. In contrast, node BC 000 will require further splitting to discover such progress niches in its child nodes. The reader can refer to Figure \ref{sm:fig:non-guided-holmes} in section \ref{sm:sec:additional_visualisations} for visualizing discovered patterns in each nodes of the hierarchy.

\begin{figure}[h!]
\sidebysidecaption{0.7\linewidth}{0.25\linewidth}{%
    \centering
    \setlength\tabcolsep{1pt}
    \renewcommand{\arraystretch}{0.5}
    \def\imagetop#1{\vtop{\null\hbox{#1}}}
    \begin{tabular}[t]{cc}
    VAE &  HOLMES\\
     \midrule
    \makecell{\includegraphics[height=3.7cm]{./figures/VAE_temporal_RSA_matrix.pdf} \\ \includegraphics[height=3.2cm]{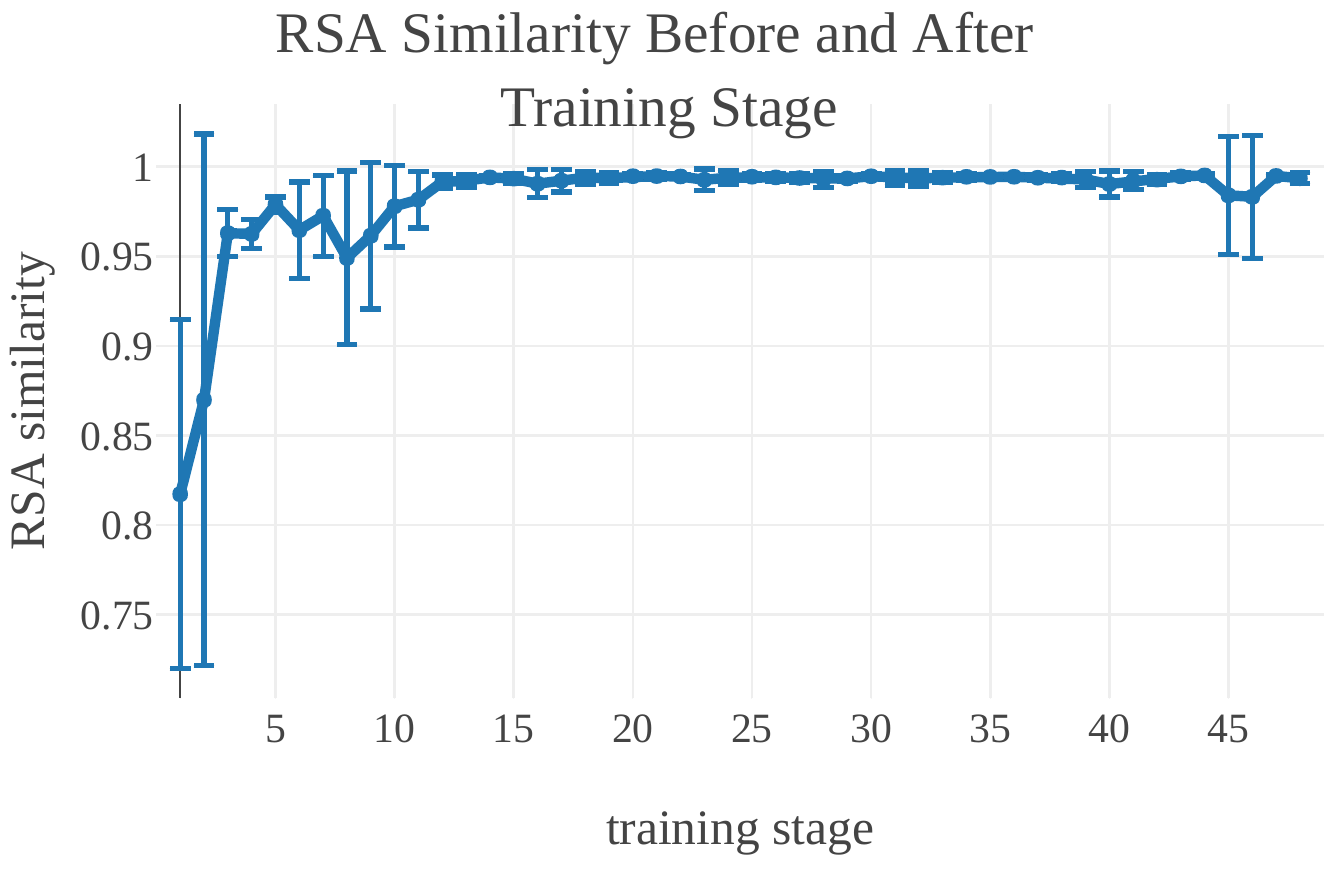}}
    & 
    \makecell{\includegraphics[height=4.2cm]{./figures/HOLMES-VAE_goalspaces_RSA_matrix.pdf} \\  \includegraphics[height=2.8cm]{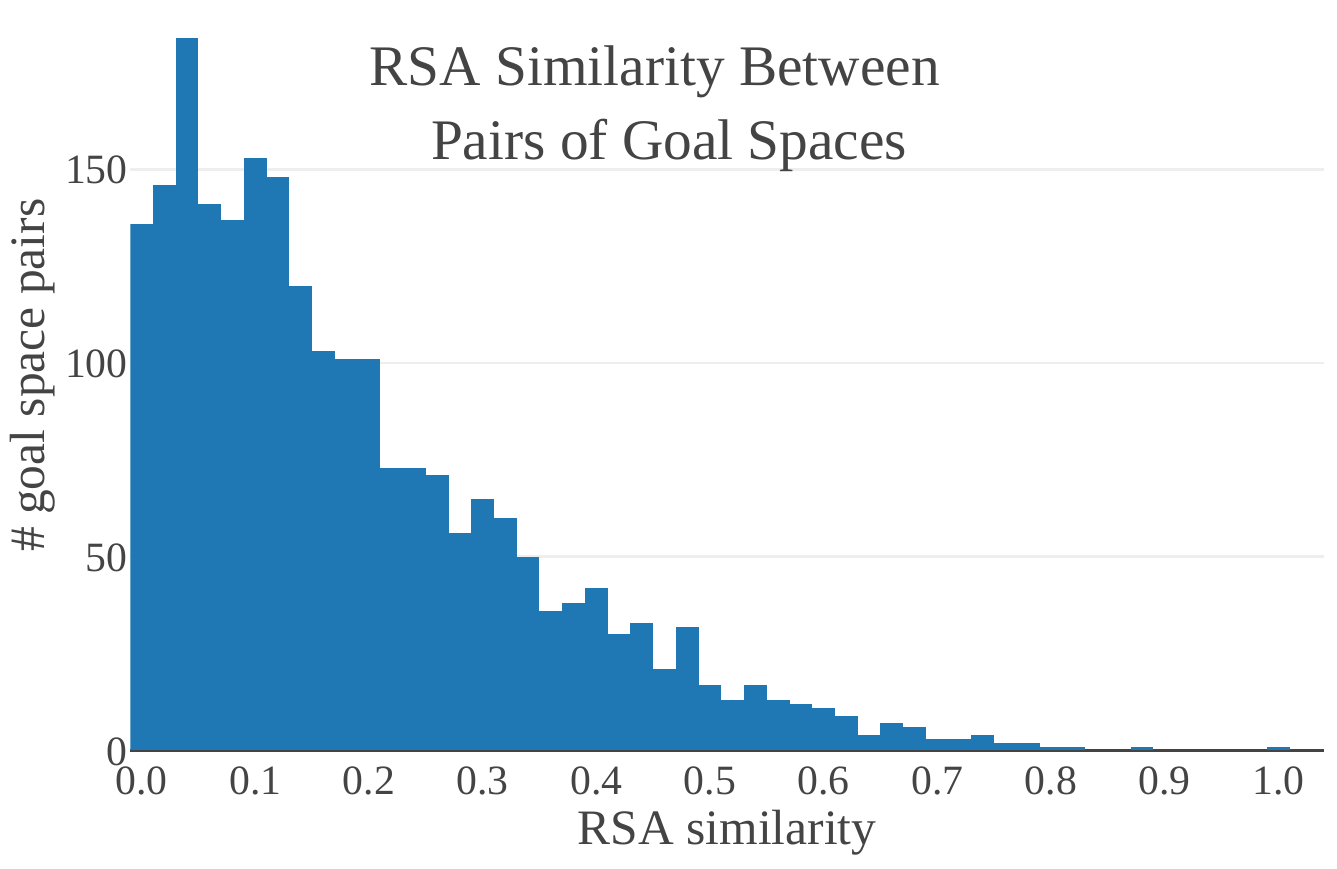}} \\
    \end{tabular}%
}{%
  \caption{RSA similarity index between 0 (dark blue, not similar at all) and 1 (yellow, identical).  (Top) RSA matrix for one experiment as shown in Figure 3 of main paper. (Bottom) statistics over the 10 repetitions: (bottom-VAE) RSA index similarity between representations coming from two consecutive training stages (mean and std); (bottom-HOLMES): histogram of RSA index similarity between all pairs of modules in HOLMES (aggregated over the 10 repetitions). }
\label{sm:fig:rsa-statistics}
}
\vspace{-10pt}
\end{figure}

Figure~\ref{sm:fig:rsa-statistics} complements Figure 3 of the main paper with statistical results over 10 repetitions. The statistical results  (bottom row) confirm our analysis: the VAE representation saturates quite early in the exploration loop and the representations learned by HOLMES modules are dissimilar from one module to another. Indeed we can see that the VAE representations of all experiments saturate after 15 training stages (high RSA $\approx 1$ between remaining consecutive training stages). The histogram of similarity index between all pairs of modules in HOLMES (for all experiments) show a concentration between [0,0.3] (i.e. very low similarity).

\subsection{Additional IMGEP baselines with a monolithic BC space }
\label{sm:subsec:monolithic_baselines}

\begin{figure}[h!]
\hspace*{-1.8cm}
\setlength\tabcolsep{1pt}
\begin{tabular}{>{\centering\arraybackslash}m{2cm} >{\centering\arraybackslash}m{3cm} >{\centering\arraybackslash}m{5cm} >{\centering\arraybackslash}m{5cm}}
 & \textbf{RSA Analysis}  & \makecell{\textbf{Diversity of SLP} \\ \textbf{in \textsc{BC\textsubscript{Elliptical-Fourier}}}}  & \makecell{\textbf{Diversity of TLP} \\ \textbf{in \textsc{BC\textsubscript{Lenia-Statistics}}}} \\

  \textbf{IMGEP-BetaVAE}  &
 \includegraphics[height=3cm]{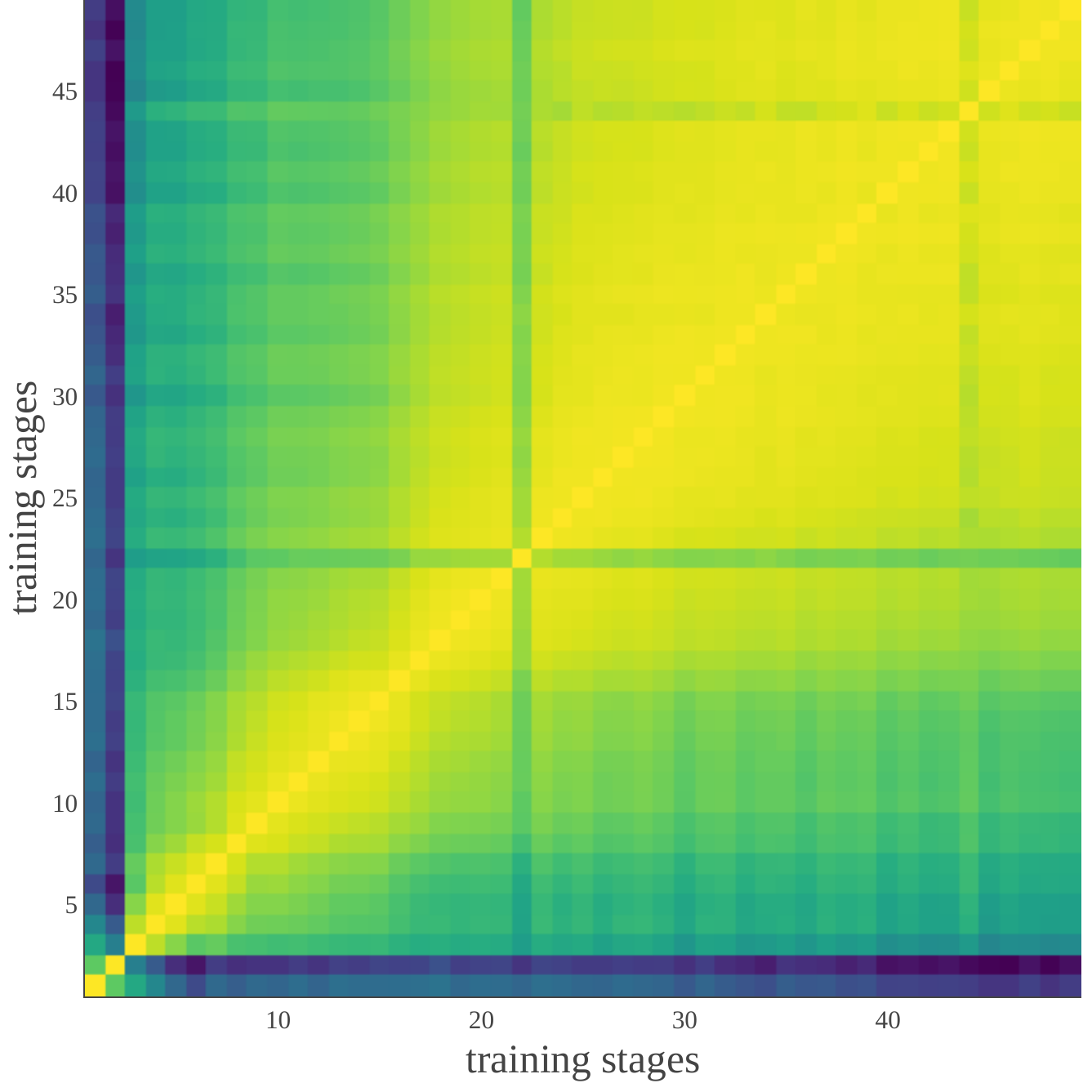}
 & \includegraphics[height=3.5cm]{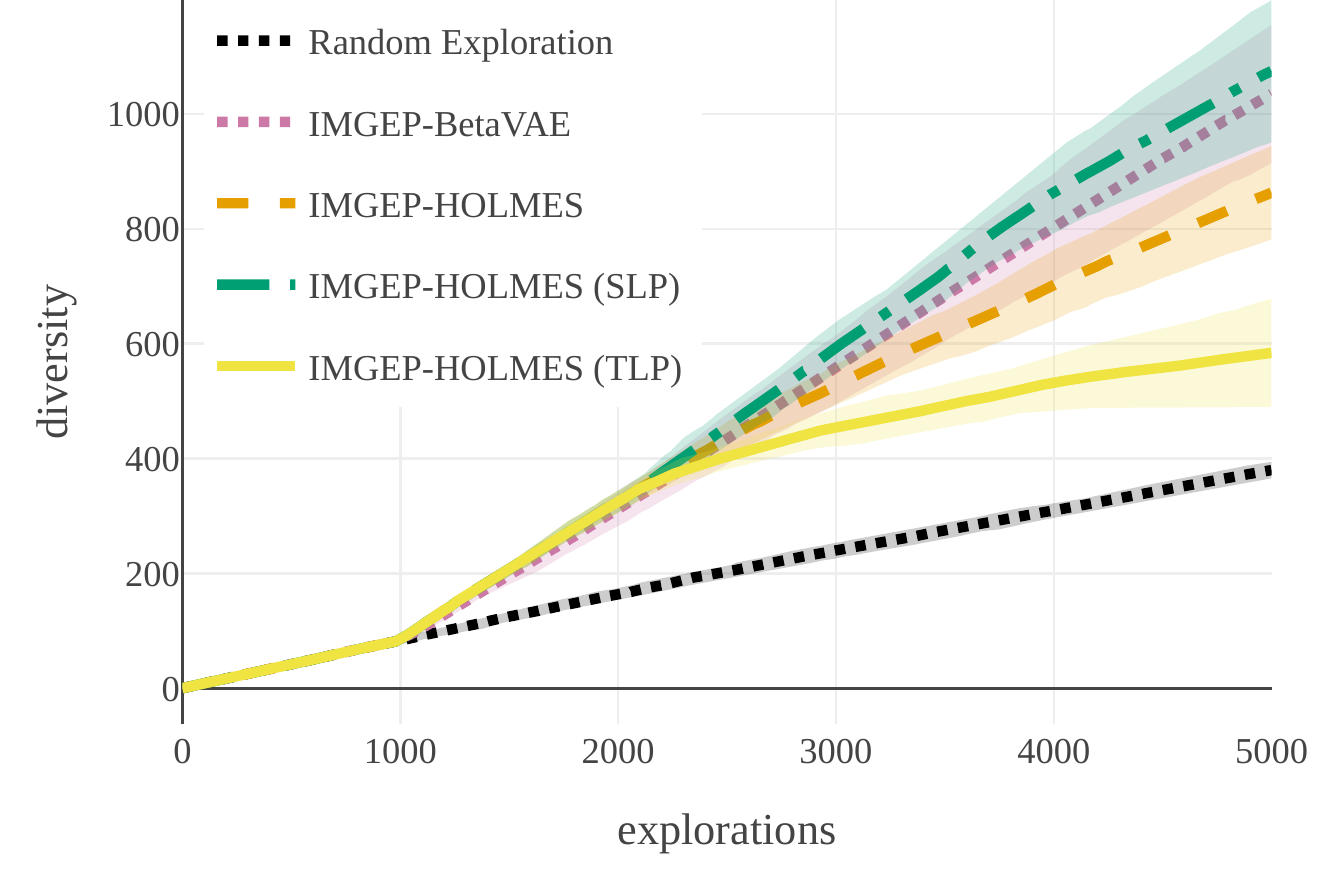}  
 & \includegraphics[height=3.5cm]{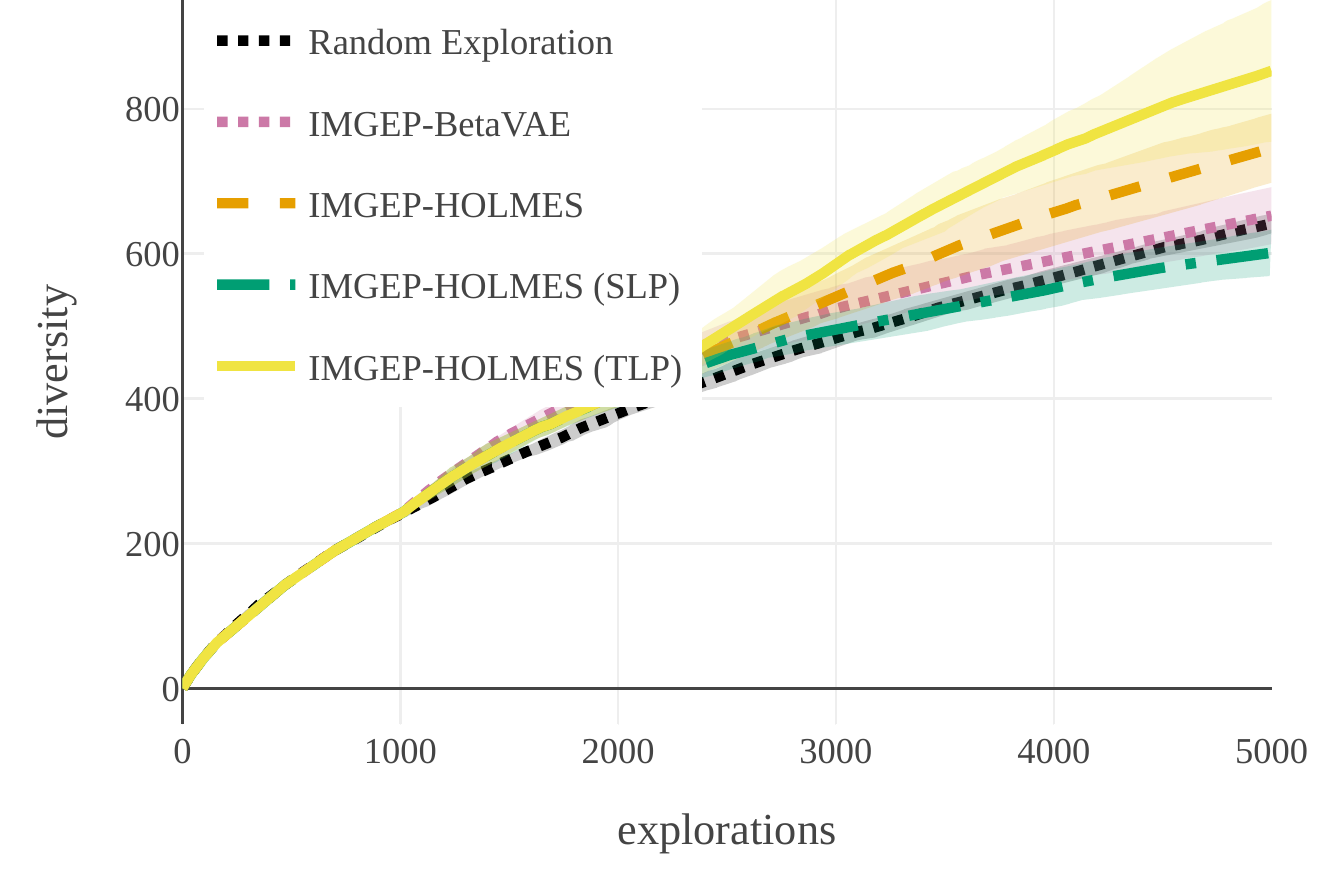}  \\

  \textbf{IMGEP-BetaTCVAE}  &
 \includegraphics[height=3cm]{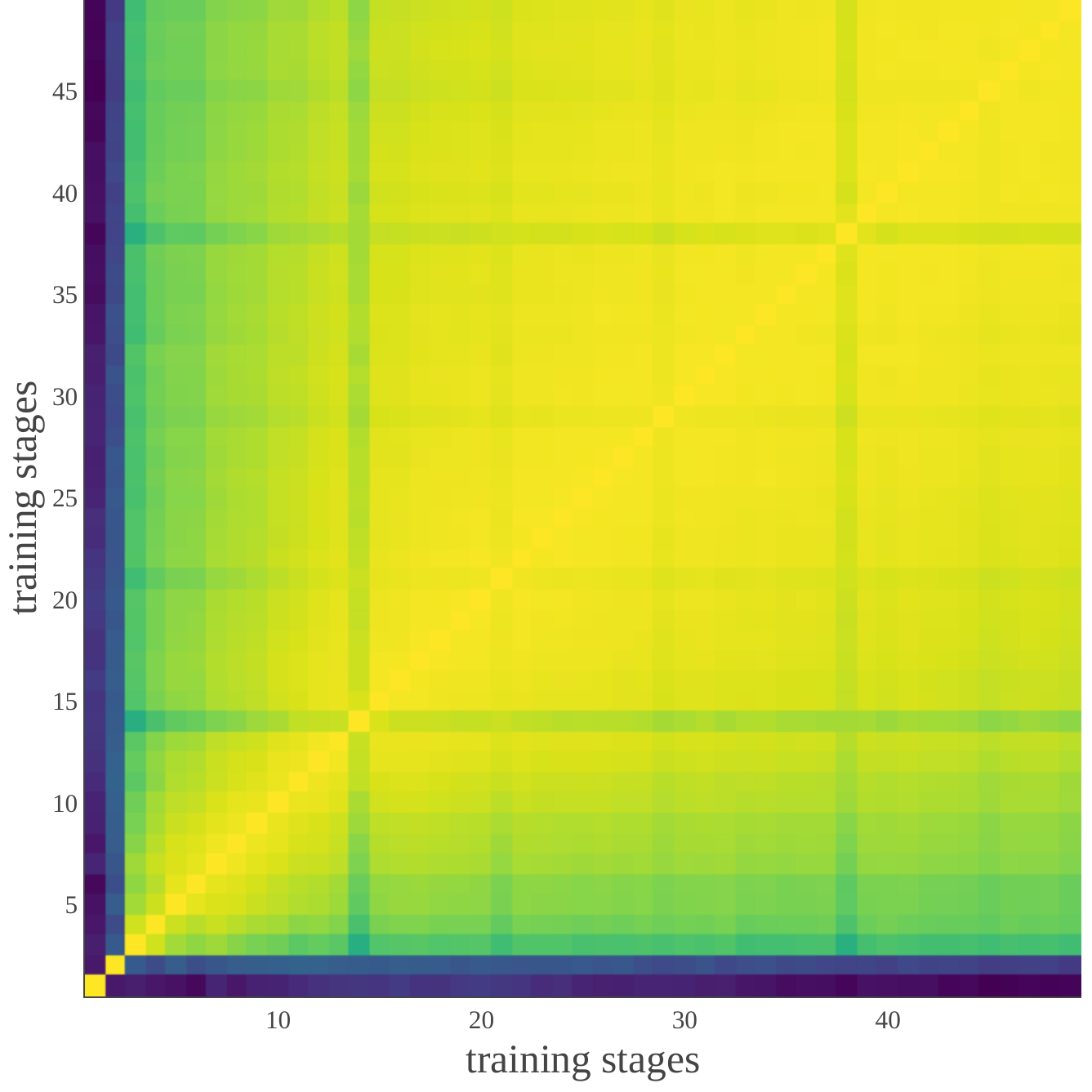}
 & \includegraphics[height=3.5cm]{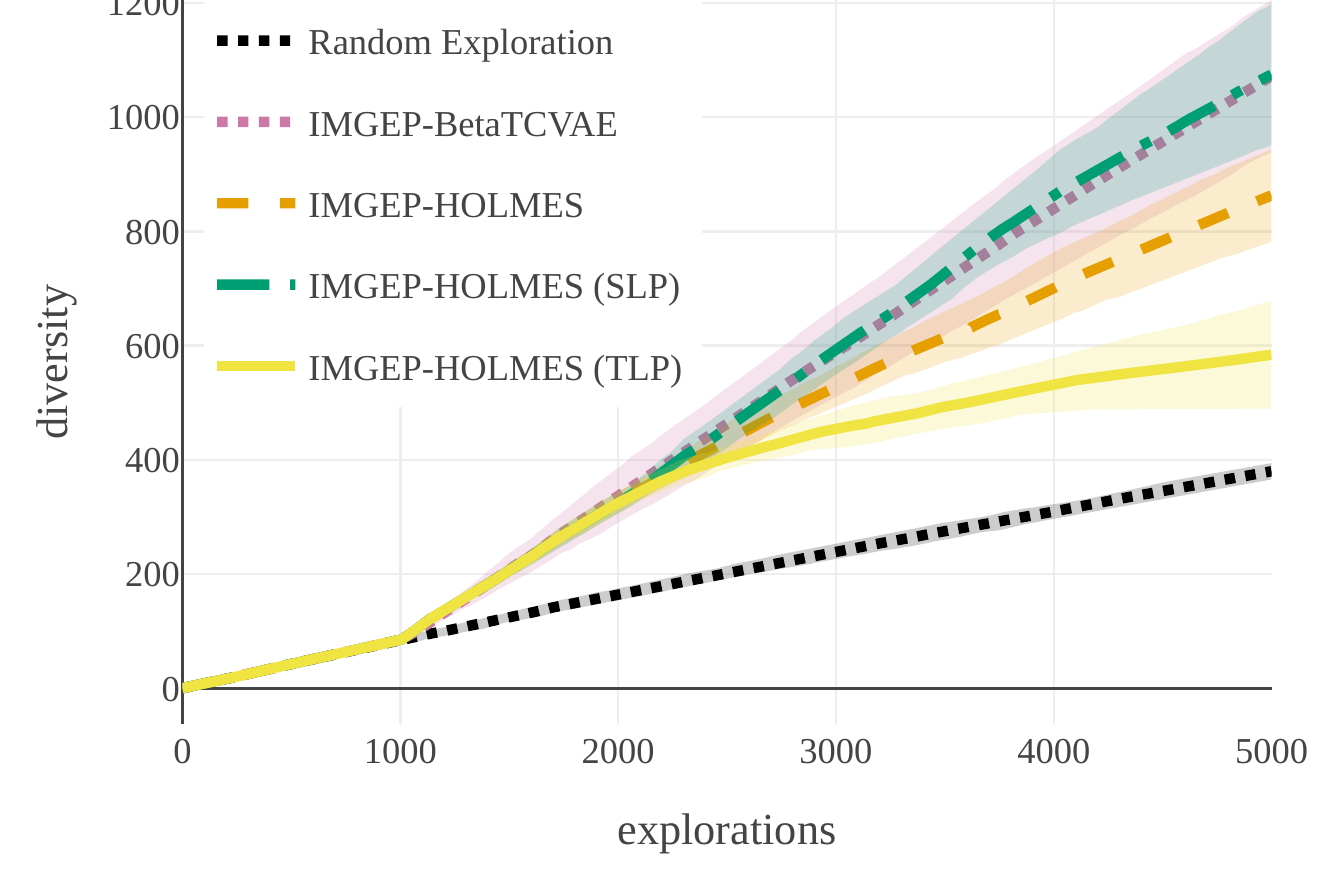}  
 & \includegraphics[height=3.5cm]{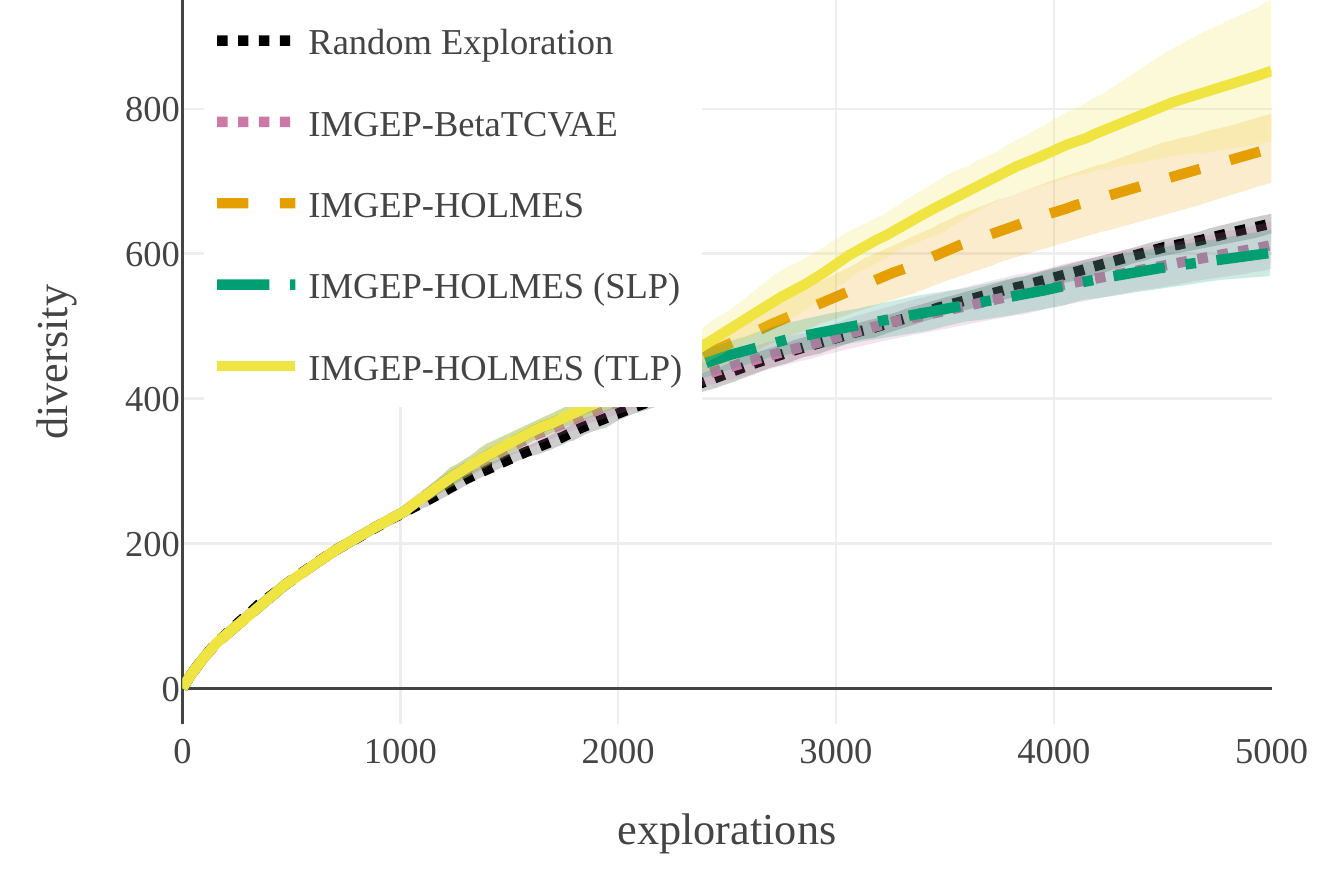}  \\

  \textbf{IMGEP-TripletCLR}  &
 \includegraphics[height=3cm]{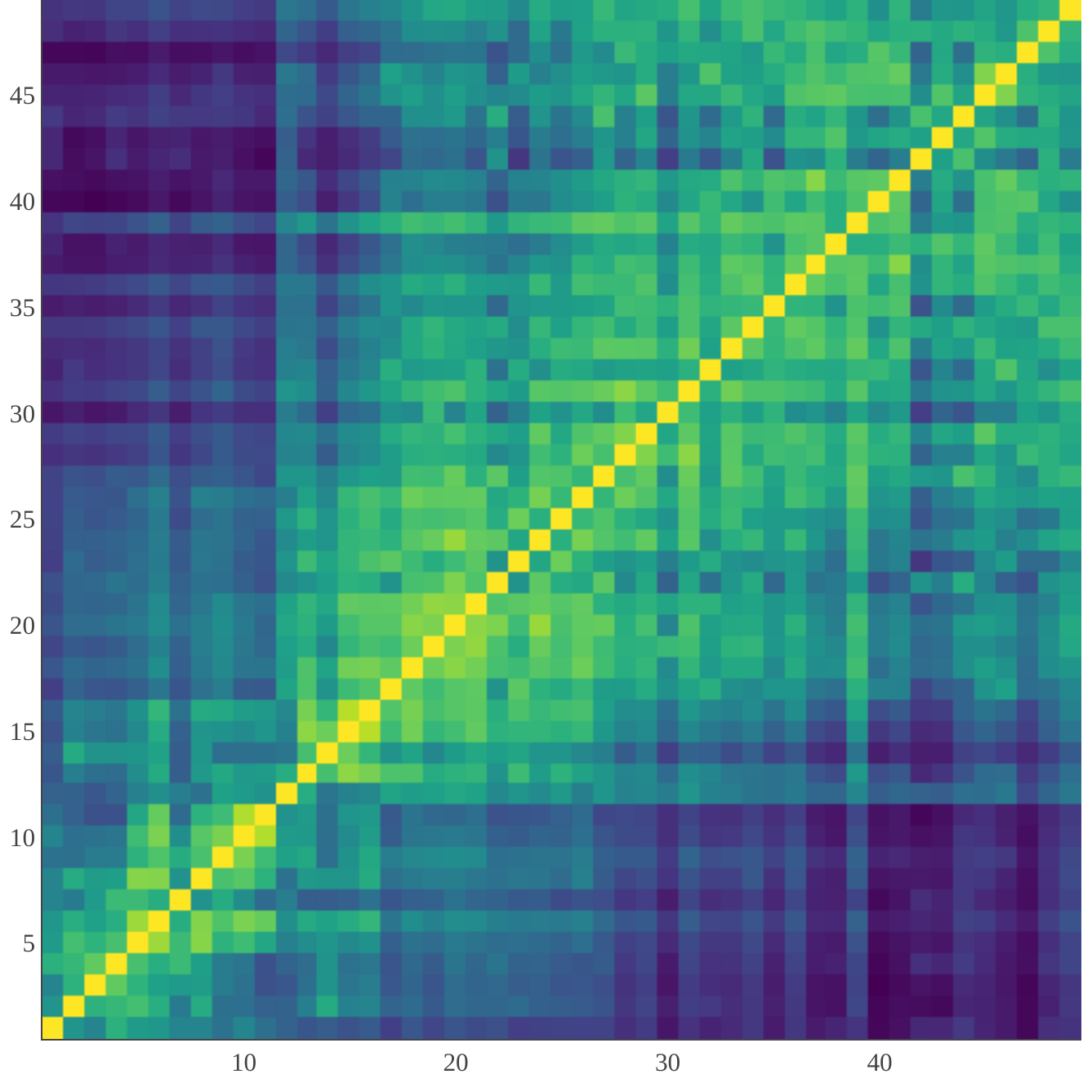}
 & \includegraphics[height=3.5cm]{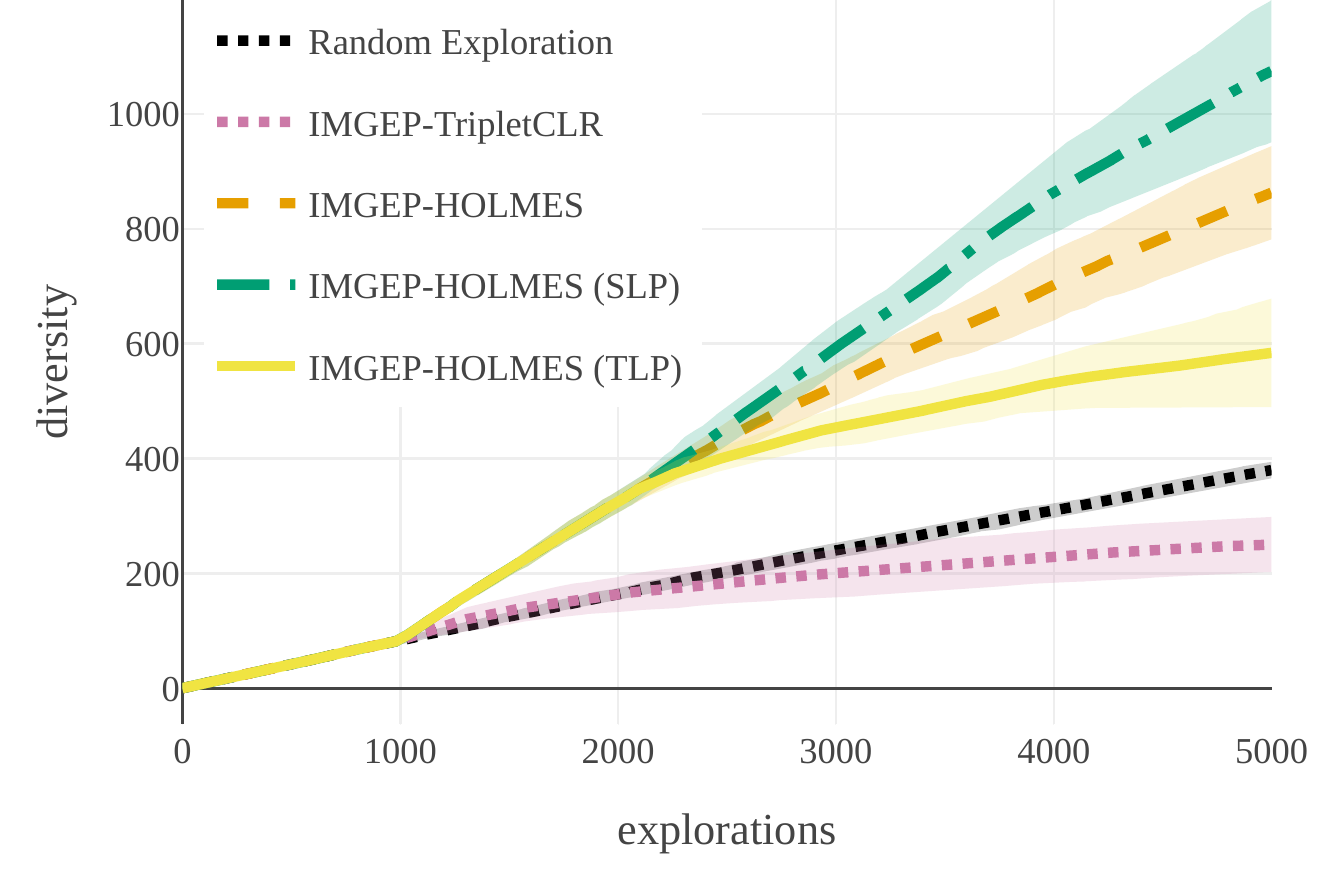}  
 & \includegraphics[height=3.5cm]{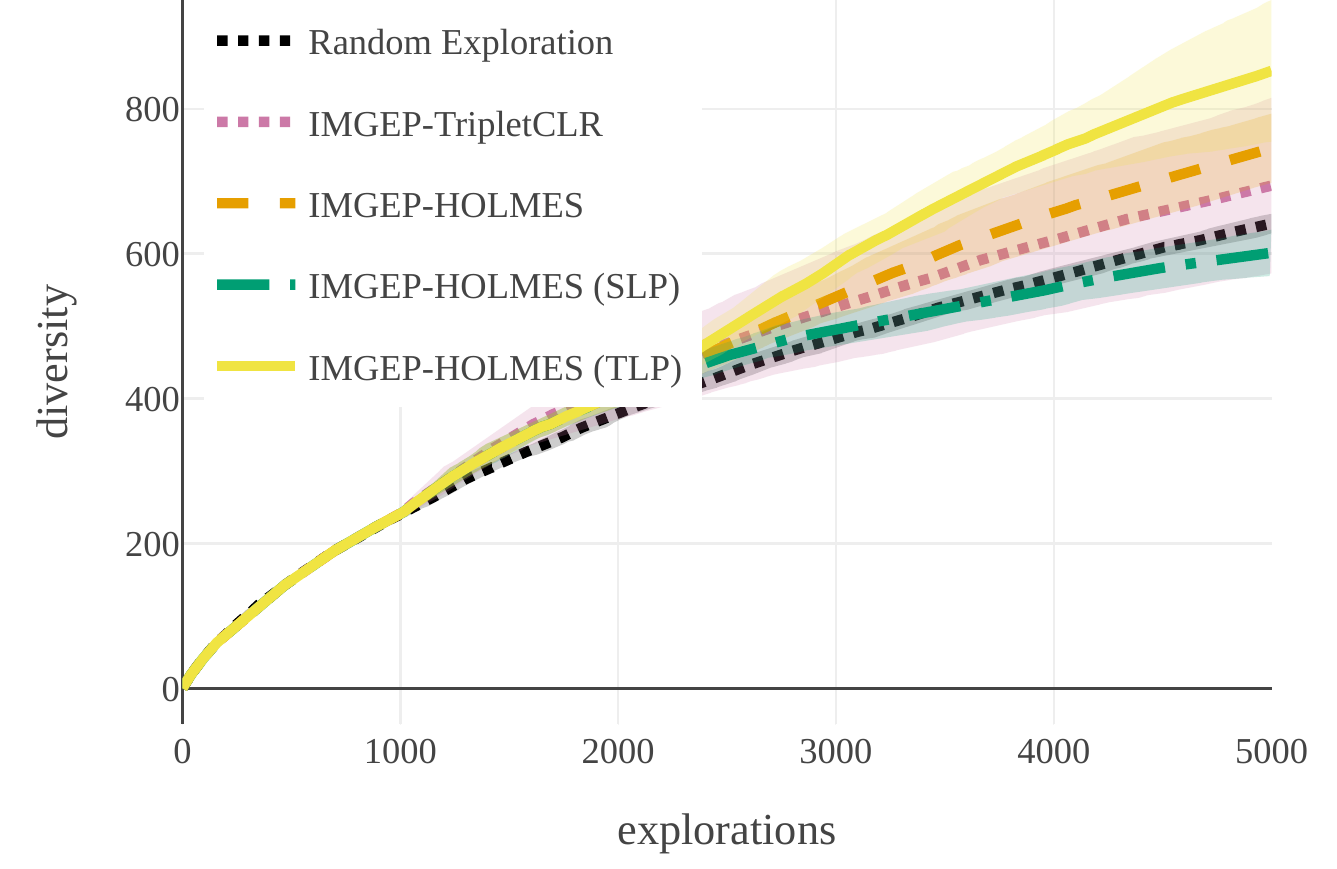}  \\

  \textbf{IMGEP-SimCLR}  &
 \includegraphics[height=3cm]{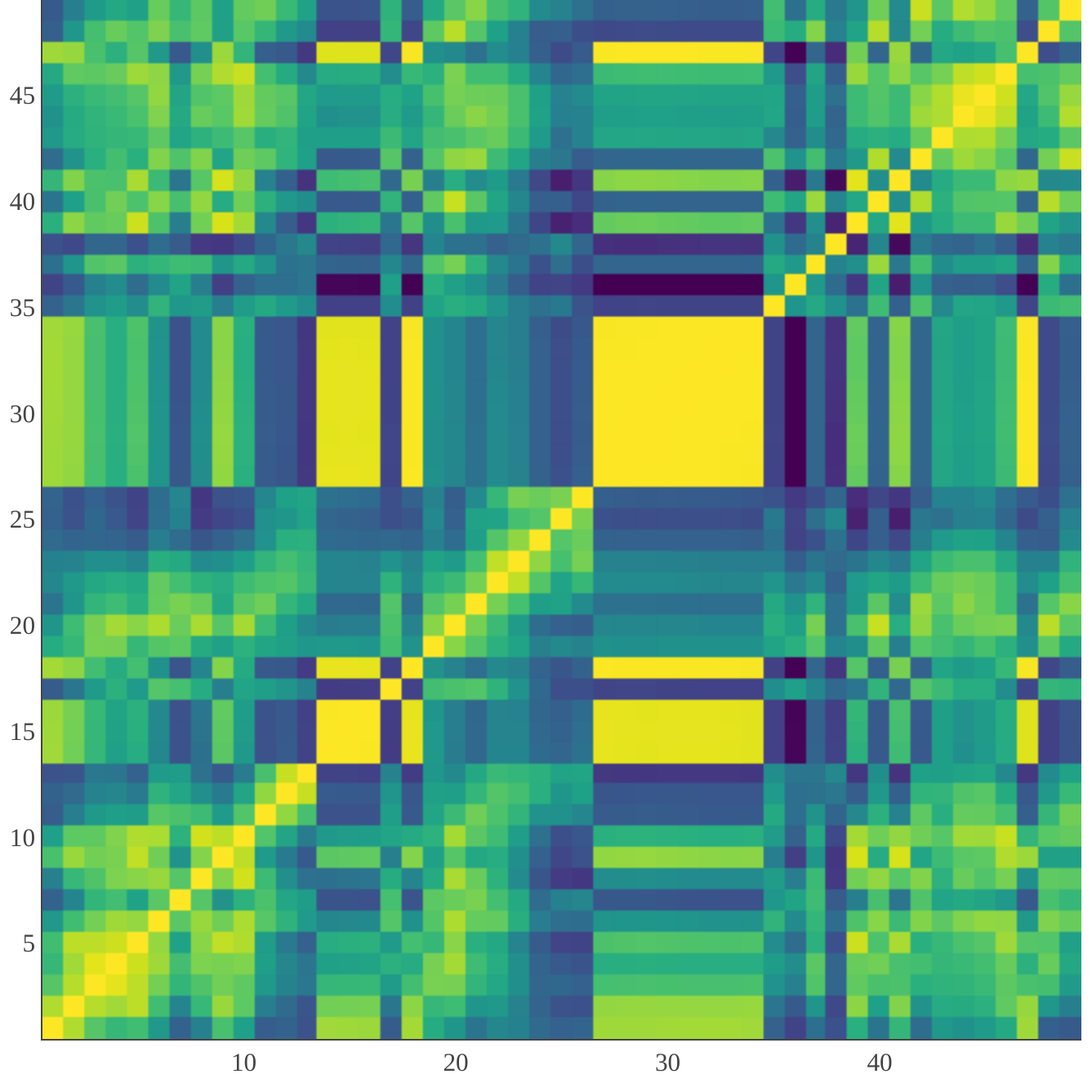}
 & \includegraphics[height=3.5cm]{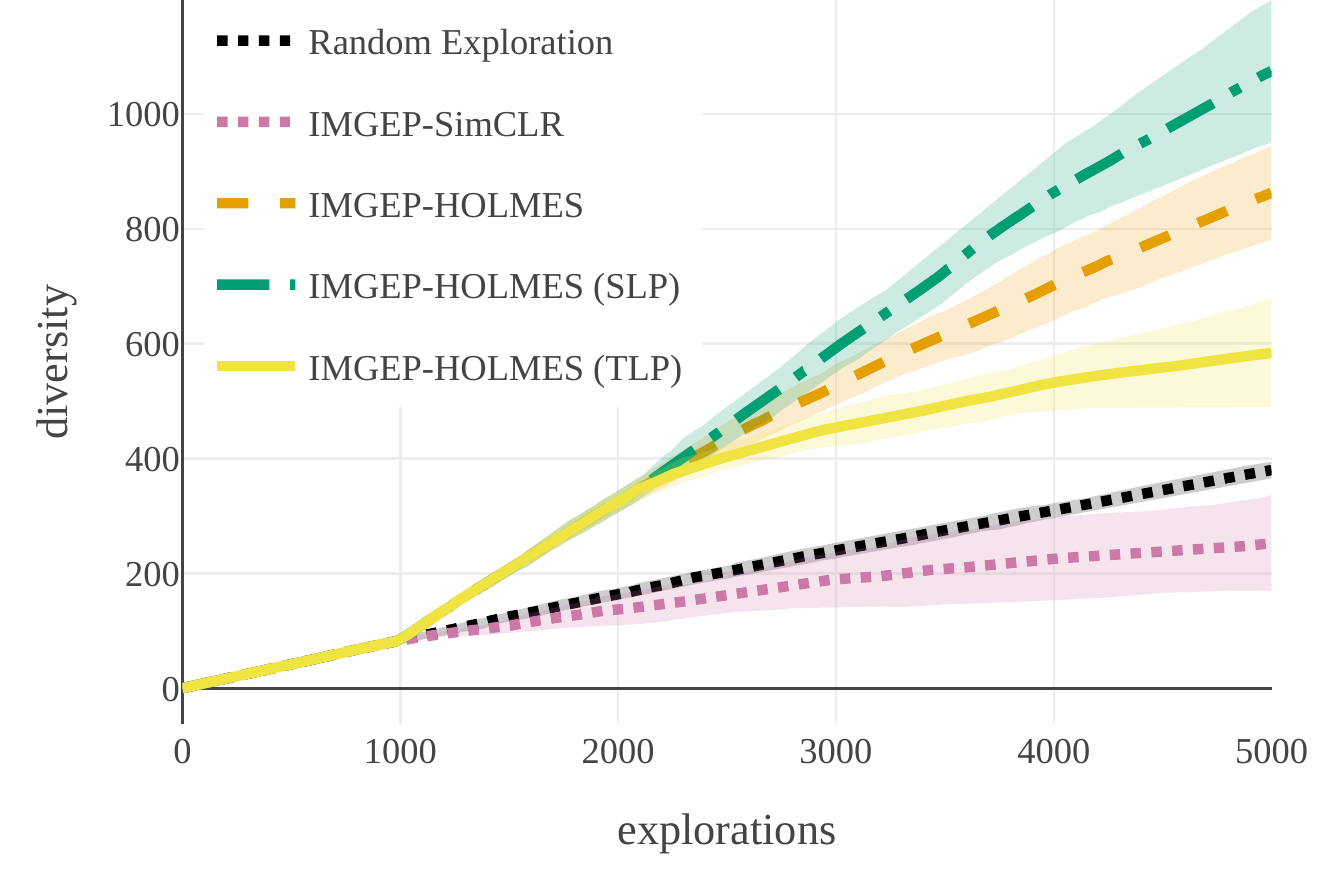}  
 & \includegraphics[height=3.5cm]{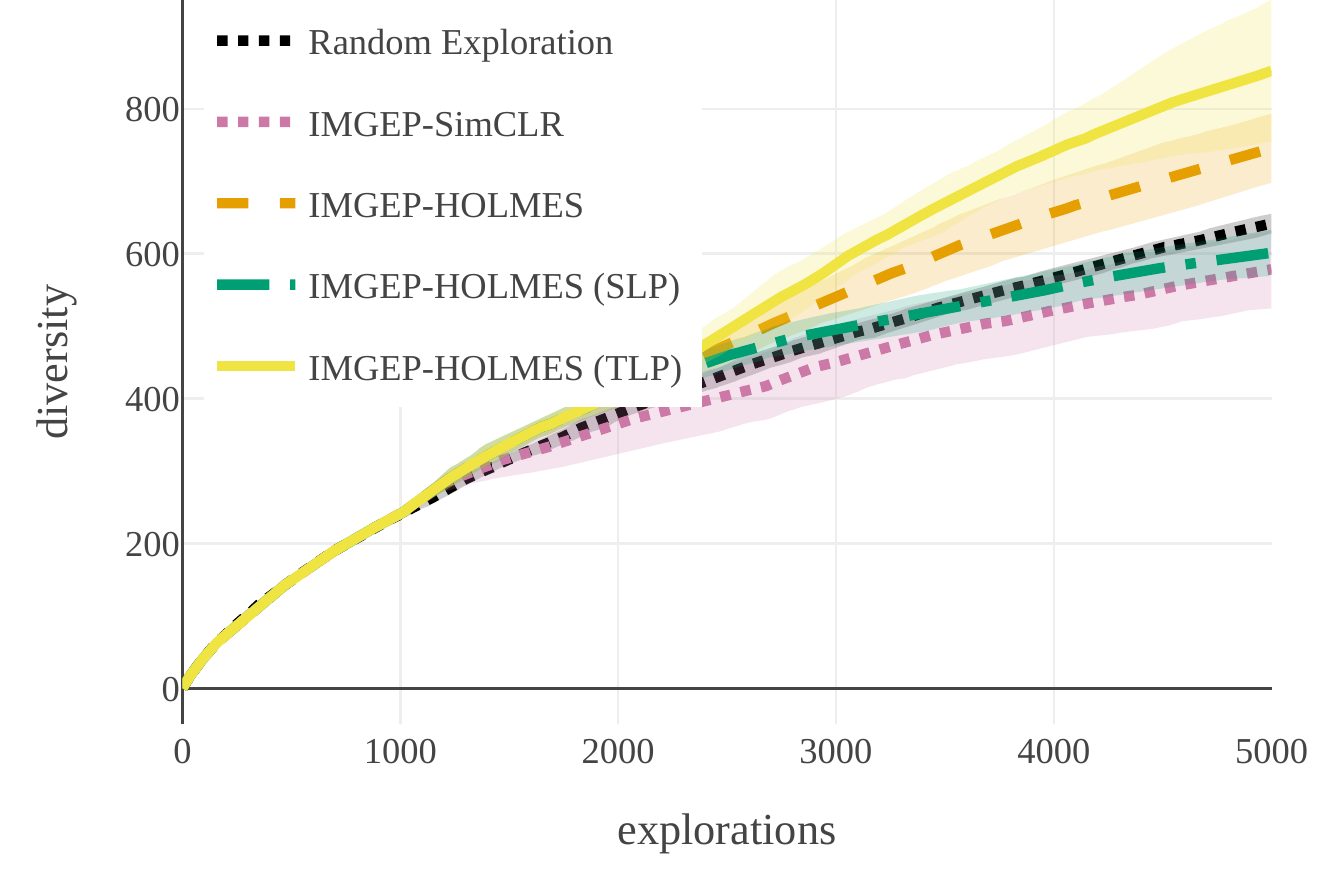}  \\

  \textbf{IMGEP-BigVAE}  &
 \includegraphics[height=3cm]{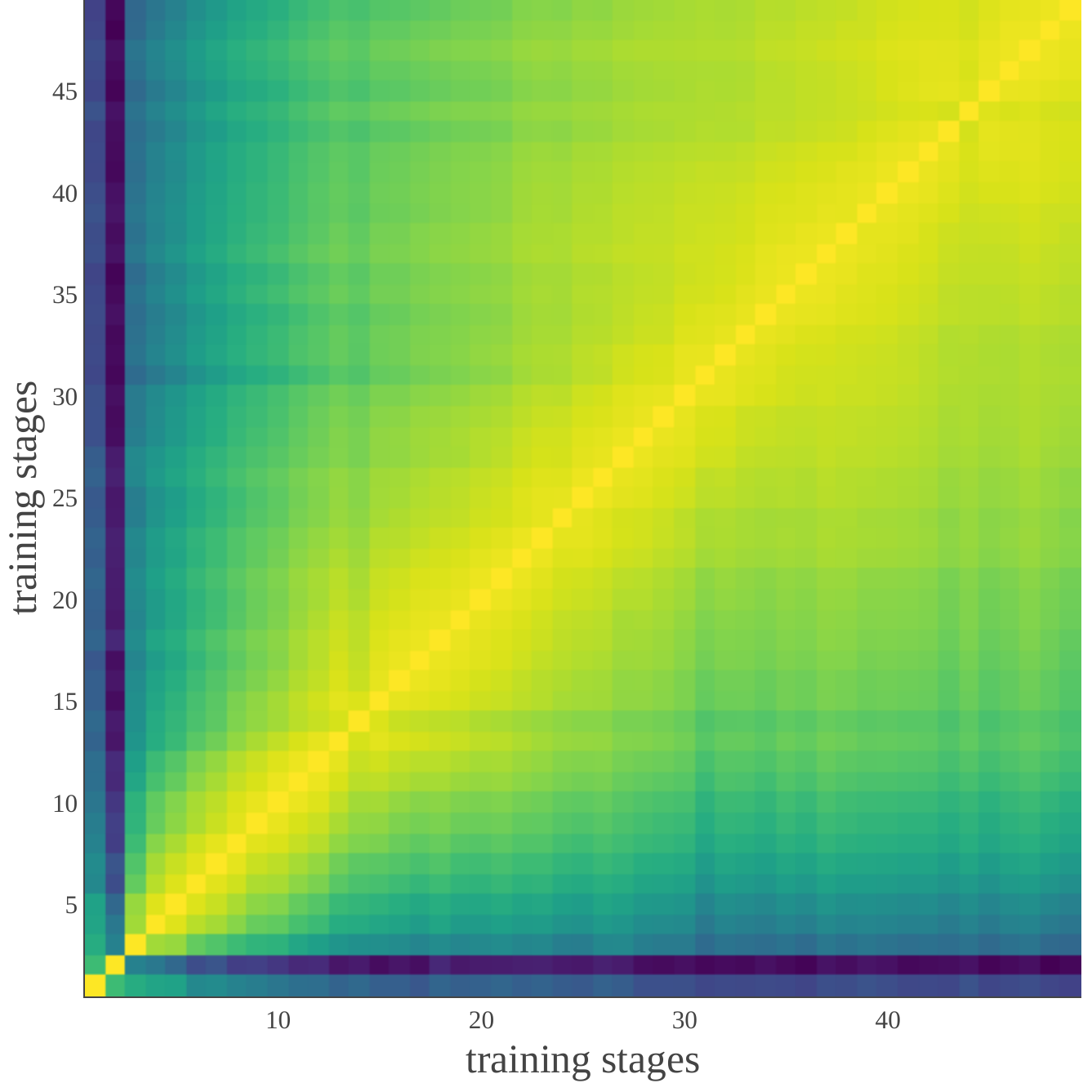}
 & \includegraphics[height=3.5cm]{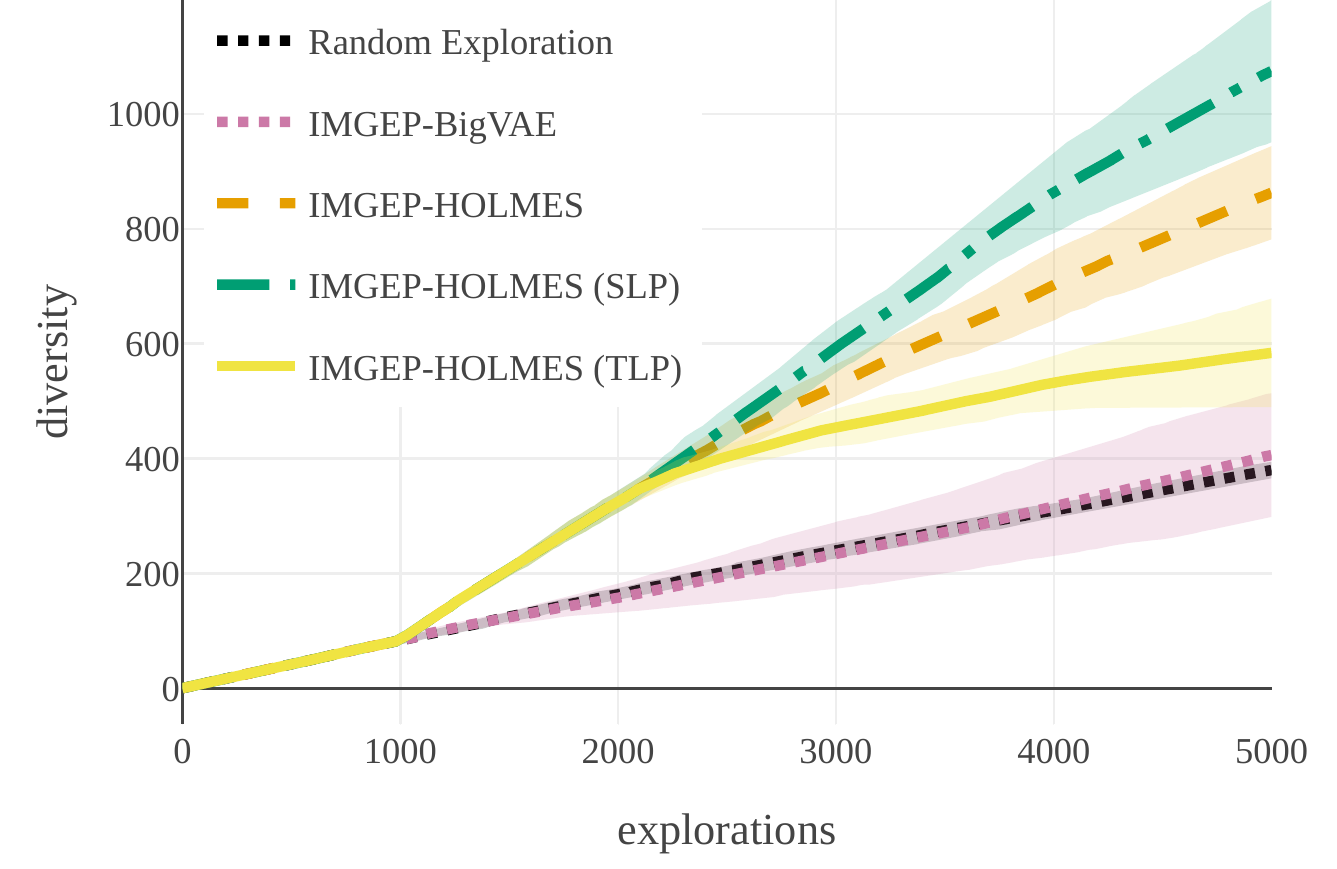}  
 & \includegraphics[height=3.5cm]{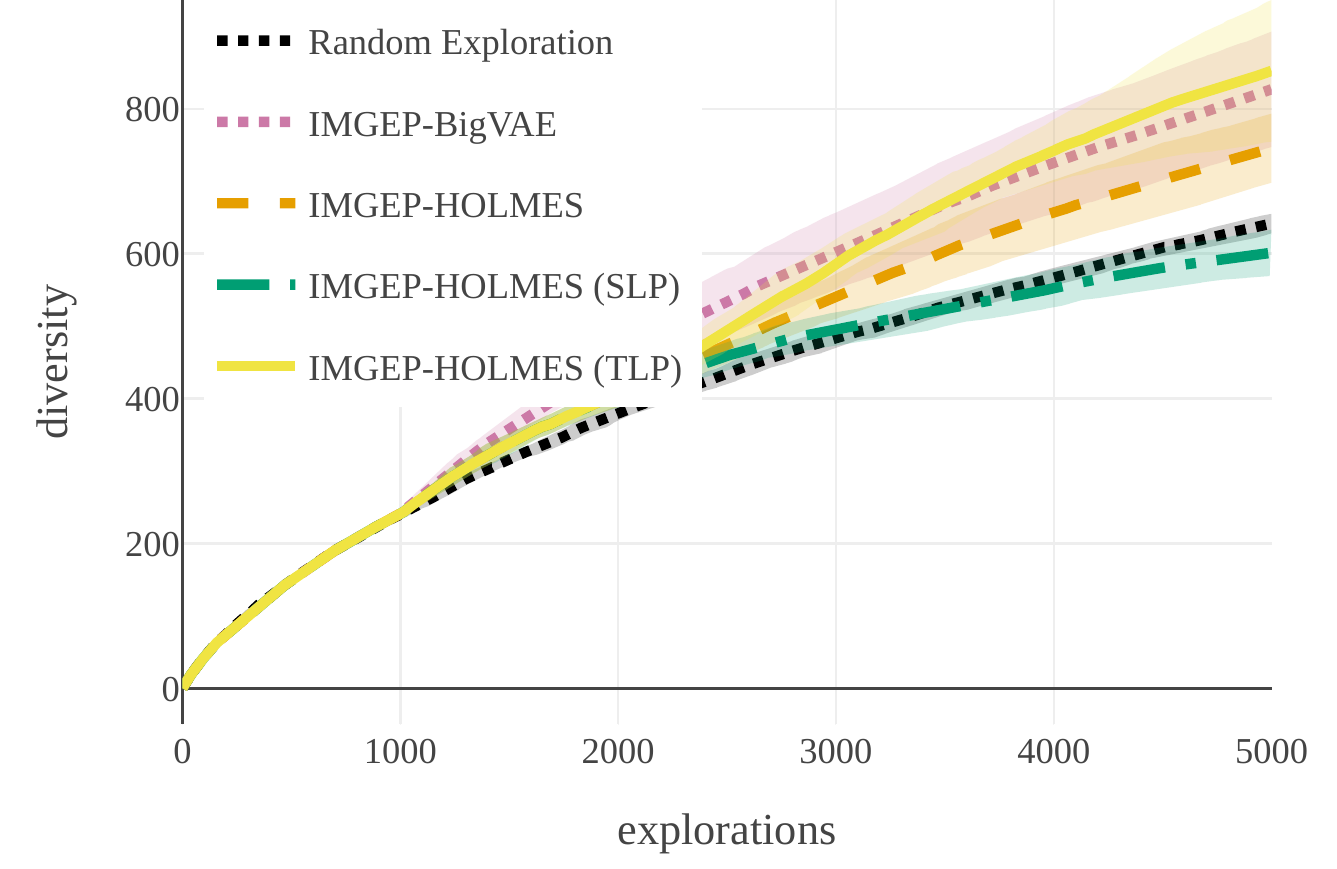}  \\

\end{tabular}
 \caption{This figure complements the results presented in the main paper, where we replace the baseline with the monolithic BC space (IMGEP-VAE) with different architectures and training strategies. Each row is a baseline denoted as IMGEP-X (where X represents the training strategy used for training the monolithic representation). For each row, we display: (left) RSA matrix where representations are compared in time between the different training stages, as shown in Figure 3 of the main paper; (middle-right) exact same plots than Figure 5 of the main paper where we replace the monolithic IMGEP-VAE baseline (pink curve) by the other baseline (of the current row). Therefore only the pink curve vary between the different graphs of one column. }
\label{sm:fig:monolithic_baselines}
\end{figure}

In section~\ref{subsec:results_incrementalBC} of the main paper, we compared the incremental training of behavioral training between an IMGEP equipped with a monolithic VAE (IMGEP-VAE) and an IMGEP equipped with the hierarchy of VAEs (IMGEP-HOLMES). 

\paragraph*{Baselines}

In this section, we consider different baselines for the training strategy of the monolithic architecture: \textbf{BetaVAE}~\cite{burgess2018understanding}, \textbf{BetaTCVAE}~\cite{chen2018isolating}, \textbf{TripletCLR}~\cite{chechik2010large,schroff2015facenet}, \textbf{SimCLR}~\cite{chen2020simple} and \textbf{BigVAE}. All the baselines have the same encoder architecture and training procedure than the main baseline IMGEP-VAE (as detailed in section~\ref{sm:subsec:imgep_variants}). The baselines differ in their approach to train the encoder network, including several variants of variational-autoencoders and contrastive approaches.

The first two variants BetaVAE~\cite{burgess2018understanding} and BetaTCVAE~\cite{chen2018isolating} build on the VAE framework and augment the VAE objective with the aim to enhance interpretability and disentanglement of the latent variables. Therefore only the training loss of the VAE (see section~\ref{sm:subsec:HOLMES}) differs: \begin{itemize}
    \item The BetaVAE objective re-weights the $b$ term by a factor $\beta > 1$: \\
    $\mathcal{L}_{\textsc{BetaVAE}} (\theta, \phi; \textbf{x}, \textbf{r} )= \underbrace{\mathbb{E}_{\hat{p}(\textbf{x})}\left(\mathbb{E}_{q_\phi(\textbf{r}|\textbf{x})}\left( -\log p_\theta(\textbf{x}|\textbf{r}) \right)\right)}_a + \beta \times \underbrace{\mathbb{E}_{\hat{p}(\textbf{x})}\left( D_{\textsc{KL}} \left( q_\phi(\textbf{r}|\textbf{x}) || p(\textbf{r}) \right) \right)}_b $ \\
    Our baseline uses $\beta=10$.
    \item The BetaTCVAE objective augments the VAE objective with an additional regularizer that penalizes the \textit{total correlation} (dependencies between the dimensions of the representation): \\
    \begin{align*}
     \mathcal{L}_{\textsc{BetaTCVAE}} (\theta, \phi; \textbf{x}, \textbf{r}) &= \mathbb{E}_{\hat{p}(\textbf{x})}\left(\mathbb{E}_{q_\phi(\textbf{r}|\textbf{x})}\left( -\log p_\theta(\textbf{x}|\textbf{r}) \right)\right) + \\
     \alpha \times \underbrace{I_{q_\phi} (\textbf{x}|\textbf{r})}_{mutual\:information} &  + \beta \times \underbrace{TC(q_\phi(\textbf{r}))}_{total\:correlation} + \gamma \times \underbrace{\sum_j D_{KL} (q_{\phi} (z_j) || p(z_j))}_{elementwise\:KL} 
     \end{align*}
    Because TC is not tractable, they~\cite{chen2018isolating} propose two methods based on importance sampling to estimate it: \textit{minibatch weighted sampling} (mws) and \textit{minibatch stratified sampling (mss)}. Our baselines uses $\alpha=1$,  $\beta=10$,  $\gamma=1$ and $mss$ importance sampling.
\end{itemize}

The second two variants TripletCLR~\cite{chechik2010large,schroff2015facenet} and SimCLR~\cite{chen2020simple} use contrastive approaches as training strategy for the encoder. Contrary to the VAE variants, these approaches drop the decoder networks and pixel-wise reconstruction as their training objective operates directly in the latent space. The encoders are trained to maximize agreement between differently augmented versions of the same observation $o$. We used to 2 variants for the contrastive loss: \begin{itemize}
    \item Triplet Loss:
    $ \mathcal{L}_{\textsc{TripletCLR}} (A,P,N) = \max \left( d(R(A), R(P)) - d(R(A), R(N)) + \alpha, 0 \right) $ where $R$ is the embedding network, $A$ is an anchor input (pattern $o$ in the training dataset), $P$ is the positive input (augmented version of $o$), $N$ is the negative input (other pattern $o'$ randomly sampled in the training dataset), $d(\cdot,\cdot)$ is the distance in the latent space (we use cosine similarity) and $\alpha$ is a margin between positive and negative pairs (we use $\alpha=1$)
    \item SimCLR Loss:  $\mathcal{L}_{\textsc{SimCLR}} (A,P) = - \log{ \frac{\exp{sim (z_A, z_P) / \tau}}{\sum_{N}  \mathbbm{1}_{[N \neq A]} \exp{sim (z_A, z_N) / \tau}} }$  where $\tau$ denotes the temperature parameter (we use $\tau=0.1$); $sim$ the similarity distance (we use cosine similarity); and $z$ represent the latent features onto which operates the contrastive loss. Please note that for this variant the encoder is coupled to a \textit{projection head} network $g(\cdot)$ such that $ o \overset{R}{\rightarrow} r \overset{g}{\rightarrow} z$ (We use g: FC 16$\rightarrow$16 + RelU, FC 16$\rightarrow$16). Here the positive pair ($A$, $P$) is contrasted with all negative pairs ($A$, $N$) in the current batch. The final loss is computed across all positive pairs in a mini-batch.
\end{itemize}

Finally the BigVAE baseline uses the same architecture and training strategy than the main VAE baseline, but with a much larger embedding capacity (368 dims instead of 16 dims) corresponding to the total embedding capacity of HOLMES if we would concatenate its 23 BC latent spaces.

\paragraph*{Results} The results are summarized in Figure~\ref{sm:fig:monolithic_baselines} and corroborate with the insights in the main paper. \begin{itemize}
    \item Lack of \textit{plasticity} for the all VAE variants, i.e. inability to adapt the learned features to novel niches of patterns. The bias that we observed in the main paper is confirmed  in the RSA analysis, even when changing the training objective or the encoding capacity. Interestingly, IMGEP-BetaVAE and IMGEP-BetaTCVAE show the same profile of discovered diversities than VAE (good at finding a diversity of SLPs but bad for TLPs) whereas the IMGEP-BigVAE seems to have a reversed bias (good at finding a diversity of TLPs but bad for SLPs). We attribute this effect to the difficulty of VAEs with low embedding capacity to capture textures with fine-grained structures (i.e. TLPs) whereas when given a higher encoding-capacity they can more accurately represent TLPs. Therefore the variants with small capacity representations seem better suited for exploring diverse SLPs (to the detriment of TLPs) whereas BigVAE seem better suited for exploring diverse TLPs (to the detriment of SLPs).

    \item Lack of \textit{stability} for all the contrastive variants, where features are drastically different from one training stage to the other. Contrary to the VAE variants, those approaches do not exhibit a strong \textit{bias} in their BC and therefore do not seem to differ much from the \textit{default} diversity found in Lenia (represented with the black curve), at least for the two \textit{types} of diversity measured in Figure~\ref{sm:fig:monolithic_baselines}.
\end{itemize}

\clearpage
\subsection{Ablation Study: Impact of the Lateral Connections}
\label{sm:subsec:impact_lateral_connections}

\begin{figure}[h!]
\setlength\tabcolsep{1pt}
\begin{tabular}{>{\centering\arraybackslash}m{2cm} >{\centering\arraybackslash}m{4cm} >{\centering\arraybackslash}m{9cm}}
 & \textbf{RSA Matrix}  & \textbf{RSA Statistics}   \\
 
  \textbf{\small no connection}  &
 \includegraphics[height=3cm]{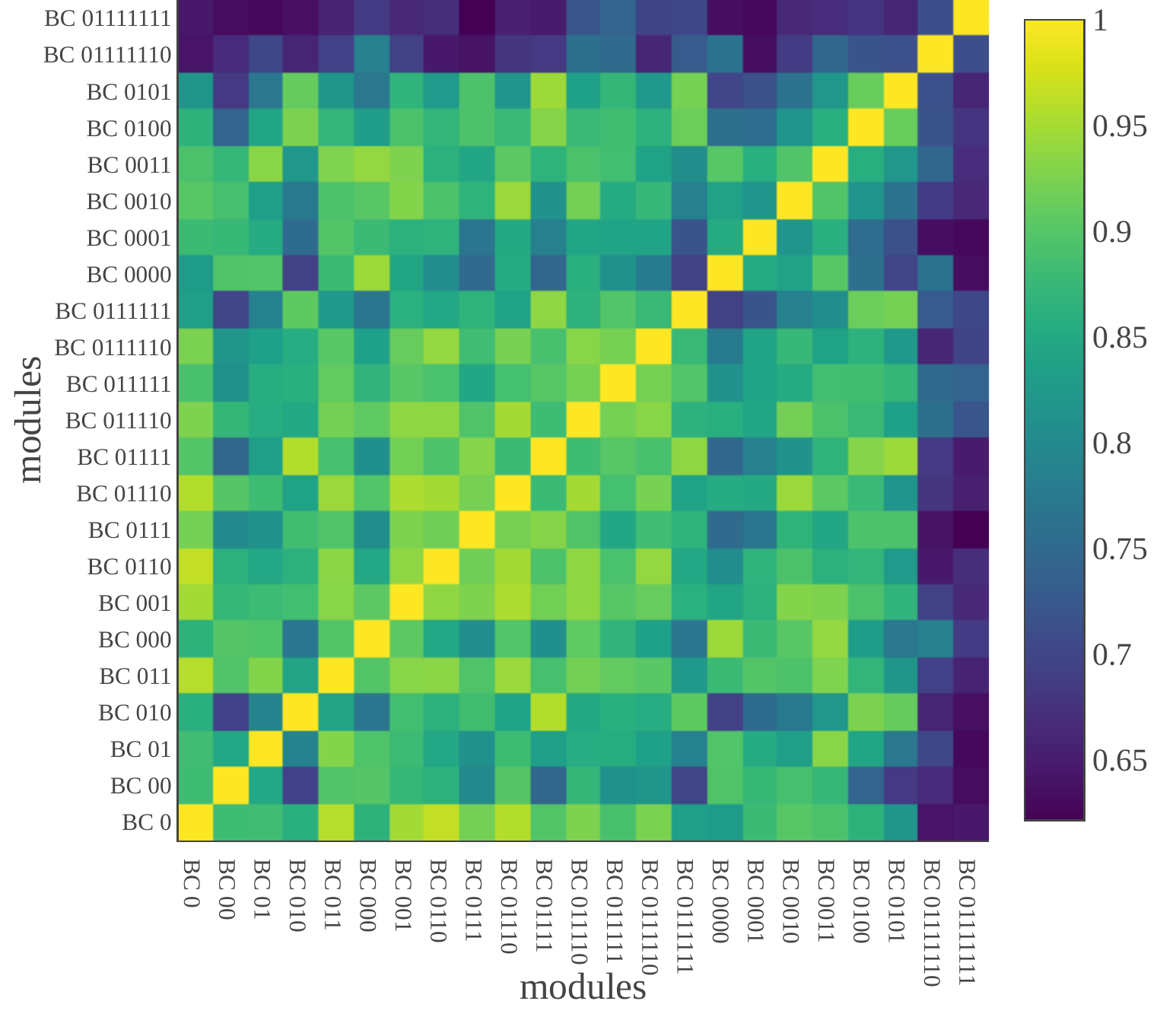}
 & \includegraphics[height=2.5cm]{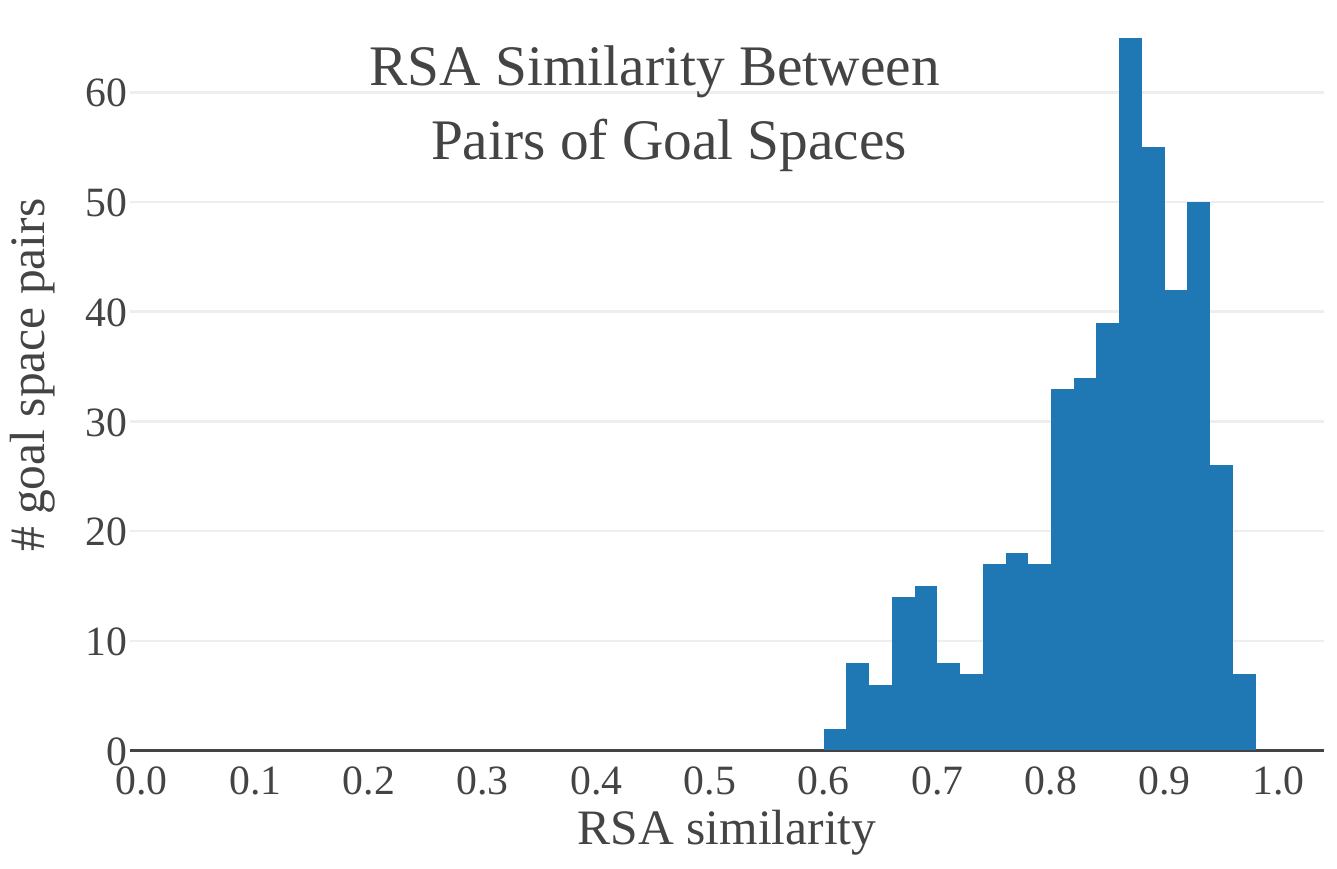}  \\

  \textbf{\small only lf\_c}  &
 \includegraphics[height=3cm]{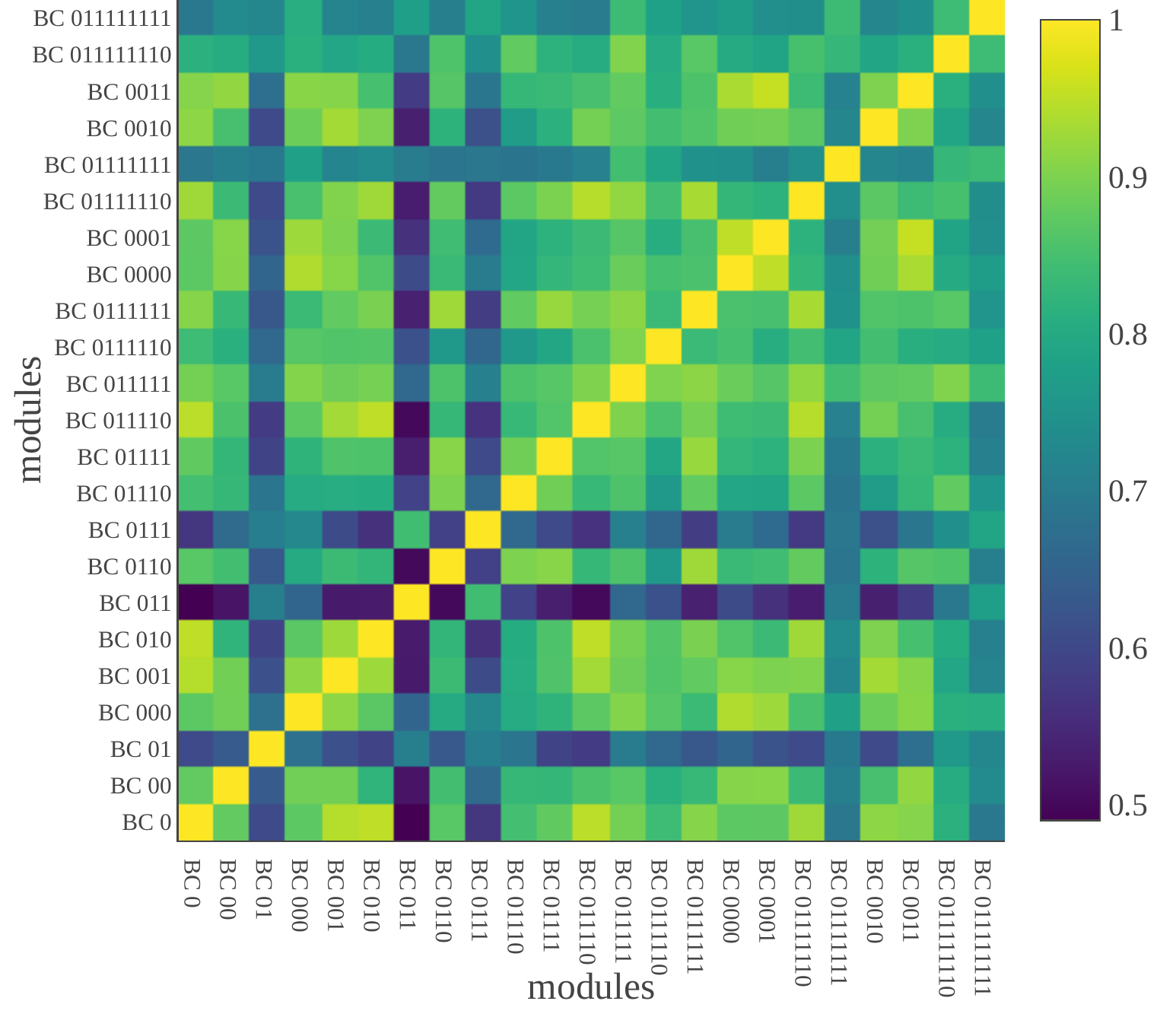}
 & \includegraphics[height=2.5cm]{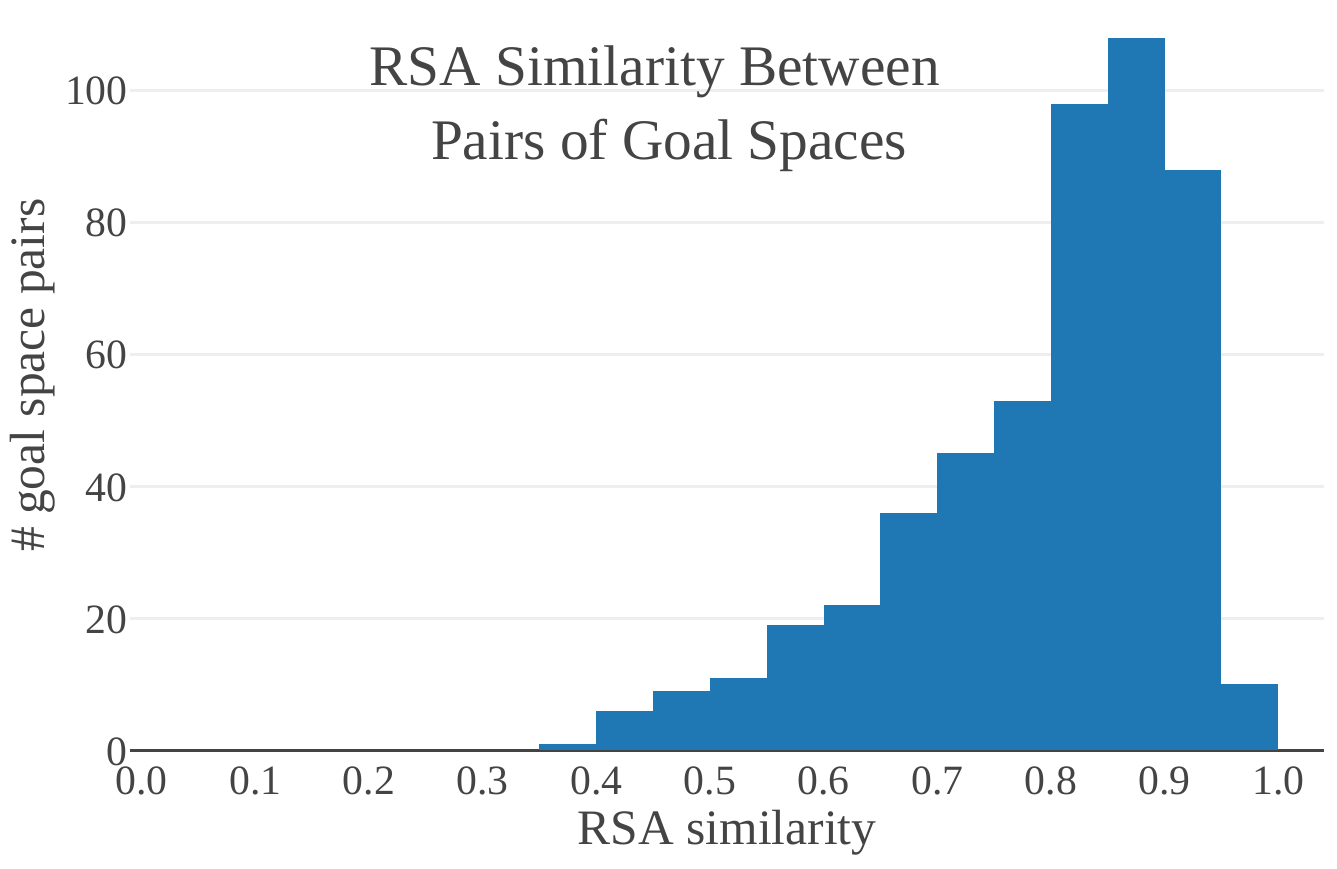}  \\

  \textbf{\small only gfi\_c}  &
 \includegraphics[height=3cm]{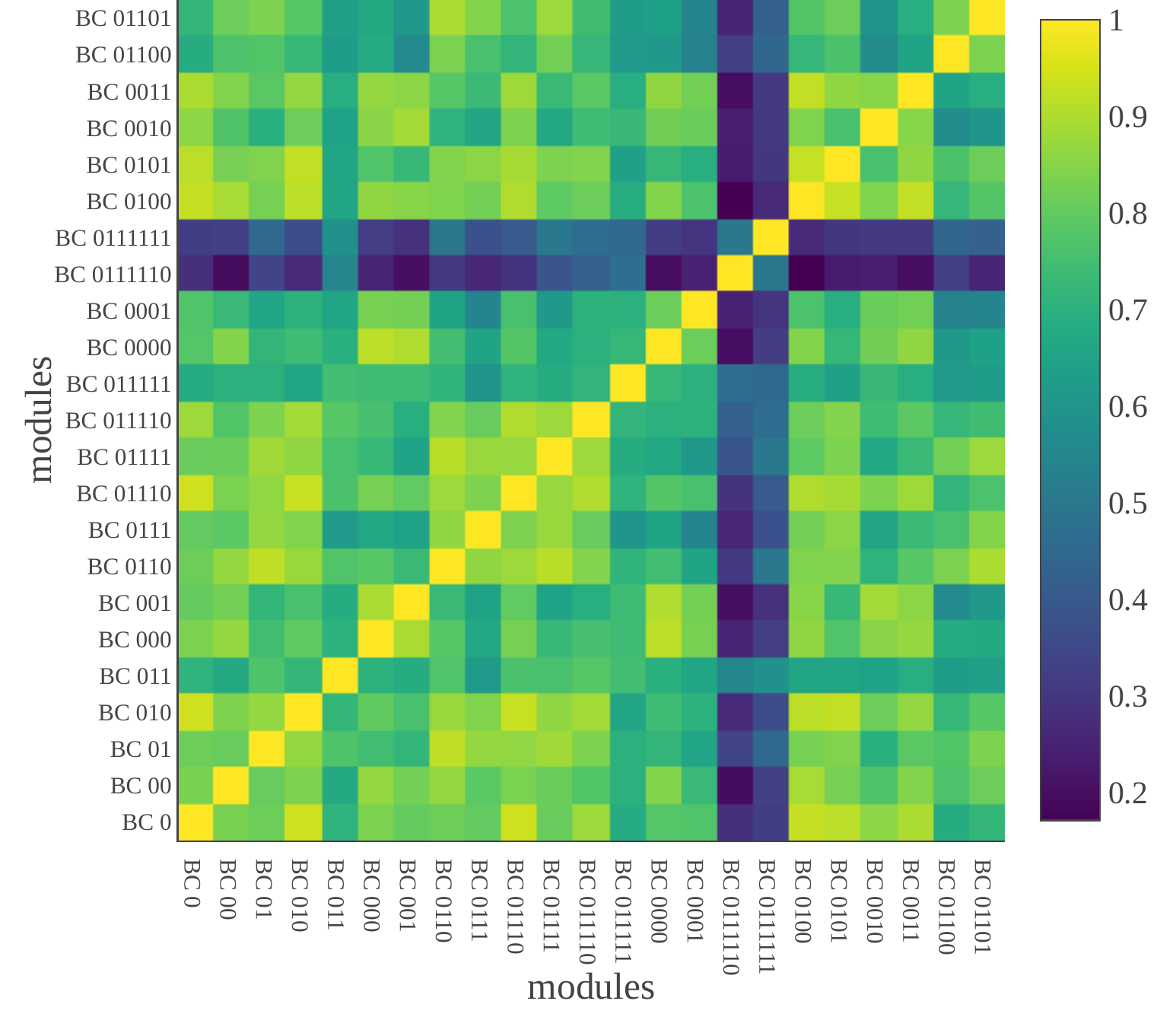}
 & \includegraphics[height=2.5cm]{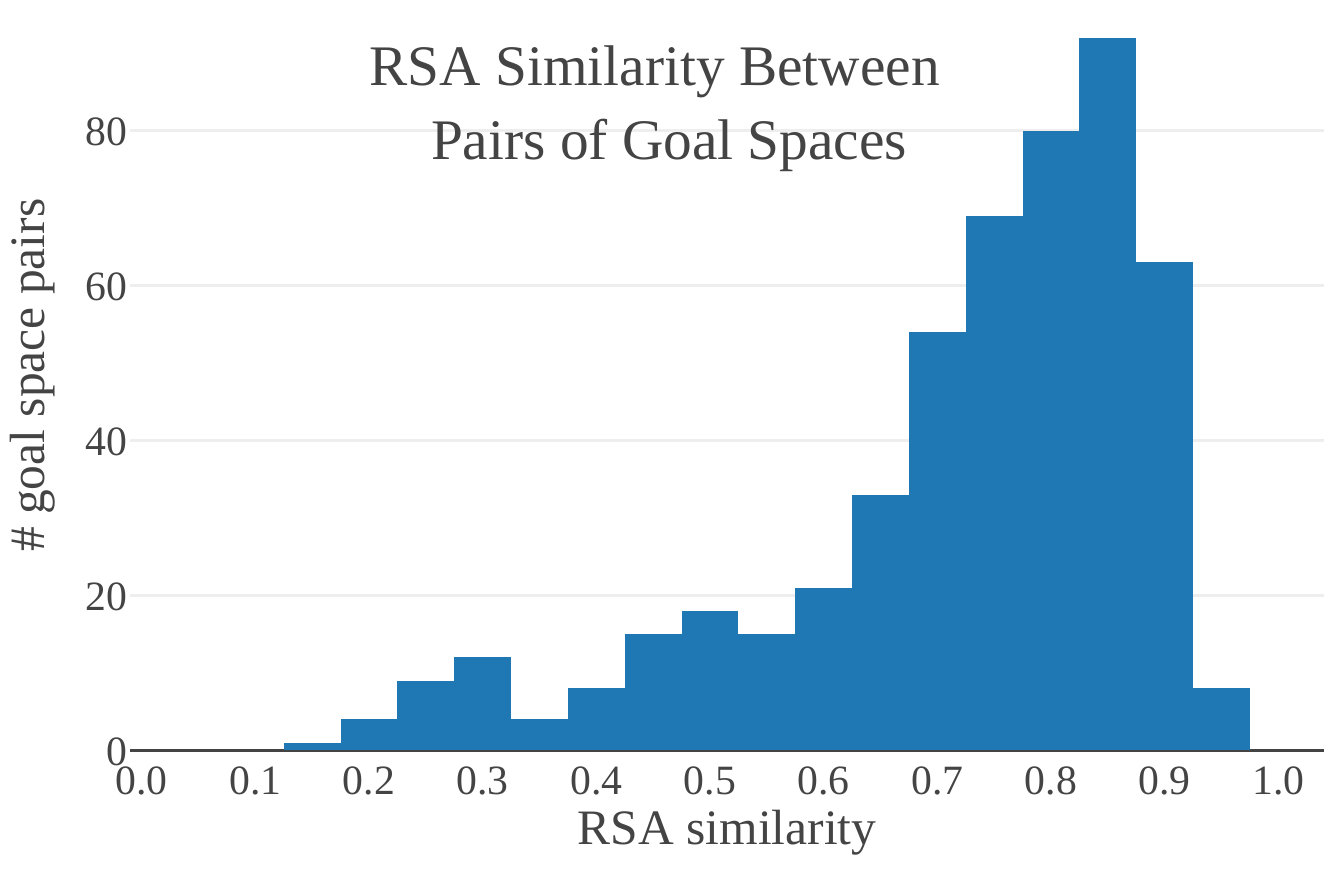}  \\

  \textbf{\small only lfi\_c}  &
 \includegraphics[height=3cm]{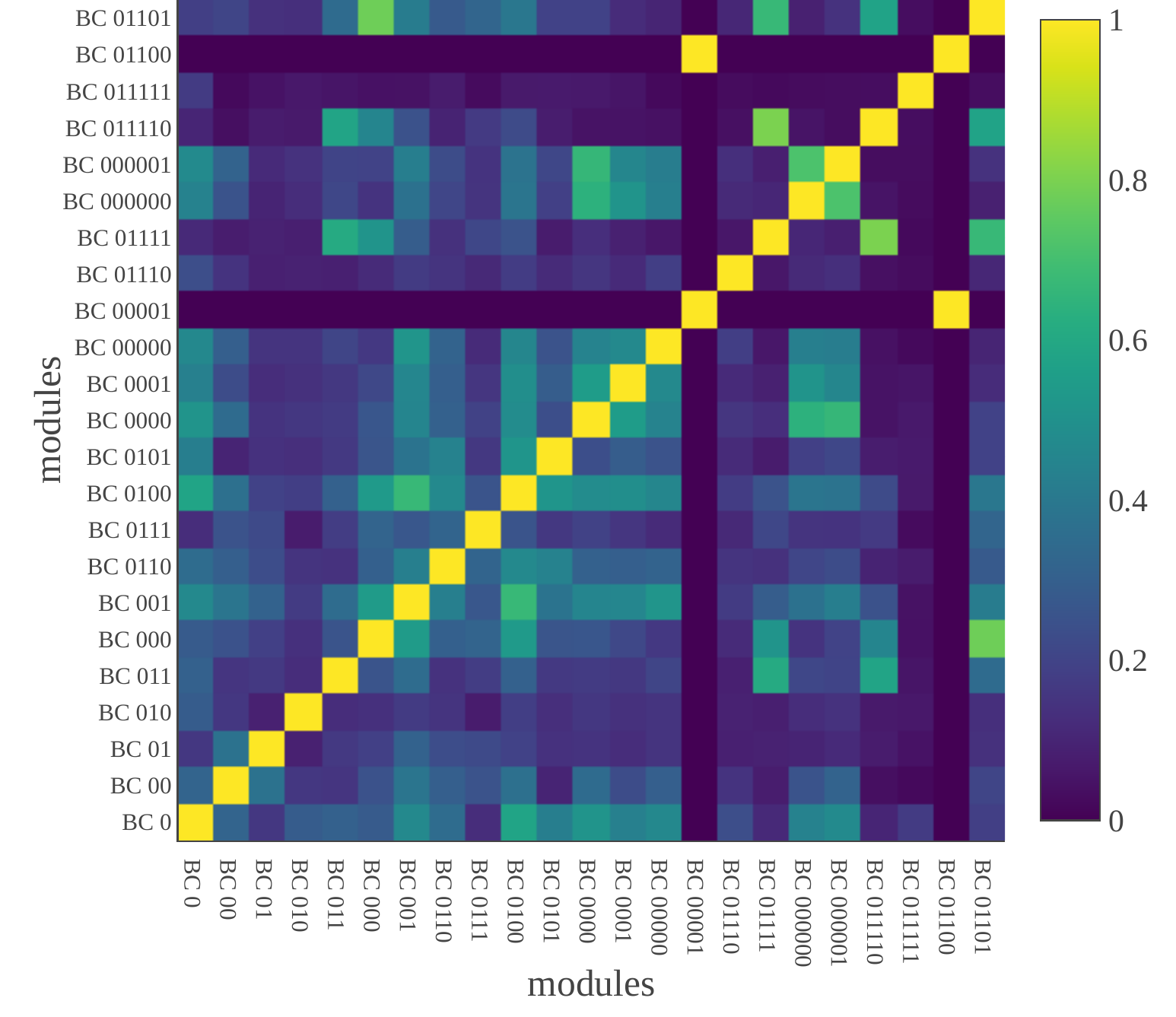}
 & \includegraphics[height=2.5cm]{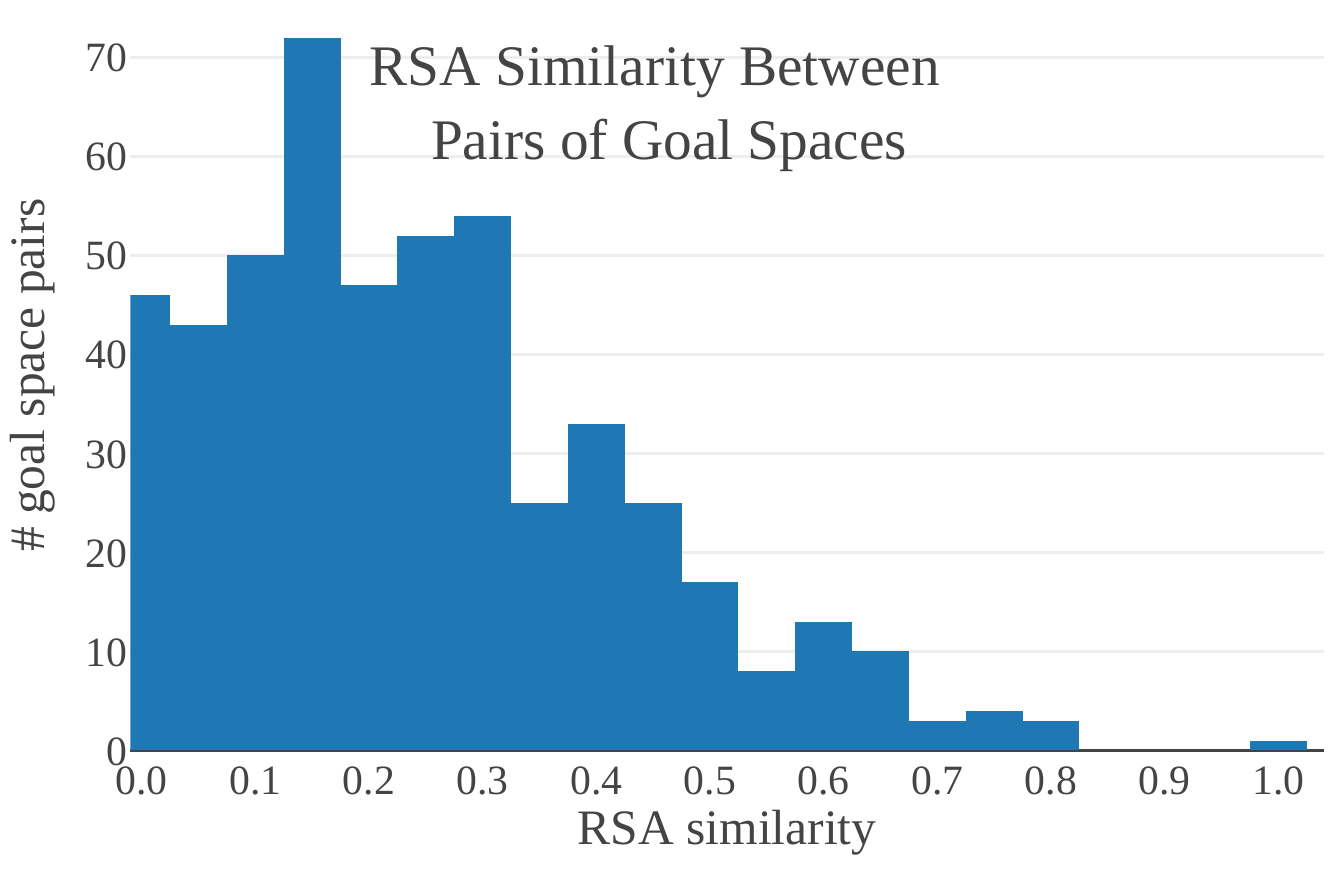}  \\

  \textbf{\small only recon\_c}  &
 \includegraphics[height=3cm]{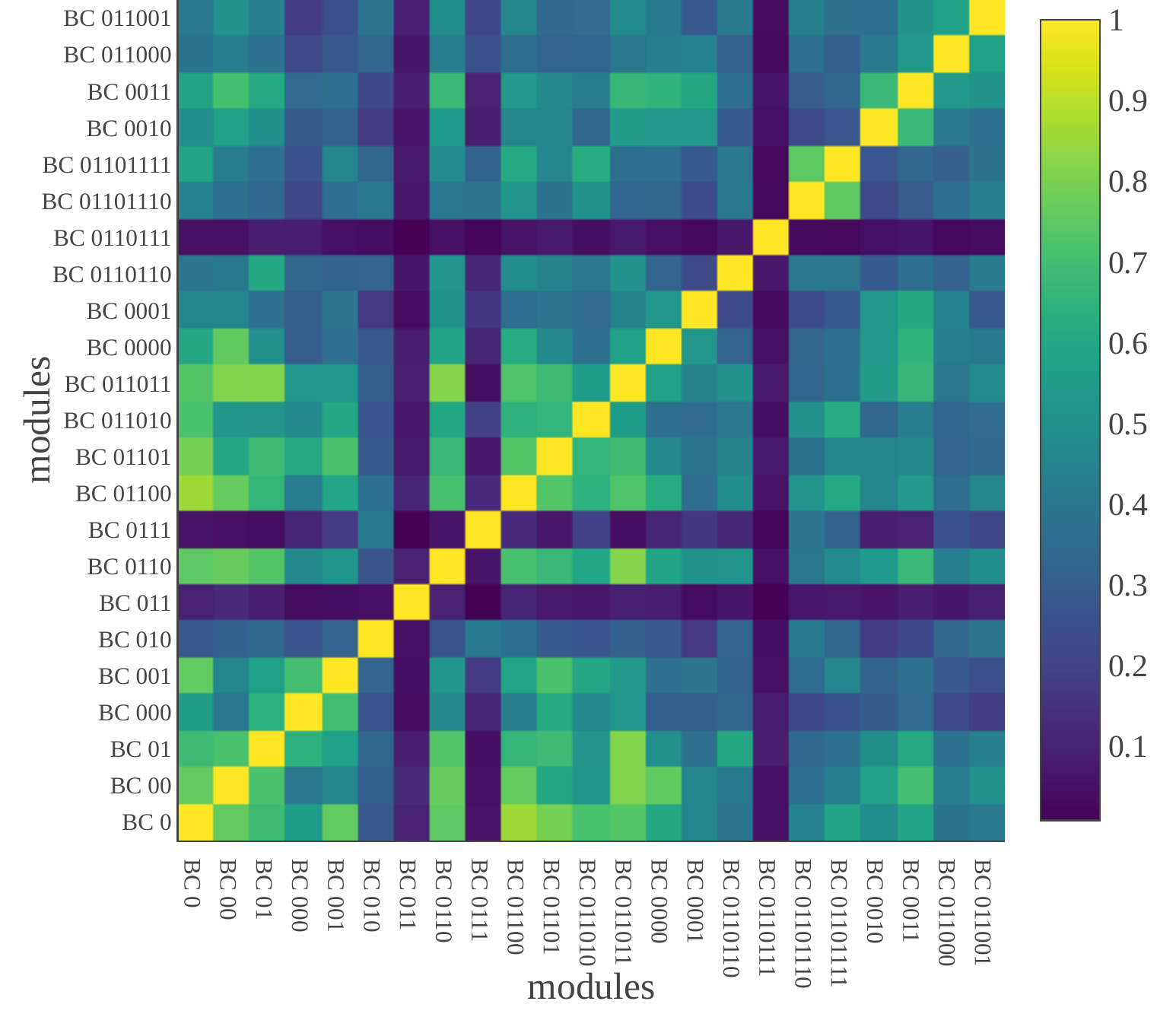}
 & \includegraphics[height=2.5cm]{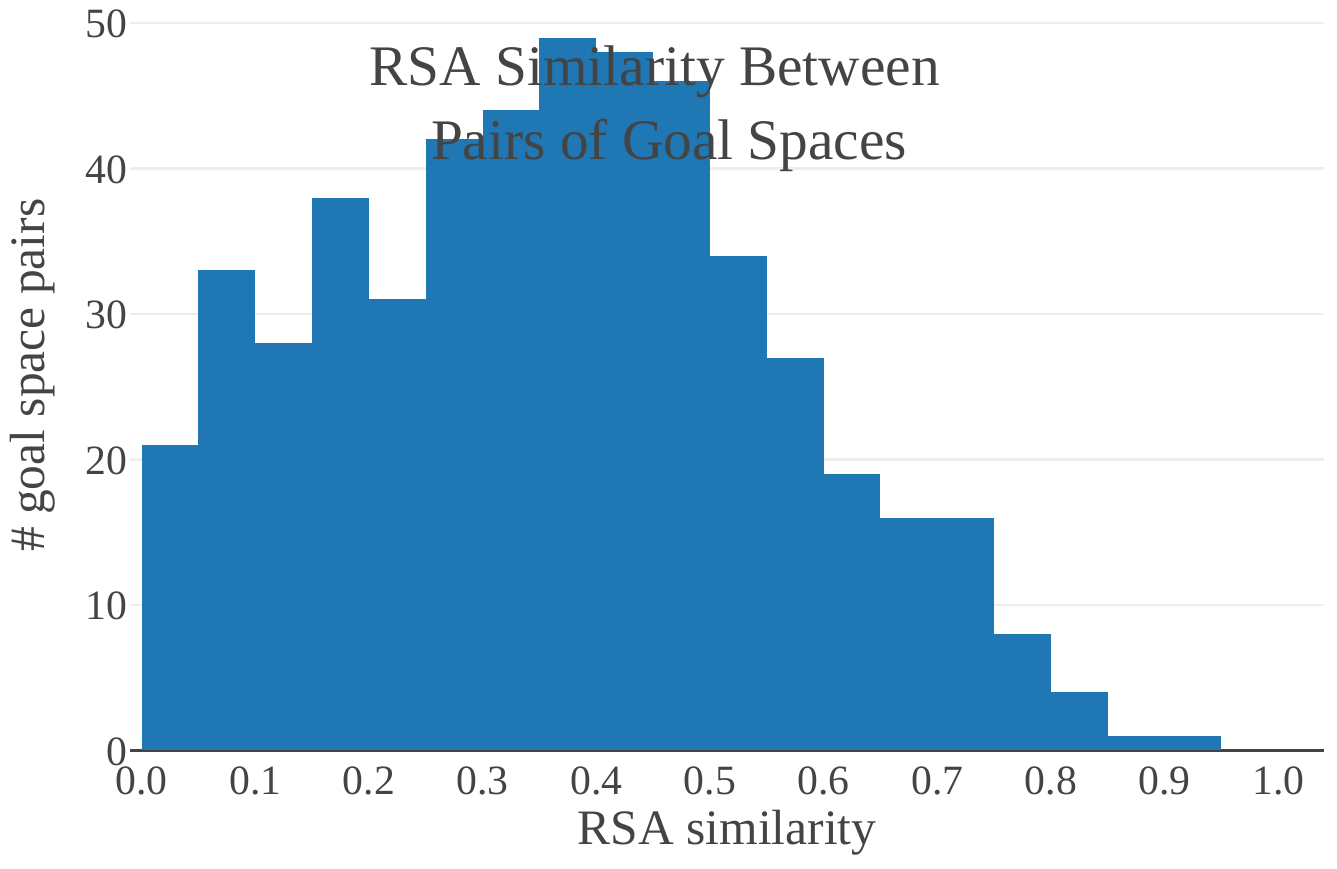}  \\

   \textbf{\small lf\_c +  gfi\_c +  lfi\_c + recon\_c (used in the main paper)}  &
 \includegraphics[height=3cm]{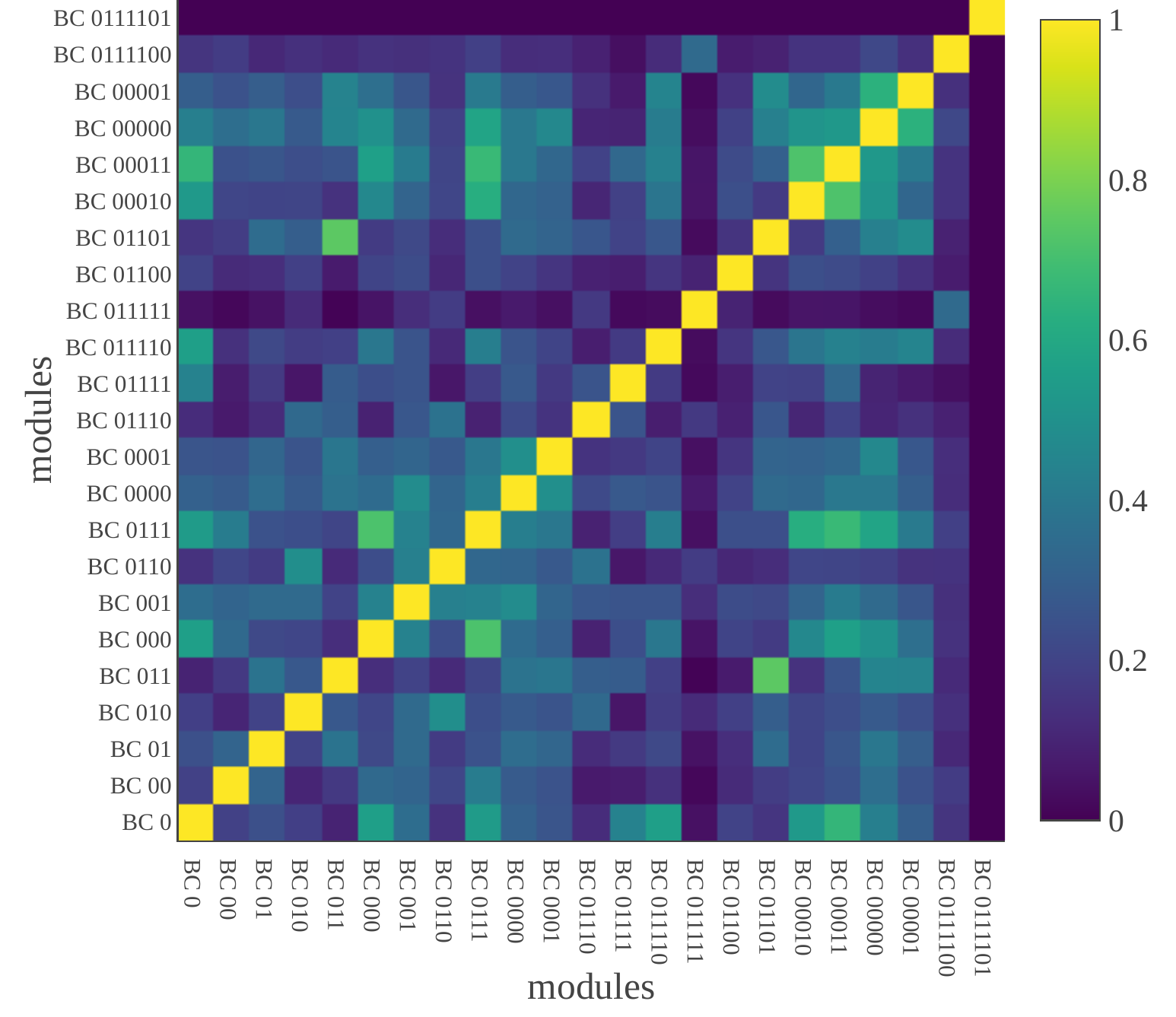}
 & \includegraphics[height=2.5cm]{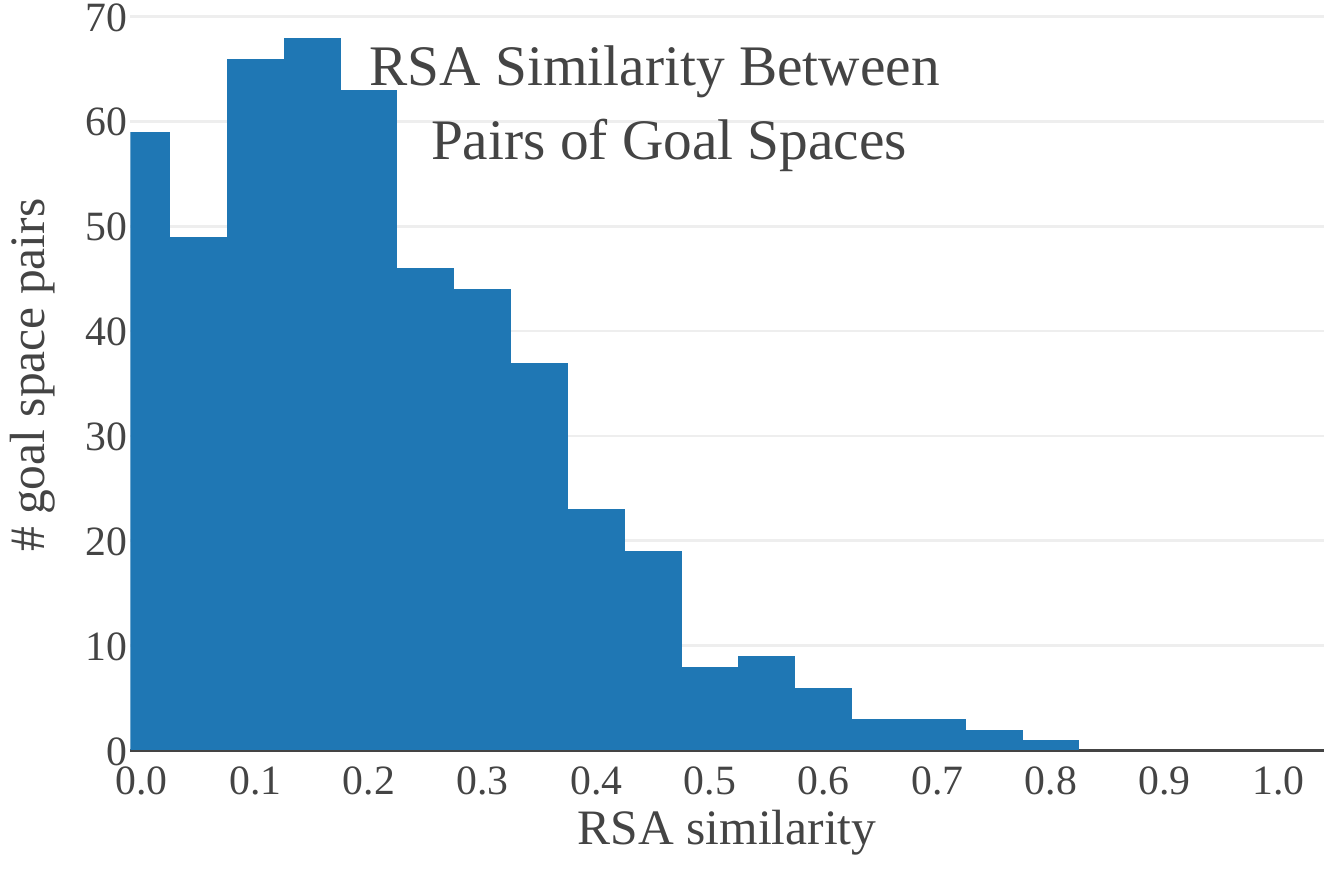}  \\

\end{tabular}
 \vspace{-10pt}
 \caption{RSA Analysis of the effect of the \textit{lateral connections} on the ability for HOLMES to learn \textit{diverse} module BCs. Each row is an ablation experiment with the corresponding connection scheme. (Left) RSA matrix for one experiment repetition (seed 0), with similarity index between 0 (dark blue, not similar at all) and 1 (yellow, identical). Representations are compared at the end of exploration between the different modules (ordered by their creation time on the left-to-right x-axis). (Right) Histogram of RSA index similarity between all pair of modules (aggregated over the 10 repetitions). }
\label{sm:fig:ablation_connections}
\end{figure}

We conducted 5 ablation experiments of the IMGEP-HOLMES variant presented in the main paper, each has 3 repetitions with different seeds. Each ablation experiment considered a different connection scheme with either zero or only one connection among \textbf{lf\_c}, \textbf{gfi\_c}, \textbf{lfi\_c} and \textbf{recon\_c} (proposed connections in HOLMES, see Figure~\ref{sm:fig:HOLMES_focus} and Table~\ref{sm:tab:holmesVAE_architecture}).
As shown in Figure~\ref{sm:fig:ablation_connections}, the lateral connections are \textit{essential} to learn \textit{diverse} behavioral characterizations among the different modules of the hierarchy. Indeed we can see that IMGEP-HOLMES without any connection (first row in the figure) learns BCs that are highly similar from one module to another (histogram concentrated around [0.8, 1] RSA indexes, i.e. very similar). We can also see that connections toward the last layers of the decoder are seemingly the more important (lfi\_c and recon\_c) as IMGEP-HOLMES with only one of such connection succeeds to learn dissimilar BCs per module (the histogram is shifted toward lower RSA indexes). However, the connection at the encoder level (lf\_c) and close to the embedding level (gfi\_c) seem less necessary, or at least alone are not sufficient to allow HOLMES modules escaping the bias inherent to the VAE learning (show a similar histogram of RSA indexes than the no-connection variant). The connection scheme used in the main paper (last row in the figure) seems to be the best suited to learn diverse BCs (histogram concentrated around [0.8, 1] RSA indexes, i.e. very dissimilar).

\clearpage
\section{Comparison of HOLMES with Related Methods}
\label{sm:sec:CURL}

In this section we provide a comparison of HOLMES with recent work in the literature: CURL~\cite{rao2019continual}, CN-FPM~\cite{lee2020neural} and pro-VLAE~\cite{li2020progressive}. Those approaches also propose to dynamically expand the network capacity of a VAE in the context of \textit{continual} representation learning, and therefore share similarities with HOLMES. In table~\ref{sm:table:HOLMES_comparison}, we provide a high-level comparison of the proposed approaches which compare the different architectures according to their \textit{structural bias}, handling of \textit{catastrophic forgetting}, architecture for \textit{dynamic expansion}, handling of \textit{transfer} between the different group of features, criteria for the \textit{expansion trigger}, and if they are performing \textit{data partitioning} (i.e. learn different set of features for different niches of observations).

\begin{table}[h!]
  \caption{High-level comparison of the general choices of HOLMES with those of previous methods: CURL~\cite{rao2019continual}, CN-FPM~\cite{lee2020neural} and pro-VLAE~\cite{li2020progressive}. Please refer to the original papers for more details. }
  \label{sm:table:HOLMES_comparison}
  \centering
  \resizebox{\linewidth}{!}{%
  \begin{tabular}{lcccc}
    \toprule
        & \textbf{CURL} &  \textbf{CN-DPM} &  \textbf{pro-VLAE} & \textbf{HOLMES} \\
    \midrule
    \makecell{\textbf{Structural} \\ \textbf{Bias}} & \makecell{Mixture of Gaussians \\ (in a single VAE latent space)} & \makecell{Dirichlet Process Mixture \\ (flat set of VAE modules)} & \makecell{Hierarchical Levels \\ (in a single VAE)} & \makecell{Hierarchical Mixture \\ (binary tree of VAE modules)} \\
    
    \midrule
    
    \makecell{\textbf{Catastrophic} \\ \textbf{Forgetting}} & Generative Replay & Freeze & 	\makecell{\enquote{fade-in} coefficient \\ (same network) } & Freeze\\
    
    \midrule
    
    \makecell{\textbf{Dynamic} \\ \textbf{Expansion}} & \makecell{New component \\ in the MoG} &	\makecell{New module \\ VAE} & \makecell{New feature layer  \\ in the VAE} & \makecell{New module \\ VAE}\\
    
    \midrule
    
    \textbf{Transfer} &  \makecell{Single Shared Network \\ with several \enquote{heads}} & \makecell{Lateral connections \\ (exhaustive as in PNN~\cite{rusu2016progressive}))} &   \makecell{Single Shared Network \\ with several \enquote{levels}}  &	\makecell{Lateral connections  \\ (parent-to-children only)}\\
    
    \midrule
    
    \makecell{\textbf{Expansion} \\ \textbf{Trigger}} & \makecell{Short-Term \\ Memory Size} & \makecell{Short-Term \\ Memory Size} & Predetermined  & \makecell{Node \\ Saturation} \\

    \midrule
    
    \makecell{\textbf{Data} \\ \textbf{Partitioning}} & \makecell{Soft partitioning \\ }  & \makecell{Soft partitioning \\ (coupling of each VAE with discriminator)} & 	\makecell{None \\ (same network) }  & \makecell{Hard Partitionning \\ (boundary in $BC_i$)} \\
    
    \bottomrule
  \end{tabular}
  }
\end{table}

As we can see, while HOLMES shares \textit{conceptual} ideas with those approaches, our approach has key differences: \begin{enumerate}
    \item It uses a hierarchy of different latent spaces whereas CURL uses a single latent space, CN-DPM uses a flat set of different latent spaces and pro-VLAE uses a fixed-set of latent spaces (different levels in one network)
    \item CURL and CN-DPM show results in the context of continual multi-task classification and demonstrate that their modular architecture can separate well the latents allowing to unsupervisedly discriminate between the different input observations / tasks (eg: discriminate digits in MNIST at test time when they have been sequentially observed at train time). However, CURL does not use different features for the different niches of observations and it is not clear if the flat approach of CN-DPM does learn different features between the different modules. However HOLMES targets to learn dissimilar set of features per BC in order to achieve \textit{meta-diversity}. 
    \item Pro-VLAE is not applied in the context of continual learning but rather proposes to progressively learn features at different levels in the VAE layers, showing that it can successfully disentangle the features. Even though disentanglement is a key property to avoid redundant features, we believe that it is also key to have diverse set of features for the different niches of observed instances. 
\end{enumerate}

\clearpage
\section{Additional Visualisations}
\label{sm:sec:additional_visualisations}

\begin{figure}[h!]
\centering
 \includegraphics[width=\linewidth]{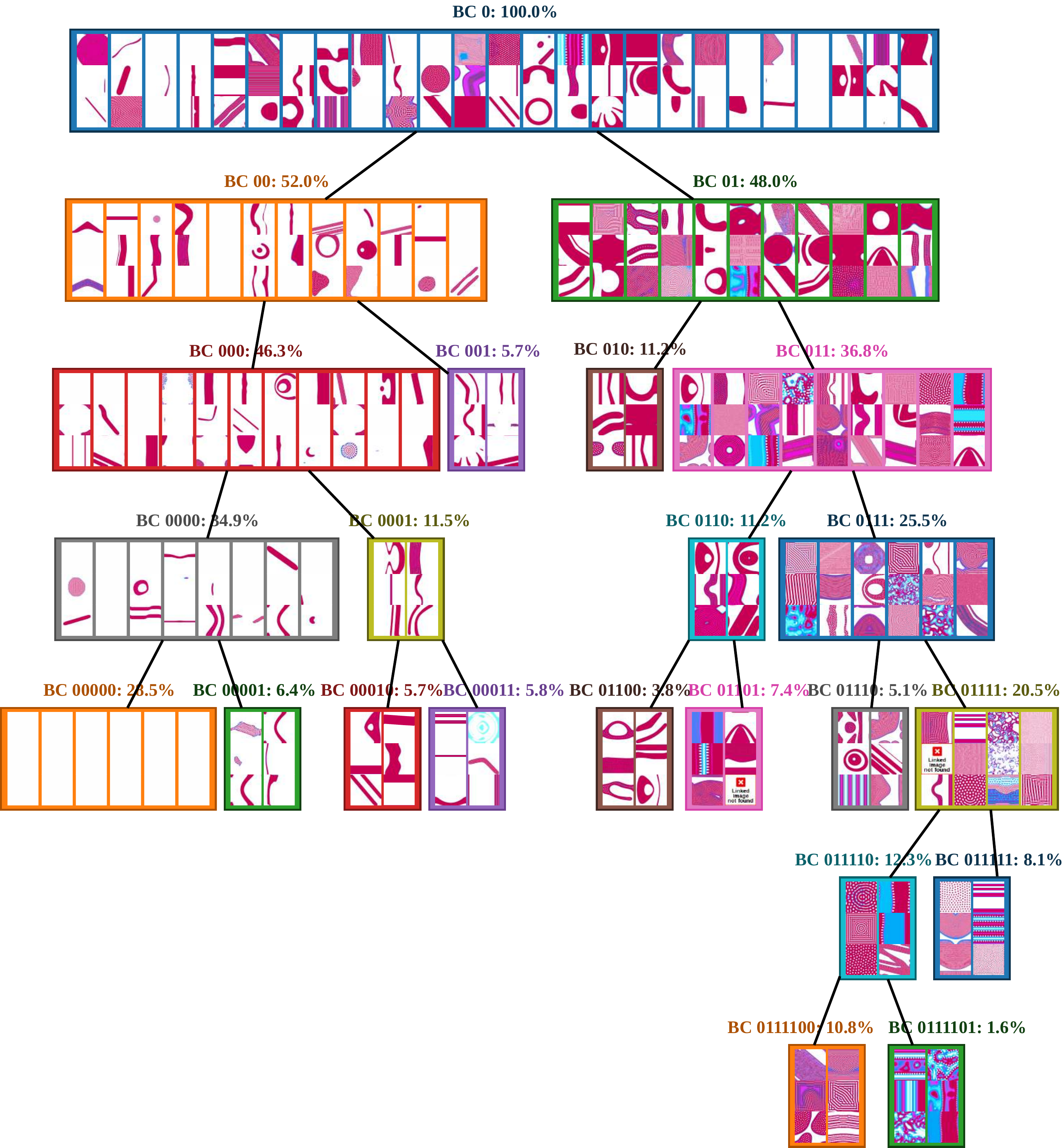} 
 \caption{Examples of patterns discovered by IMGEP-HOLMES (non-guided) within the learned tree hierarchy. The hierarchy is the same as in Figure~\ref{sm:fig:rsa-hierarchy}. In each node is displayed the percentage of discovered patterns directed through that node, as well as a set of pattern images representative of the diversity within that node (the set is built with the procedure described in section~\ref{sm:subsec:human_evaluation}). The number of patterns per node reflects the indicated percentage.}
\label{sm:fig:non-guided-holmes}
\end{figure}

Figure~\ref{sm:fig:non-guided-holmes} shows examples of patterns discovered by IMGEP-HOLMES (non-guided) within the learned tree hierarchy. The patterns shown in the root node are representative of the diversity of all the discovered patterns in that particular run (100\% of the patterns). The boundaries fitted when splitting each non-leaf node (see procedure in section~\ref{sm:subsec:HOLMES}) makes each pattern follow a particular path in the hierarchy, from the root node to a leaf node. Goals are sampled by the IMGEP by first sampling a leaf uniformly among all existing leafs, then sampling uniformly in the hypercube fitted around currently reached goals within that leaf (see section~3.2 of the main paper). However, we observe that the percentage indicated in each node does not reflect this uniformity (for example, only 5.7\% of the patterns fall in the leaf BC 001). The interpretation is that leafs with low percentages correspond to unstable niches: when a goal is sampled in such a leaf, the small mutation applied in the parameter-sampling policy is sufficient to produce a pattern which is different enough to fall in another leaf.

We qualitatively observe in Figure~\ref{sm:fig:non-guided-holmes}  that the boundaries fitted during the splitting procedure tend to separate the patterns into visually distinct categories. For example, the proportion of TLPs is much higher in BC 01 compared to BC 00 ; the leaf BC 00000 contains only blank patterns while its sibling BC 00001 contains only SLPs ; the nodes below BC 01111 (bottom-right of the tree) contains only TLPs.

\begin{figure}[h!]
\centering
 \includegraphics[width=\linewidth]{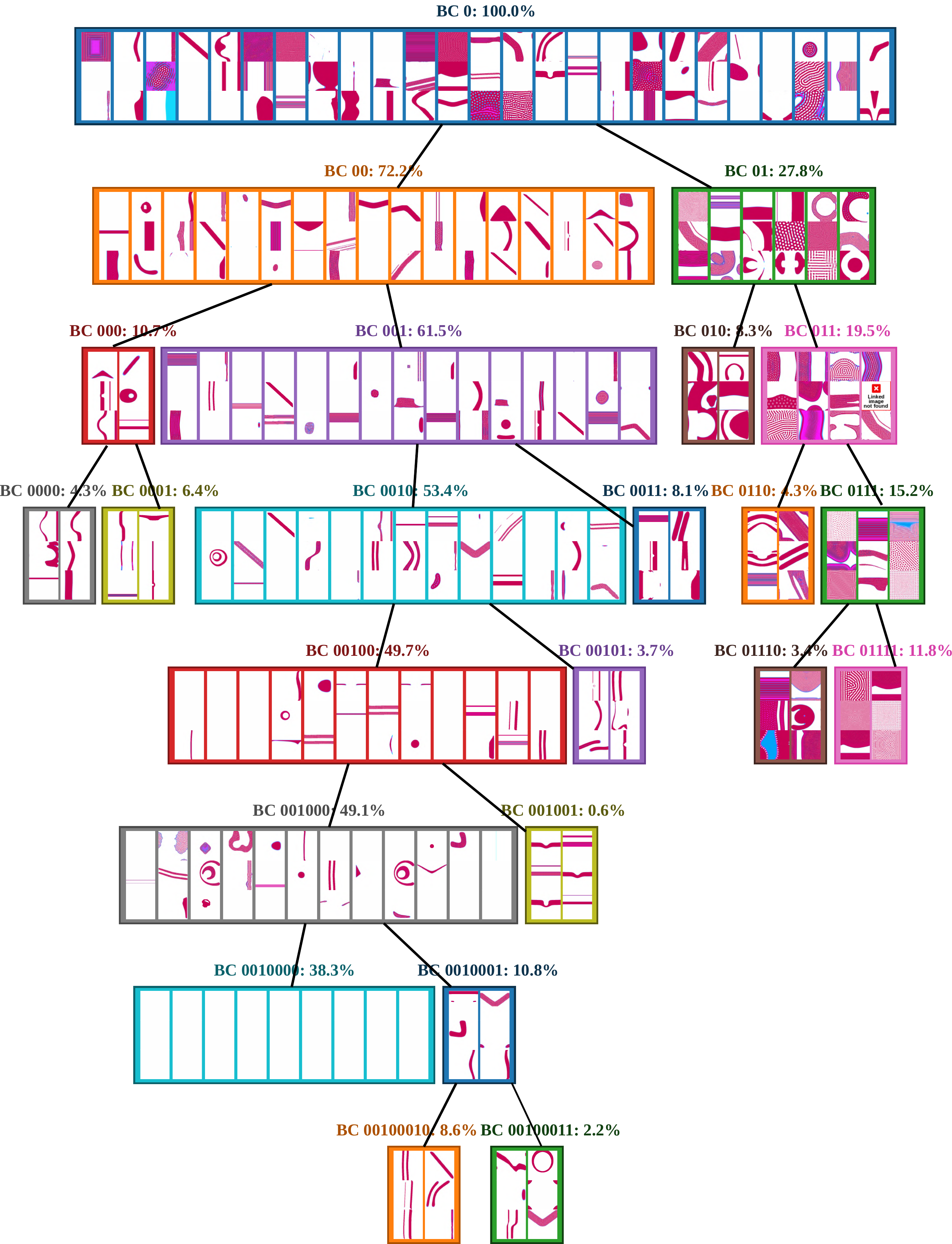} 
 \caption{Examples of patterns discovered by IMGEP-HOLMES within the learned tree hierarchy, when guided towards SLPs through simulated user feedback as described in section 4.3 of the main paper. Same convention as in Figure~\ref{sm:fig:non-guided-holmes}.}
\label{sm:fig:slp-holmes}
\end{figure}

\begin{figure}[h!]
\centering
 \includegraphics[width=\linewidth]{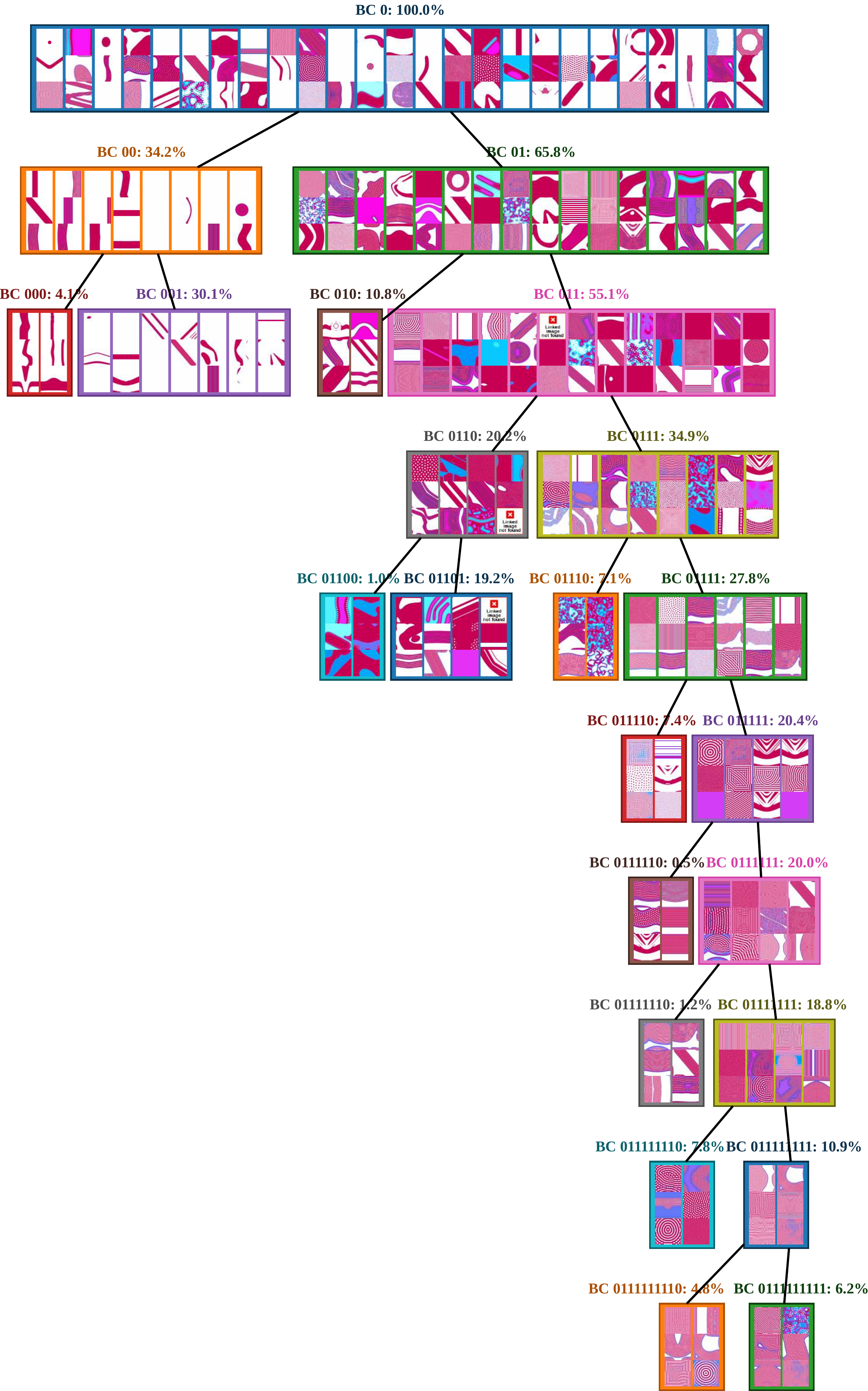} 
 \caption{Examples of patterns discovered by IMGEP-HOLMES within the learned tree hierarchy, when guided towards TLPs through simulated user feedback as described in section 4.3 of the main paper. Same convention as in Figure~\ref{sm:fig:non-guided-holmes}.}
\label{sm:fig:tlp-holmes}
\end{figure}

Figures~\ref{sm:fig:slp-holmes} and \ref{sm:fig:tlp-holmes} show discovered patterns when IMGEP-HOLMES is guided towards SLPs or TLPs, respectively, through simulated user feedback as described in section 4.3 of the main paper. We observe that the user guidance is able to dramatically affect both the diversity of the discovered patterns and the structure of the hierarchy . When guided towards SLPs, most of the discovered patterns are SLPs (most TLPs in Figure~\ref{sm:fig:slp-holmes} are concentrated in the leafs BC 01110 and BC 01111 which represent approximately 15\% of the discovered patterns). On the contrary, when guided towards TLPs, most of the discovered patterns are TLPs (most SLPs in Figure~\ref{sm:fig:tlp-holmes} are concentrated in the leafs BC 000 and BC 001 which represent approximately 34\% of the discovered patterns). As a consequence of this bias toward either SLPs or TLPs, we observe that HOLMES has created more branches in the direction of the desired patterns (either SLPs or TLPs) in order to enrich their corresponding representations.

Additional visualisations can be found on the project website \url{http://mayalenE.github.io/holmes/}.

\end{document}